\pdfoutput=1
\documentclass[11pt,onecolumn]{cleantechnicalreport}

\usepackage[authoryear,compress,round]{natbib}

\usepackage{hybridfrontmatter}
\usepackage{bookmark} %
\usepackage{longtable}
\usepackage{array}
\usepackage{multirow}
\usepackage{xspace}
\usepackage{flafter} %
\usepackage{xurl} %
\usepackage{tikz}
\usetikzlibrary{arrows.meta,positioning,fit,backgrounds,calc,shapes.geometric,decorations.pathreplacing}
\usepackage{pgfplots}
\pgfplotsset{compat=1.18}

\usepackage{titlesec}
\titlespacing*{\paragraph}{0pt}{3pt}{3pt} %

\newcolumntype{L}[1]{>{\raggedright\arraybackslash}p{#1}}

\AtBeginDocument{\setlength{\LTcapwidth}{\linewidth}}

\definecolor{STwoColor}{HTML}{4263EB}    %
\definecolor{HarnessColor}{HTML}{4263EB} %
\definecolor{FMColor}{HTML}{7950F2}      %
\definecolor{SOneColor}{HTML}{1098AD}    %
\definecolor{SZeroColor}{HTML}{F76707}   %
\definecolor{WorldColor}{HTML}{868E96}   %
\definecolor{EnvColor}{HTML}{37B24D}     %
\definecolor{SafetyColor}{HTML}{E03131}  %
\definecolor{ToolColor}{HTML}{AE3EC9}    %
\definecolor{EmbColor}{HTML}{D9480F}     %

\newif\ifreviewchanges
\reviewchangestrue
\DeclareRobustCommand{\oldtext}[1]{\ifreviewchanges{\color{blue}\hypersetup{citecolor=blue,linkcolor=blue,urlcolor=blue}#1}\fi}

\title{Envisioning Agentic Robotics: A New Era of Robotic Systems}
\author{Grid Robotics}
\reportwebsite{https://grid-robotics-official.github.io/agentic-robotics-roadmap}
\reportcontact{contact@gridrobotics.space}
\begin{abstract}

The release of GPT-6 Astra marks the emergence of a new paradigm in robotics: \textit{agentic robotics}. Within weeks, frontier models with agentic harnesses matched or exceeded task-specific policies on many embodied benchmarks. Instead of training a dedicated policy for each family of tasks, an agentic robotic system applies a foundation model to open-world embodied tasks. It operates in a closed loop: observing the environment, interpreting user intent, reasoning about the task, invoking skills and tools, verifying outcomes, and adapting its strategy. Despite these advances, the community has yet to converge on a shared definition, a reliable evaluation framework, or a research roadmap.
We argue that \emph{Embodied System 2}, realized by a foundation model together with an embodied harness, forms the decision-making core of agentic robotics.
Its central challenge is to translate general-purpose reasoning into effective and verifiable actions under physical constraints such as limited reaction time, mechanical limits, unpredictable environment changes, and irreversible consequences. 
Motivated by this perspective, we propose an evaluation framework that measures both model capabilities and closed-loop system performance. 
Furthermore, we identify the principles that carry over from digital to physical agents and the challenges that must be addressed to achieve reliable physical autonomy.
On top of the Embodied System 2 and the evaluation framework, we envision a research roadmap toward sustained autonomy: the ability to perform work safely and reliably with progressively broader responsibility, from executing individual primitives to fulfilling standing roles.

\end{abstract}

\begin{document}
\maketitle
\vspace*{-6mm}
\par\noindent
\begin{minipage}{\linewidth}
\captionsetup{type=figure}
\centering
\includegraphics[width=0.98\linewidth]{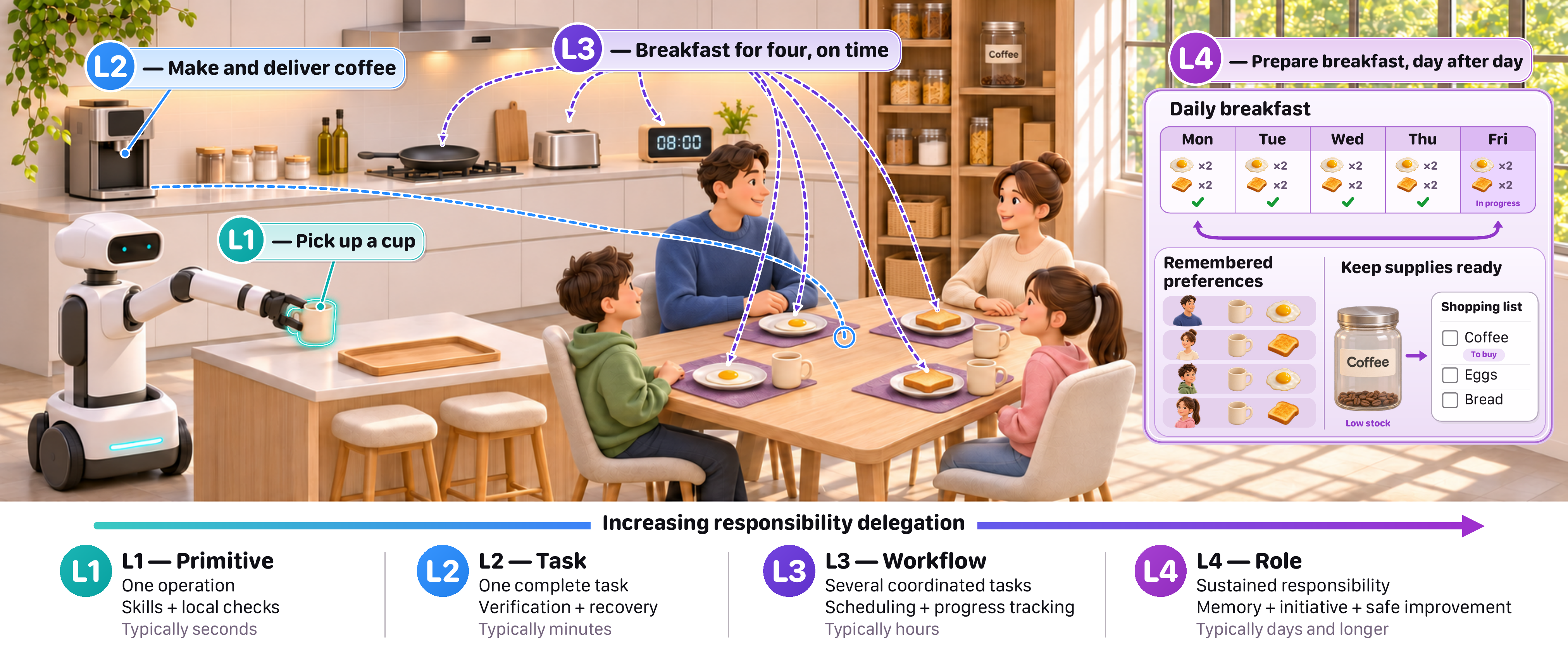}
\caption{\textbf{A vision for agentic robotics: from executing tasks to fulfilling roles.} A home breakfast scenario illustrates progressively broader responsibility entrusted to robots: from executing a primitive (L1), completing a task (L2), coordinating a workflow (L3), to fulfilling a standing role (L4).}
\label{fig:teaser}
\end{minipage}
\par

\clearpage
\tableofcontents
\clearpage

\section{Introduction}
\label{sec:introduction}

A central goal of robotics is to enable robots to translate human intent into desired outcomes in the physical world. Early autonomous robots often integrated hand-engineered components for perception, planning, and control within modular or layered architectures \citep{nilsson1984shakey,gat1998three}. 
Since then, advances in deep learning have allowed robots to learn increasingly complex behaviors.
Policies learned in simulation, mostly with deep reinforcement learning, have produced agile locomotion \citep{hwangbo2019anymal} and versatile humanoid whole-body control by tracking human motion data \citep{liao2025beyondmimic,luo2026sonic,liao2026supermimic}. 
However, these simulation-trained policies remain specialized to specific tasks and embodiments, limiting their ability to generalize across diverse and open-ended environments.
To move beyond task-specific controllers and enable robots to follow human instructions, the field has increasingly turned toward models that unify perception, language understanding, and action generation within a single framework.
Among them, vision-language-action (VLA) models, exemplified by the $\pi$ series, combine vision-language pretraining with diverse robot data to learn general policies that map visual observations and language instructions to actions \citep{pi2025pi05,black2025pi0}. 
More recently, world-action models (WAM) integrate video-based modeling of physical dynamics with robot policy learning \citep{dreamzero,fastwam,li2026causal}. Together, these advances have broadened robot capabilities. Yet reliable generalization remains a central challenge, as policy performance can still deteriorate when scenes or instructions change \citep{zhou2025liberopro}.

\begin{figure}[t]
\centering
\begin{tikzpicture}[
  font=\small,
  execute at begin node={\hyphenpenalty=10000\relax},
  box/.style={draw=#1!80!black, fill=#1!9, rounded corners=3pt, line width=0.6pt,
    text width=5.0cm, align=center, inner sep=4pt, minimum height=0.85cm},
  half/.style={draw=#1!80!black, fill=#1!9, rounded corners=3pt, line width=0.6pt,
    text width=2.25cm, align=center, inner sep=4pt, minimum height=0.85cm},
  plain/.style={text width=5.3cm, align=center, inner sep=2pt},
  arr/.style={-{Latex[length=2.2mm]}, semithick, black!80},
  fb/.style={-{Latex[length=2.2mm]}, semithick, black!50, densely dashed, rounded corners=4pt},
  tag/.style={font=\footnotesize\itshape, black!65},
  node distance=4.5mm
]
\node[plain] (ah) {\textbf{(a) Policy-centric robotics}};
\node[plain, minimum height=0.9cm, below=of ah] (ain) {task instruction};
\node[box=SOneColor, below=of ain] (apol) {\textbf{Action policy (S1)}\\ \footnotesize reasoning and subtask prediction might be included};

\node[plain, right=0.8cm of ah] (bh) {\textbf{(b) Agentic robotics}};
\node[plain, minimum height=0.9cm, below=of bh] (bin) {user intent};
\node[box=STwoColor, below=of bin] (bs2) {\textbf{Foundation model and embodied harness (S2)}\\[2pt]
  {\footnotesize interpret intent, orchestrate skills and tools, verify outcomes, and adapt decisions}};
\node[draw=SOneColor!80!black, fill=SOneColor!9, rounded corners=3pt, line width=0.6pt,
  minimum width=5.28cm, minimum height=1.92cm, below=of bs2] (bs1) {};
\node[anchor=north, font=\small\bfseries, inner sep=3pt] at (bs1.north) {Skills and tools (S1)};
\node[draw=SOneColor!80!black, fill=white, rounded corners=2pt, line width=0.5pt,
  text width=4.85cm, align=center, inner sep=2.5pt, minimum height=0.55cm, anchor=north,
  font=\footnotesize] (bpol) at ($(bs1.north)+(0,-0.5)$) {\textbf{Policies and skills}: VLA, WAM};
\node[draw=SOneColor!80!black, fill=white, rounded corners=2pt, line width=0.5pt,
  text width=4.85cm, align=center, inner sep=2.5pt, minimum height=0.55cm, anchor=north,
  font=\footnotesize] (btool) at ($(bpol.south)+(0,-0.12)$) {\textbf{Tools}: perception, planning, code};
\node[box=SZeroColor, below=of bs1] (bctl) {\textbf{Control (S0)}};
\node[box=WorldColor, below=of bctl] (brob) {\textbf{Robot}};
\node[box=EnvColor, below=of brob] (benv) {\textbf{Environment}};
\node[box=EnvColor] (aenv) at (apol |- benv) {\textbf{Environment}};
\node[box=SZeroColor] (actl) at ($(apol)!0.333333!(aenv)$) {\textbf{Control (S0)}\\ \footnotesize turn action targets into execution};
\node[box=WorldColor] (arob) at ($(apol)!0.666667!(aenv)$) {\textbf{Robot}};
\draw[arr] (ain) -- (apol);
\draw[arr] (apol) -- node[tag, right] {control target} (actl);
\draw[arr] (actl) -- node[tag, right] {motor command} (arob);
\draw[arr] (arob) -- node[tag, right] {physical action} (aenv);
\draw[fb] (aenv.west) -- ++(-0.45,0) |- node[tag, pos=0.25, left, align=right] {observation\\ and feedback} (apol.west);

\draw[arr] (bin) -- (bs2);
\draw[arr] (bs2) -- node[tag, right] {tool call} (bs1);
\draw[arr] (bs1) -- node[tag, right] {control target} (bctl);
\draw[arr] (bctl) -- node[tag, right] {motor command} (brob);
\draw[arr] (brob) -- node[tag, right] {physical action} (benv);
\draw[fb] (benv.east) -- ++(0.45,0) |- node[tag, pos=0.25, right, align=left] {observation\\ and feedback} (bs2.east);
\end{tikzpicture}
\caption{\textbf{Two paradigms.} (a)~An action policy maps the task instruction and observation to action targets. It might include reasoning, subtask prediction, or in-context adaptation. (b)~S2 combines a foundation model with an embodied harness. Together, they interpret intent, orchestrate skills and tools, verify outcomes, and adapt subsequent decisions. S1 provides callable skills and tools. S0 performs body control. The S1 path is illustrative; S2 may also call S0 or external tools.}
\label{fig:paradigm}
\end{figure}
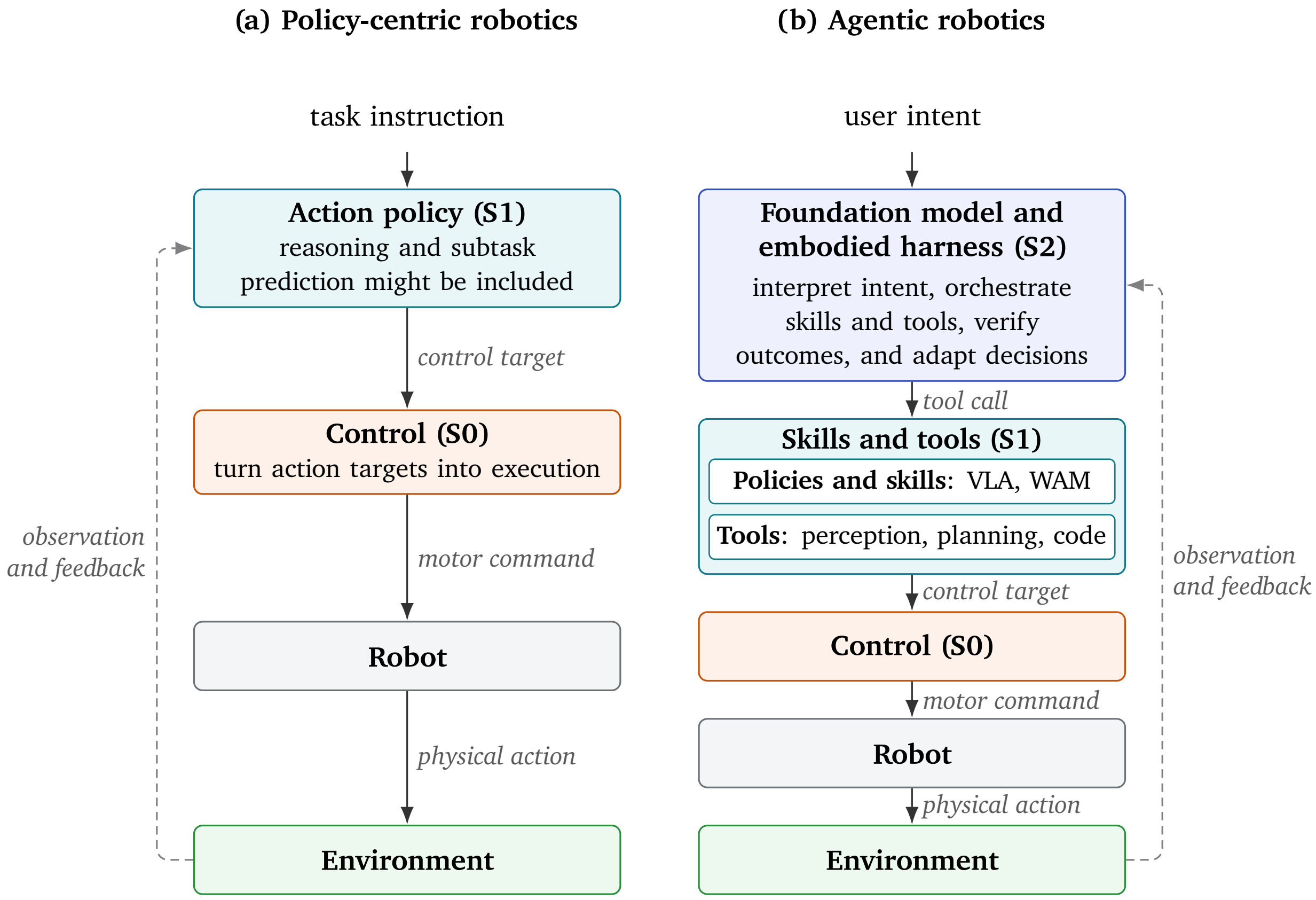

To address this generalization challenge, a major line of research focuses on scaling and diversifying training data for VLA and WAM policies \citep{egoscale,dyna2026dyna2}. These efforts draw on diverse data sources, including real-robot data \citep{pi2025pi05}, UMI demonstrations \citep{chi2024umi}, egocentric human data \citep{egoscale}, and multimodal web data \citep{brohan2023rt2}. The release of GPT-6 Astra \citep{openai2026gpt6astra,openai2026gpt6astrasystemcard} and its early applications to robotics offered a different perspective on this challenge. In an early evaluation, the model outperformed trained policies on many robotic benchmarks without task-specific fine-tuning \citep{zhang2026unexpected}. Subsequent work explored its use with learned policies and robotic tools \citep{galbot2026systematically,gao2026hexashell}. These results suggest that generalization can also be achieved by relying on a general-purpose model's knowledge and reasoning to organize and adapt behavior during execution. 
This perspective motivates a paradigm shift from \emph{policy-centric robotics} to \emph{agentic robotics}, by shifting the focus from learned action policies (Figure~\ref{fig:paradigm}a) to an agent that pursues goals by coordinating skills and tools, evaluating outcomes, and adapting its decisions (Figure~\ref{fig:paradigm}b).

Despite this progress, the community still lacks a shared understanding of agentic robotics or a reliable evaluation framework.
The scope of the paradigm, its relationship to existing forms of robot autonomy, and the organization of its constituent functions remain insufficiently clarified \citep{qin2026towards}.
Furthermore, differences in experimental conditions and system configurations also make it difficult to compare results across studies.
For example, RoboDojo evaluations of Astra vary in the task-knowledge sources they use and the action budgets they permit \citep{zhang2026unexpected,sensenova2026has}; separately, Harness VLA shows substantial performance gains from execution memory when the planner and VLA are held fixed \citep{zhang2026harnessvla}.
All of these results underscore the need to establish what constitutes the agentic system, which capabilities are being assessed, and under what conditions comparisons are meaningful.
These conceptual and methodological challenges call for a unified framework that clarifies the agentic workflow by specifying the functional roles of its components and how they interact.
Such a framework can provide a shared conceptual foundation for characterizing agentic robotics, developing valid and comparable evaluations at the model and system levels, and formulating a research roadmap.

To address this, we propose a framework of agentic robotics that distinguishes three functional roles (Figure~\ref{fig:architecture}). 
System~2 (S2) interprets intent, reasons about goals and constraints, orchestrates skills and tools, and verifies outcomes to guide subsequent decisions. System~1 (S1) provides skills and callable modules, while System~0 (S0) supplies the body interface and real-time control. 
Specifically, S2 is the decision-making core of agentic robotics, realized by a foundation model together with an \emph{embodied harness}.
The foundation model supplies knowledge, understanding, and reasoning, while the harness organizes observations, maintains plans and memory, exposes skills and tools, and supports execution and verification. 
Learned policies, perception layers, motion planning modules, and generated programs are included in S1. 
S2 may orchestrate these capabilities or send action targets directly to S0, while retaining responsibility for verifying outcomes and adapting its decisions. 
With these roles defined, we investigate how the foundation model and harness should work together under physical constraints, and how to evaluate their contributions to closed-loop system performance.

Building on this division of responsibilities, we first explain how S2, S1, and S0 work together (Section~\ref{sec:overview}). 
We then articulate S2 and examine its foundation model, its harness, and the physical constraints under which they operate (Section~\ref{sec:s2}).
Next, we develop an evaluation framework across nine dimensions that distinguishes model capabilities from whole system performance, and use it to assess the available evidence (Section~\ref{sec:evaluation}). 
For a closed-loop system, the framework emphasizes independent outcome verification, transparent reporting, and evaluation while the environment continues to evolve.
Furthermore, we examine which design principles carry over from digital agents to physical systems and which assumptions must be reconsidered in the physical world (Section~\ref{sec:digital-physical}).
These analyses inform a roadmap organized around four levels of delegable work, from executing primitives (L1), completing tasks (L2), coordinating workflows (L3), to fulfilling standing roles (L4) (Figure~\ref{fig:teaser}; Section~\ref{sec:road-ahead}). 
The roadmap connects this progression to four research directions, including continual improvement from physical experience. 
Finally, we conclude with several guidelines to verify and evaluate the progress of agentic robotics.

\section{Overview of Agentic Robotics}
\label{sec:overview}

We adopt a system-level view of agentic robotics: an agent interprets the user's intent, decides how to proceed, orchestrates skills and tools at run time, and verifies their physical effects against the intended goal. 
Feedback from verification guides subsequent decisions, enabling the system to adapt its actions, recover from failures, and seek clarification or assistance when needed. 
We first present a reference architecture that defines the responsibilities and interfaces of S2, S1, and S0. 
We then go through the agentic loop and examine three system-level properties: operation across multiple timescales, responsiveness during deliberation, and state persistence across tasks.

\subsection{Reference Architecture}
\label{sec:components}

\begin{figure}[!tp]
\centering
\begin{tikzpicture}[
  font=\footnotesize,
  execute at begin node={\hyphenpenalty=10000\relax},
  >={Latex[length=2mm]},
  comp/.style={draw=#1!75!black, fill=#1!8, rounded corners=2pt, align=left, inner sep=3pt,
    font=\scriptsize, line width=0.5pt},
  lab/.style={font=\scriptsize\itshape, black!70, inner sep=1pt},
]
\newcommand{\abox}[4]{\parbox[c][#2][t]{#1}{\raggedright\textbf{\footnotesize #3}\\ #4}}
\node[comp=WorldColor, text width=6.2cm, align=center, anchor=south] (user) at (9.1,1.45)
  {\textbf{\footnotesize User}\\ explicit or implicit intent, preference, interruption};
\draw[->, thick] (8.5,1.45) -- node[lab, left, xshift=-1.5pt] {intent} (8.5,0.9);
\draw[<-, thick] (9.7,1.45) -- node[lab, right, xshift=1.5pt] {question, report} (9.7,0.9);

\draw[draw=STwoColor!75!black, fill=STwoColor!3, rounded corners=4pt, line width=0.8pt] (3.3,0.9) rectangle (14.9,-5.98);
\node[anchor=north west, font=\small\bfseries, text=STwoColor!70!black] at (3.4,0.85) {System 2: foundation model and embodied harness};
\node[anchor=north west, lab] at (3.45,0.42) {on demand or event-driven: interpret intent, decide, orchestrate skills and tools, verify outcomes};
\draw[draw=STwoColor!60!black, fill=white, rounded corners=3pt] (3.5,0.0) rectangle (10.083,-5.78);
\node[anchor=north west, font=\footnotesize\bfseries] at (3.6,-0.05) {Embodied harness};
\node[comp=STwoColor, anchor=north west] (cm) at (3.65,-0.5)
  {\abox{2.856cm}{1.40cm}{Context manager}{observations, events; world, user, and body state; progress; history}};
\node[comp=STwoColor, anchor=north west] (ti) at (6.867,-0.5)
  {\abox{2.856cm}{1.40cm}{Tool interface}{S1 skills, S0 interfaces, code execution, search, external APIs}};
\node[comp=STwoColor, anchor=north west] (pl) at (3.65,-2.26)
  {\abox{2.856cm}{1.40cm}{Planner}{task decomposition, candidate plans, replanning}};
\node[comp=STwoColor, anchor=north west] (me) at (6.867,-2.26)
  {\abox{2.856cm}{1.40cm}{Memory}{short- and long-term; retrieval, writing, update, delete}};
\node[comp=EnvColor, anchor=north west] (ve) at (3.65,-4.02)
  {\abox{6.072cm}{1.40cm}{Verifier}{check each step and the task from evidence}};
\draw[draw=FMColor!60!black, fill=white, rounded corners=3pt] (11.383,0.0) rectangle (14.75,-5.78);
\node[anchor=north west, font=\footnotesize\bfseries] at (11.483,-0.05) {Foundation model};
\node[comp=FMColor, anchor=north west] (kn) at (11.533,-0.5)
  {\abox{2.856cm}{1.40cm}{Knowledge}{world, embodied, programming}};
\node[comp=FMColor, anchor=north west] (un) at (11.533,-2.26)
  {\abox{2.856cm}{1.40cm}{Understanding}{perception, space, physics, behavior, body kinematics}};
\node[comp=FMColor, anchor=north west] (re) at (11.533,-4.02)
  {\abox{2.856cm}{1.40cm}{Reasoning}{intent, planning, tool use, direct control, recovery, interruption}};
\draw[->, thick] (10.083,-1.305) -- node[lab, above] {context} (11.383,-1.305);
\draw[<-, thick] (10.083,-3.065) -- node[lab, above, align=center] {response} (11.383,-3.065);

\node[comp=SafetyColor, text width=2.05cm, align=flush left, anchor=north west, execute at begin node={\hyphenpenalty=10000\relax}] (evt) at (-0.45,-2.0)
  {\textbf{\footnotesize Real-time event monitor}\\ high-rate watch: user command, body and scene change, execution anomaly};
\draw[->, thick, SafetyColor!75!black] (evt.east |- 0,-2.85) -- node[lab, above] {interrupt} (3.3,-2.85);
\draw[->, thick] (user.west) -- node[lab, above] {command} (evt.north |- user.west) -- (evt.north);

\node[comp=SOneColor, text width=11.39cm, anchor=north west] (s1) at (3.3,-6.88)
  {\textbf{\footnotesize System 1: skills and callable modules}\\ perception, control policies (VLA, WAM, etc.), motion planning};
\node[comp=SZeroColor, text width=11.39cm, anchor=north west] (s0) at (3.3,-8.53)
  {\textbf{\footnotesize System 0: body interface and local real-time control}\\ whole-body tracking, inverse kinematics, joint control};
\node[comp=WorldColor, text width=11.39cm, anchor=north west] (body) at (3.3,-10.18)
  {\textbf{\footnotesize Robot body}\\ sensors (cameras, depth, IMU, force) and actuators (joints, grippers, base)};
\node[comp=EnvColor, text width=11.39cm, anchor=north west] (world) at (3.3,-11.83)
  {\textbf{\footnotesize Physical world}\\ objects, people, and space};

\draw[->, thick] (5.3,-5.98) -- node[lab, right] {decision, tool call} (5.3,-6.88);
\draw[<-, thick] (10.3,-5.98) -- node[lab, right] {execution result} (10.3,-6.88);
\draw[->, thick] (5.3,-7.70) -- node[lab, right] {control target} (5.3,-8.53);
\draw[<-, thick] (10.3,-7.70) -- node[lab, right] {measured state} (10.3,-8.53);
\draw[->, thick] (5.3,-9.35) -- node[lab, right] {actuator command} (5.3,-10.18);
\draw[<-, thick] (10.3,-9.35) -- node[lab, right] {sensor state} (10.3,-10.18);
\draw[->, thick] (5.3,-11.00) -- node[lab, right] {physical action} (5.3,-11.83);
\draw[<-, thick] (10.3,-11.00) -- node[lab, right] {world state} (10.3,-11.83);
\draw[->, densely dashed, thick, STwoColor!75!black] (14.9,-5.45) -- (15.2,-5.45) -- node[lab, sloped, below, STwoColor!75!black] {optional direct call} (15.2,-8.94) -- (s0.east |- 0,-8.94);

\draw[thick] (body.west) -- node[lab, above, pos=0.55] {body movement} ([xshift=-0.55cm]evt.south |- body.west);
\draw[->, thick] (world.west) -- node[lab, above, pos=0.55] {scene change} ([xshift=-0.55cm]evt.south |- world.west) -- ([xshift=-0.55cm]evt.south);
\draw[->, thick, SafetyColor!75!black, densely dashed] ([xshift=0.55cm]evt.south) -- ([xshift=0.55cm]evt.south |- 0,-8.94) -- node[lab, above, align=center, SafetyColor!75!black] {immediate\\ reaction} (3.3,-8.94);
\node[lab, SafetyColor!75!black, align=right, anchor=east] at ($([xshift=0.55cm]evt.south |- 0,-6.13)+(-0.05,0)$) {safety\\ rules};

\draw[->, line width=1.1pt, EnvColor!70!black] (body.east) -- (15.85,-10.60) -- node[lab, sloped, above, EnvColor!60!black, pos=0.5] {observation} (15.85,0.45) -- (14.9,0.45);
\end{tikzpicture}
\caption{\textbf{A reference framework for agentic robotics.} An Embodied S2 consists of a foundation model and an
embodied harness, whose five core parts manage context, tools, planning, memory, and
verification. S2 runs on demand or in response to events, whereas S1 and S0 execute at
their own rates. A real-time event monitor outside the three systems watches the user,
the robot body, and the physical world at a high rate. It interrupts S2 when an event may
change the task and, under predefined safety rules, triggers an immediate reaction in S0
without waiting for S2. S2 may call S0 directly through the tool interface;
S0 keeps its own real-time loop.}
\label{fig:architecture}
\end{figure}
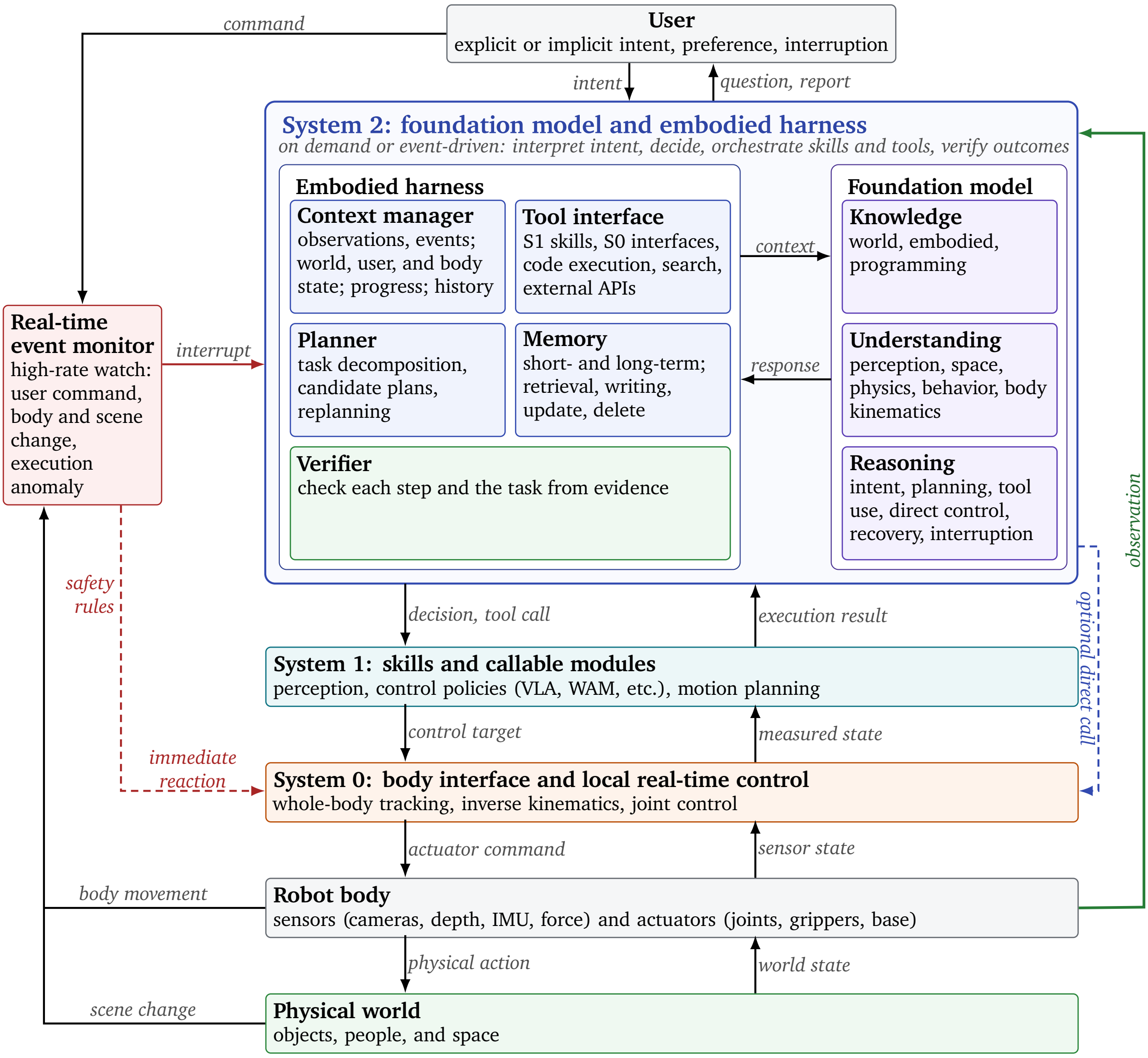

In general, an agentic robot requires three functional roles: S2 interprets intent, reasons about goals and constraints, orchestrates capabilities, and verifies outcomes; S1 provides skills and callable modules; and S0 provides the body interface and local real-time control.
Note that this decomposition is functional: one model may serve several roles, and one role may span several models or modules. 
Figure~\ref{fig:architecture} shows the three systems and their interfaces, together with the user, the robot body, the physical world, and a real-time event monitor. 
The systems typically run on different timescales, but timing alone does not distinguish them. We describe them from S2 downward, focusing on their responsibilities and interactions.

\paragraph{System~2: Intent, Reasoning, Orchestration, and Verification.}
S2 is the functional layer that interprets intent, reasons about goals and constraints to make decisions, orchestrates skills and tools accordingly, and verifies outcomes to adapt subsequent decisions toward the intended goals. 
It decides whether to act, gather information, ask for clarification, continue, retry, replan, or stop. When acting, it chooses a capability and specifies the call, which may invoke an S1 skill or a tool, or send an action target directly to S0.
It checks the returned evidence against the intended goal before deciding on the next subtask. 
In hierarchical systems such as Hi Robot \citep{shi2025hirobot} and Gemini Robotics \citep{gemini2025robotics,gemini2025robotics15,gemini2026robotics25}, a high-level vision-language model contributes to the orchestration: it interprets instructions and feedback before delegating the next subtask to an action model.

In our framework, S2 is realized by a foundation model and an embodied harness. 
The \emph{S2 foundation model} supplies knowledge, understanding, and reasoning, and proposes the next decision from the context it receives.
The \emph{embodied harness} supplies the runtime that assembles the model’s context and turns its decisions into constrained, observable, and verifiable execution.
Concretely, it has five core parts. 
A \emph{context manager} assembles current observations, task progress, interaction history, and the state of the world, the user, and the body. 
A \emph{tool interface} exposes S1 skills, permitted S0 calls, code execution, and other tools to S2.
A \emph{planner} maintains and revises task plans given the foundation model's proposals or dedicated planning routines. 
A \emph{memory} stores, retrieves, updates, and removes information beyond the current context. 
A \emph{verifier} judges from evidence whether a step achieved its subgoal and whether the task is complete. 
Existing systems implement some of these parts: Inner Monologue \citep{huang2022inner} feeds scene and skill-success signals back into planning; SCOPER \citep{wei2026scoper} updates an evidence memory and verifies progress to guide recovery. 

\paragraph{System~1: Skills and Callable Modules.}
S1 provides skills and callable modules that perform the functions S2 delegates, including action execution, perception, and motion planning. 
Specifically, \emph{action execution} modules map observations to actions within the scope of a subtask, such as VLA policies \citep{black2025pi0,pi2025pi05,nvidia2025gr00t}, world-action models (WAMs) \citep{dyna2026dyna21}, locomotion policies, and navigation policies.
\emph{Perception} modules extract task-relevant estimates from sensor data through detection \citep{minderer2022simple}, segmentation \citep{carion2025sam3}, pointing \citep{deitke2024molmo}, and pose estimation \citep{wen2024foundationpose}.
\emph{Motion planning} modules turn a goal into a feasible motion \citep{sundaralingam2023curobo,sundermeyer2021contact,kim2025pyroki,huang2024rekep}. 
A single call may combine several of these categories.
For example, a grasp call may segment the object, select a grasp function, plan a path, and execute it. 
Programs written by S2 likewise serve as S1 skills when they execute delegated behavior \citep{yang2026robotworld}.

\paragraph{System~0: Body Interface and Local Real-Time Control.}
S0 is the embodiment-specific layer that executes control targets in a real-time loop.
Examples are whole-body tracking and balance for a humanoid, gait tracking and stabilization for a quadruped, inverse kinematics and trajectory tracking for an arm, and chassis control for a mobile base.
For humanoids, whole-body tracking is increasingly handled by general-purpose controllers—often called behavior foundation models—that are pretrained at scale \citep{yuan2025bfmsurvey}, for example by learning to track large human-motion datasets \citep{luo2026sonic,liao2026supermimic,fu2026cosa05}.
S0 receives control targets from S1, or from S2 through a permitted action interface, and reports state estimates such as joint positions, contact, and pose.
It maintains stability and enforces the body's control limits without waiting for S2 to deliberate. 

\paragraph{Real-Time Event Monitor.}
A lightweight \emph{real-time event monitor} runs alongside the three systems.
At a high rate, it watches the user, the robot body, and the physical world for events that cannot wait for S2 to decide. 
Such events include a sudden command or interjection from the user, an unexpected body movement caused by a collision or fault, and a change in the scene such as a person entering the workspace.
In practice, the monitor may use a small streaming vision-language model (VLM), rule-based detectors, or both. 
It interrupts S2 when an event may change the task.
When an event demands an immediate reaction, it may apply its predefined safety rules and trigger a protective response in S0, such as stopping or slowing the current motion.

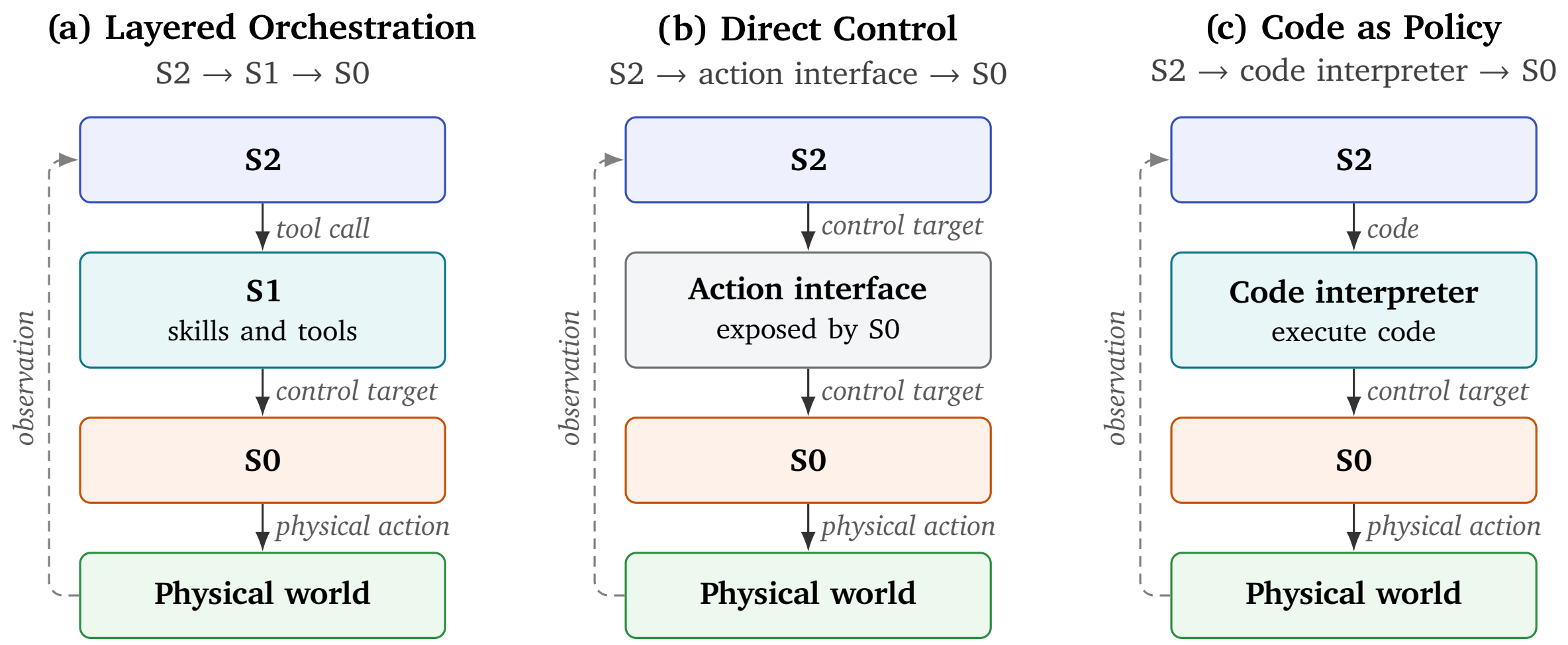
\begin{figure}[tp]
\centering
\begin{tikzpicture}[
  font=\footnotesize,
  execute at begin node={\hyphenpenalty=10000\relax},
  box/.style={draw=#1!80!black, fill=#1!9, rounded corners=3pt, line width=0.6pt, text width=3.4cm, align=center, inner sep=3.5pt, minimum height=0.85cm},
  head/.style={font=\small\bfseries, text width=4.8cm, align=center, inner sep=1pt},
  sub/.style={font=\footnotesize, text width=4.8cm, align=center, inner sep=1pt, black!75},
  arr/.style={-{Latex[length=2mm]}, semithick, black!80},
  fb/.style={-{Latex[length=2mm]}, semithick, black!50, densely dashed, rounded corners=4pt},
  tag/.style={font=\scriptsize\itshape, black!65},
]
\foreach \x/\h/\sub/\b/\bsub/\cb/\d/\lab in {
  0/{(a) Layered Orchestration}/{S2 $\rightarrow$ S1 $\rightarrow$ S0}/{S1}/{skills and tools}/SOneColor/{tool call}/{control target},
  5.45/{(b) Direct Control}/{S2 $\rightarrow$ action interface $\rightarrow$ S0}/{Action interface}/{exposed by S0}/WorldColor/{control target}/{control target},
  10.9/{(c) Code as Policy}/{S2 $\rightarrow$ code interpreter $\rightarrow$ S0}/{Code interpreter}/{execute code}/SOneColor/{code}/{control target}} {
  \node[head] (h) at (\x,0.05) {\h};
  \node[sub] at (\x,-0.38) {\sub};
  \node[box=STwoColor] (s2) at (\x,-1.25) {\textbf{S2}};
  \node[box=\cb, minimum height=1.15cm] (mid) at (\x,-2.75) {\textbf{\b}\\{\scriptsize \bsub}};
  \node[box=SZeroColor] (s0) at (\x,-4.25) {\textbf{S0}};
  \node[box=EnvColor] (w) at (\x,-5.6) {\textbf{Physical world}};
  \draw[arr] (s2) -- node[tag, right] {\d} (mid);
  \draw[arr] (mid) -- node[tag, right] {\lab} (s0);
  \draw[arr] (s0) -- node[tag, right] {physical action} (w);
  \draw[fb] (w.west) -- ++(-0.3,0) |- (s2.west);
  \node[tag, rotate=90, anchor=south] at ($(s2.west)!0.5!(w.west)+(-0.32,0)$) {observation};
}
\end{tikzpicture}
\caption{\textbf{Three typical execution paths.} One system may use all three paths within a single task. For simplicity, the robot body is omitted. (a)~Layered orchestration: S2 invokes S1 skills and tools, which produce control targets that S0 executes. (b)~Direct control: S2 controls S0 directly through an action interface; S0 still runs its own real-time control loop. (c)~Code as policy: S2 writes code that a runtime interpreter executes in the role of S1.}
\label{fig:exec-paths}
\end{figure}

In general, commands flow downward and evidence flows upward across the three systems. 
In the downward direction, we divide the execution of a decision into three typical paths by how S2 reaches the body, as shown in Figure~\ref{fig:exec-paths}.
In \emph{layered orchestration}, S2 invokes S1 skills and tools, which produce the actions that S0 executes. 
In \emph{direct control}, S2 bypasses S1 and controls S0 through an action interface, as several systems built on frontier models do \citep{zhang2026unexpected,yang2026robotworld}. 
Direct control does not place the foundation model inside the real-time control loop: S0 still runs its own local loop, and the action interface defines what S2 may send to it. 
In \emph{code as policy}, S2 writes code that plays the role of S1 at run time \citep{liang2022code}; the code may
execute several actions before S2 decides again.
Note that these paths are not alternative architectures: one system may use all three within a single task. 
In the upward direction, S0 reports the state of the body. 
Each call should return the observations available at that point and an execution status, allowing S2 to assess the outcome. 
Outside this call chain, the event monitor signals S2 and S0 without being invoked.

The three systems also differ in how directly they depend on the body. 
S0 must account for the robot’s dynamics and control interface, whereas S1 skills depend on the actions and observations that the body provides. 
An S2 strategy of looking in the drawer before the cabinet may transfer across bodies, but only if it remains appropriate for the new robot’s capabilities, access constraints, and observations.

\subsection{Agentic Loop: From Intent to Verified Outcome}
\label{sec:loop}

\begin{figure}[!tp]
\centering
\begin{tikzpicture}[
  font=\footnotesize,
  execute at begin node={\hyphenpenalty=10000\relax},
  >={Latex[length=2mm]},
  step/.style={draw=#1!75!black, fill=#1!8, rounded corners=3pt, line width=0.6pt,
    text width=3.75cm, align=center, inner sep=4pt, minimum height=1.4cm},
  lab/.style={font=\scriptsize\itshape, black!70, inner sep=1pt},
  seg/.style={fill=#1!70, rounded corners=1pt},
  segl/.style={font=\tiny\bfseries, text=white, inner sep=0pt},
]
\node[step=STwoColor] (intent) at (0,0) {\textbf{Step 1: Interpret intent}};
\node[step=STwoColor] (decide) at (5.35,0) {\textbf{Step 2: Decide}};
\node[step=STwoColor] (invoke) at (10.7,0) {\textbf{Step 3: Invoke}};
\node[step=SOneColor] (exec) at (10.7,-2.45) {\textbf{Step 4: Execute}};
\node[draw=SafetyColor!75!black, fill=SafetyColor!8, rounded corners=3pt, line width=0.6pt,
  text width=6.4cm, align=center, inner sep=3pt] (imodel) at (5.35,1.85)
  {\textbf{Real-time event monitor}};
\node[step=EnvColor] (verify) at (5.35,-2.45) {\textbf{Step 5: Verify}};
\node[step=STwoColor] (next) at (0,-2.45) {\textbf{Step 6: Continue, replan, ask, or report}};
\draw[->, thick] (intent) -- (decide);
\draw[->, thick] (decide) -- (invoke);
\draw[->, thick] (invoke) -- (exec);
\draw[->, thick] (exec) -- (verify);
\draw[->, thick] (verify) -- (next);
\draw[->, thick] (next.north) -- node[lab, left, pos=0.5] {not done} ++(0,0.62) -| ([xshift=-0.9cm]decide.south);
\draw[->, thick, SafetyColor!75!black, densely dashed] (imodel.south) -- node[lab, right, SafetyColor!75!black] {~interrupt} (decide.north);

\begin{scope}[yshift=-4.55cm]
  \def\xa{-0.3}\def\xb{12.7}
  \node[anchor=east, font=\scriptsize] at (\xa-0.1,0) {\textbf{\textcolor{STwoColor!80!black}{S2}} on demand};
  \node[anchor=east, font=\scriptsize] at (\xa-0.1,-0.45) {\textbf{\textcolor{SOneColor!80!black}{S1}} while a tool runs};
  \node[anchor=east, font=\scriptsize] at (\xa-0.1,-0.9) {\textbf{\textcolor{SZeroColor!80!black}{S0}} 50\,Hz--1\,kHz};
  \foreach \y in {0,-0.45,-0.9} \draw[black!20] (\xa,\y) -- (\xb,\y);
  \fill[seg=STwoColor] (-0.2,-0.14) rectangle (1.9,0.14);  \node[segl] at (0.85,0) {1--3};
  \fill[seg=EnvColor]  (5.0,-0.14) rectangle (6.0,0.14);   \node[segl] at (5.5,0) {5};
  \fill[seg=STwoColor] (6.0,-0.14) rectangle (7.3,0.14);   \node[segl] at (6.65,0) {6, 2--3};
  \fill[seg=STwoColor] (9.0,-0.14) rectangle (10.3,0.14);  \node[segl] at (9.65,0) {2--3};
  \fill[seg=SOneColor] (1.9,-0.59) rectangle (5.0,-0.31);   \node[segl] at (3.45,-0.45) {4};
  \begin{scope}
    \clip[rounded corners=1pt] (7.3,-0.59) rectangle (9.35,-0.31);
    \fill[SOneColor!70] (7.3,-0.59) rectangle (9.0,-0.31);
    \fill[SOneColor!12] (9.0,-0.59) rectangle (9.35,-0.31);
    \clip (9.0,-0.59) rectangle (9.35,-0.31);
    \foreach \x in {8.66,8.75,...,9.35} \draw[SOneColor!80!black, line width=0.45pt] (\x,-0.59) -- ++(0.28,0.28);
  \end{scope}
  \draw[SOneColor!80!black, line width=0.3pt, rounded corners=1pt] (9.0,-0.31) -- (9.35,-0.31) -- (9.35,-0.59) -- (9.0,-0.59);
  \node[segl] at (8.15,-0.45) {4};
  \fill[seg=SOneColor] (10.3,-0.59) rectangle (12.6,-0.31); \node[segl] at (11.45,-0.45) {4};
  \foreach \x in {-0.25,-0.13,...,12.65}{ \draw[SZeroColor!85!black, line width=0.5pt] (\x,-1.0) -- ++(0,0.2);}
  \draw[SafetyColor!80!black, densely dashed, thick] (9.0,0.32) -- (9.0,-1.05);
  \node[anchor=south, lab, SafetyColor!75!black] at (9.0,0.32) {event monitor interrupts: tool call canceled, S2 replans};
  \draw[->, black!55] (\xa,-1.3) -- (\xb+0.1,-1.3) node[below left, font=\scriptsize, inner sep=1pt] {Time};
\end{scope}
\end{tikzpicture}
\caption{\textbf{The agentic loop.} Top: one cycle comprises six steps. 
S2 interprets intent, reasons about goals and constraints to decide how to proceed, and invokes an S1 skill, an S0 interface, or an external tool as needed. 
The invoked capability executes under its contract.
S2 then observes and verifies the outcome by answering four questions (whether the request was accepted, whether execution terminated, whether the physical effect holds, and whether the user's intent is satisfied).
Based on the verification, it continues, recovers, asks, or reports. 
The real-time event monitor, which watches the user, the body, and the scene, can interrupt S2 at any point. 
Bottom: the same steps on a schematic time axis, labeled with their step numbers and colors. 
S2 deliberates on demand, S1 runs while a tool executes, and S0 runs continuously; 
after an interruption, S2 cancels the running tool call and replans while S0 keeps the body stable; the hatched segment marks the canceled execution.}
\label{fig:loop}
\end{figure}
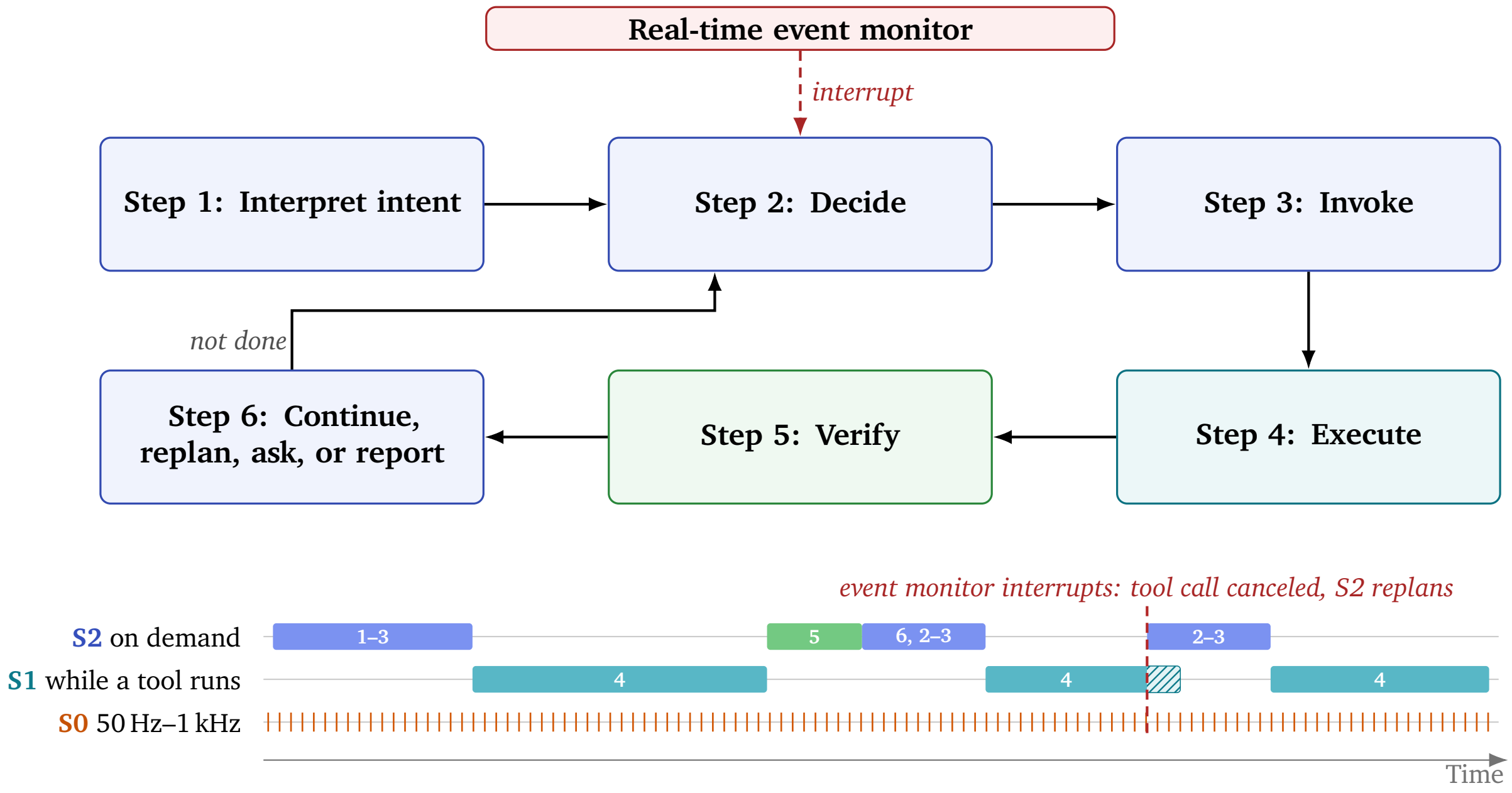

Consider the request, ``Find the red cup and bring it to the dining table.'' Given this request, S2 interprets it and plans four subtasks: locate, grasp, carry, and place the cup. It first invokes perception, and the harness records the cup's identity, location, and observation time. Then S2 invokes a grasp skill and checks the returned observation to make sure that the cup is held securely for transport. During transport, S1 and S0 maintain control while new observations inform S2's subsequent decisions. After placement, S2 verifies that the cup is supported by the table at the requested location. Each stage forwards state to the next, and the harness distinguishes a past event, ``the cup was grasped'', from a current state, ``the cup is still held''.

Suppose instead that the user interjects while the robot is grasping the red cup: ``Not the red cup, the blue one.'' The event monitor detects the interjection and interrupts S2. S2 recognizes that the revised goal conflicts with the current action and requests that the grasp be paused or canceled. A cancellation request alone is insufficient: before dispatching a conflicting action, the harness must confirm that execution has stopped and that the body is in a suitable holding state. S2 then refreshes its observations and replans to retrieve the blue cup.
Sometimes a skill reports success even though the action did not succeed. In this case, verification also catches unreported failures. For example, the grasping skill reports success on the blue cup, but the wrist camera shows that the cup is still on the shelf: the grasp was executed as commanded but missed. S2 attributes the miss to a grasp point that was occluded by the cup's handle. It therefore observes the cup from another side and retries with a different grasp rather than repeating the same call.

The example above traces six logical steps, which form a feedback cycle.
\begin{enumerate}[label=\textbf{Step~\arabic*:}, leftmargin=*, itemsep=1pt, topsep=3pt]
\item \textbf{Interpret Intent.} The user expresses intent, which may be explicit (``put the mugs in the dishwasher'') or implicit in preferences and context. S2 combines the request with the context assembled by the harness, such as a known preference that determines where a cup belongs. When the intent or the evidence is insufficient, a valid decision should be asking the user or observing again.
\item \textbf{Decide.} Given the current context and memory, S2 reasons about goals and constraints to decide how to proceed. It may select a subgoal or capability, specify a permitted control target, or supply a program.
\item \textbf{Invoke.} S2 calls an S1 skill, a permitted S0 interface, or an external tool, with arguments that respect the tool's contract, such as reference frame, units, object identifiers, and preconditions.
\item \textbf{Execute.} The invoked skill or tool carries out the call under its execution contract. Local control continues while S2 deliberates; a skill proceeds until it finishes, is paused or canceled, or a safety condition requires intervention. The event monitor continues to watch the user, the body, and the scene.
\item \textbf{Verify.} S2 distinguishes four questions when assessing the outcome: was the request \emph{accepted}, has the execution \emph{terminated}, is there evidence that the intended physical \emph{effect} holds, and does that effect satisfy the user's \emph{intent}? 
Termination may indicate completion, failure, or cancellation; it does not by itself establish success. 
A successful tool return is only intermediate evidence: a grasp can be executed exactly as commanded and still miss the object.
\item \textbf{Continue, Replan, Ask, or Report.} Completion is judged against the user's intent. If the task has not completed, S2 proceeds to the next subgoal or replans after a failure. If the task is complete, S2 reports; if the evidence is inconclusive, it observes again or asks the user.
\end{enumerate}

Across the whole system, three properties should hold.

\paragraph{Multiple Timescales.}
The components typically operate on different timescales. S0 typically runs locally at roughly 50\,Hz to 1\,kHz \citep{luo2026sonic,dyna2026dyna21,figure2026helix02}. 
S1 control skills run continuously while a subtask executes, at rates that depend on the platform and the task. 
Some commercial stacks report rates from about 5\,Hz to 200\,Hz \citep{dyna2026dyna21,figure2025helix,nvidia2025gr00t}. 
By contrast, perception and motion planning modules are invoked on demand and return results after their computation finishes. 
S2 does not run at a fixed rate: it could be invoked when a skill finishes or fails, when new observations arrive, when the event monitor reports an event, or when the user speaks. 
A frontier model in this role may take from several seconds to tens of seconds per decision \citep{xu2026rtsafe,galbot2026systematically}, although this latency is an implementation property, not a defining feature of S2. 

\paragraph{Continuous Responsiveness.}
While S2 is deliberating or a skill is executing, the physical world or the body state may change, and the user may revise an instruction. 
The event monitor should watch for these changes and remain responsive throughout deliberation and execution.
Current systems achieve this responsiveness in different ways. 
Gemini Robotics-ER~2 treats a new prompt received during generation as an interruption of model reasoning and of pending tool calls \citep{google2026erstreaming}. 
MIRA combines streaming speech understanding with interruptible motion \citep{lin2026mira}. 
ProAgent pairs a fast model that decides whether and when to intervene with a slower model that decides what to do \citep{yang2026proagent}. 
In our architecture, the safety rules reside in the event monitor, so a protective stop can be requested in time.

\paragraph{Persistent State.}
The harness should keep a persistent state. We call one task an \emph{episode} and a continuous period of use a \emph{session}, which may contain many episodes. Within an episode, the harness tracks progress, object identities, and previous attempts. Across episodes and sessions, it maintains what the robot has learned about the environment, the user's preferences, and the body state. The record of the physical world must be reconstructed from observations and could become stale without notice, as when an object is moved. Persistent memory should therefore record when and how each item was obtained and allow revision, rather than treat a stored observation as a permanent fact.

\section{Embodied System 2: Foundation of Agentic Robotics}
\label{sec:s2}

The previous section described how S2 interacts with S1 and S0.
In this section, we examine S2 itself.
We first define S2 and give a criterion for identifying it in a robotic system. 
We then examine how the foundation model and embodied harness contribute to its eight capabilities and how physical constraints shape its operation.

\subsection{Definition}
\label{sec:s2-definition}

We formally define Embodied System~2 as follows. 

\begin{tcolorbox}[enhanced, colback=STwoColor!4, colframe=STwoColor!60, boxrule=0.8pt,
  arc=4pt, left=8pt, right=8pt, top=6pt, bottom=6pt]
\textbf{Definition (Embodied System 2).} Embodied System~2 (Embodied S2) is the functional layer of an agentic robot that interprets intent, reasons about goals and constraints to make decisions, orchestrates skills and tools accordingly, and verifies outcomes to adapt subsequent decisions toward the intended goals.
\end{tcolorbox}

In this paper, Embodied S2 refers specifically to an implementation in which a foundation model and an embodied harness jointly fulfill that role.
Hereafter, we use Embodied S2 and S2 interchangeably and reserve \emph{S2 foundation model} for the model inside Embodied S2 alone.

The above definition has three notable features. 
First, it is anchored in \emph{intent} rather than in a fixed task specification: the objective is a world state that satisfies the user. Reaching it may require clarifying an ambiguous request, inferring an unstated preference, or accommodating a change of mind. 
Second, it rests on \emph{responsibility}: S2 makes high-level decisions, whereas S1 and S0 make action-level decisions within the scope that S2 delegates to them.
Third, it includes the \emph{harness}, because the state on which S2's decisions depend, namely that of the world, the user, and the body, is usually not supplied by the environment directly. This state must be assembled by the harness from partial, noisy, and aging observations.

\paragraph{Identifying S2.}
Labels such as System~2, embodied brain, planner, and orchestrator are applied inconsistently, so we identify S2 by functionality, not by name.
A component, or a set of components, \emph{acts as S2} if and only if it closes
the loop of the definition, which requires two kinds of decision:
\begin{enumerate}[leftmargin=*, itemsep=1pt, topsep=2pt, label=(\roman*)]
  \item \emph{decisions on how to proceed}: interpreting the intent and, given the goals and constraints, either choosing the next subgoal and the skill or tool that should achieve it, together with the arguments of the call, or choosing to continue, retry, replan, ask, or stop; and
  \item \emph{decisions on outcomes}: judging from evidence whether the previous step, or the task as a whole, achieved its goal, so that the next decision can adapt accordingly.
\end{enumerate}

Both kinds of decision must also be \emph{identifiable from outside}: either the
component's output interface exposes them, or a controlled replacement experiment reveals
them. 
A component that makes only one kind of decision implements part of S2 but does not act as S2 on its own. 
For example, a planner that chooses and sequences skills without checking their effects orchestrates but never verifies. 
By contrast, a success detector makes neither kind of decision: it scores an observation, but whether that score settles the outcome of a step or
of the task, and what follows from it, is decided elsewhere. 
State estimators in S0, perception modules in S1, world models, and the real-time event monitor likewise supply evidence without deciding what it means for the task: they inform S2 but do not act as S2. 
Because the criterion concerns function, it applies to any implementation, including classical planners and behavior trees.

\paragraph{Borderline Cases.}
In dual-system VLAs whose upper module emits only a latent representation, such as Helix \citep{figure2025helix} and GR00T~N1 \citep{nvidia2025gr00t}, high-level decisions cannot be read from the interface. 
We therefore do not classify the upper module as S2 unless replacement
experiments show otherwise. Conversely, a model that first emits a subtask and then the
actions to achieve it, such as $\pi_{0.5}$ \citep{pi2025pi05}, or G0.5 with its chain-of-thought stream enabled \citep{liu2026g05}, makes an
identifiable decision on how to proceed. Its subtask stage therefore implements part of
S2, even though it shares weights with S1. As the authors of Hi Robot note, ``the
separation of roles at the model level is not fundamental to this design'' \citep{shi2025hirobot}. 

Decision frequency is not part of the criterion.
A model trained for the S2 role may decide about once per second \citep{shi2025hirobot}, whereas a frontier model queried over a network may take tens of seconds per decision \citep{galbot2026systematically}.
Either acts as S2 if it makes both kinds of decision.
Speed determines whether an S2 is useful in a given setting, so we treat it as a physical constraint rather than as part of the definition.
Nor is the granularity of the outputs part of the criterion.
A component that emits end-effector targets or programs acts as S2 only if it also makes both kinds of decision, and an S2 that orchestrates skills remains S2 when it emits a pose directly (Figure~\ref{fig:exec-paths}).

\paragraph{Relation to Existing Definitions.}
Our definition builds on recent proposals that describe S2 as the embodied ``brain'' that selects skills and checks their effects \citep{luo2025robobench,wei2026scoper}, and on work that distinguishes a model from the runtime around it \citep{liang2026wams,huang2026harnesssurvey,wang2026thea}. 
Our definition differs from these works in two respects: it provides an explicit, testable criterion, and it counts the harness as part of Embodied S2.

\subsection{S2 Foundation Model}
\label{sec:s2-model}

The S2 foundation model turns the context assembled by the harness into decisions. It may be a general frontier model used without embodied training, an embodied foundation model post-trained for embodied tasks, or the reasoning stage of a larger model that also produces actions. 
In each case, it inherits broad knowledge and reasoning from large-scale pretraining. However, its role differs from that of a chatbot or coding agent, since its decisions are carried out by a body in the physical world.

\begin{figure}[tp]
\centering
\input{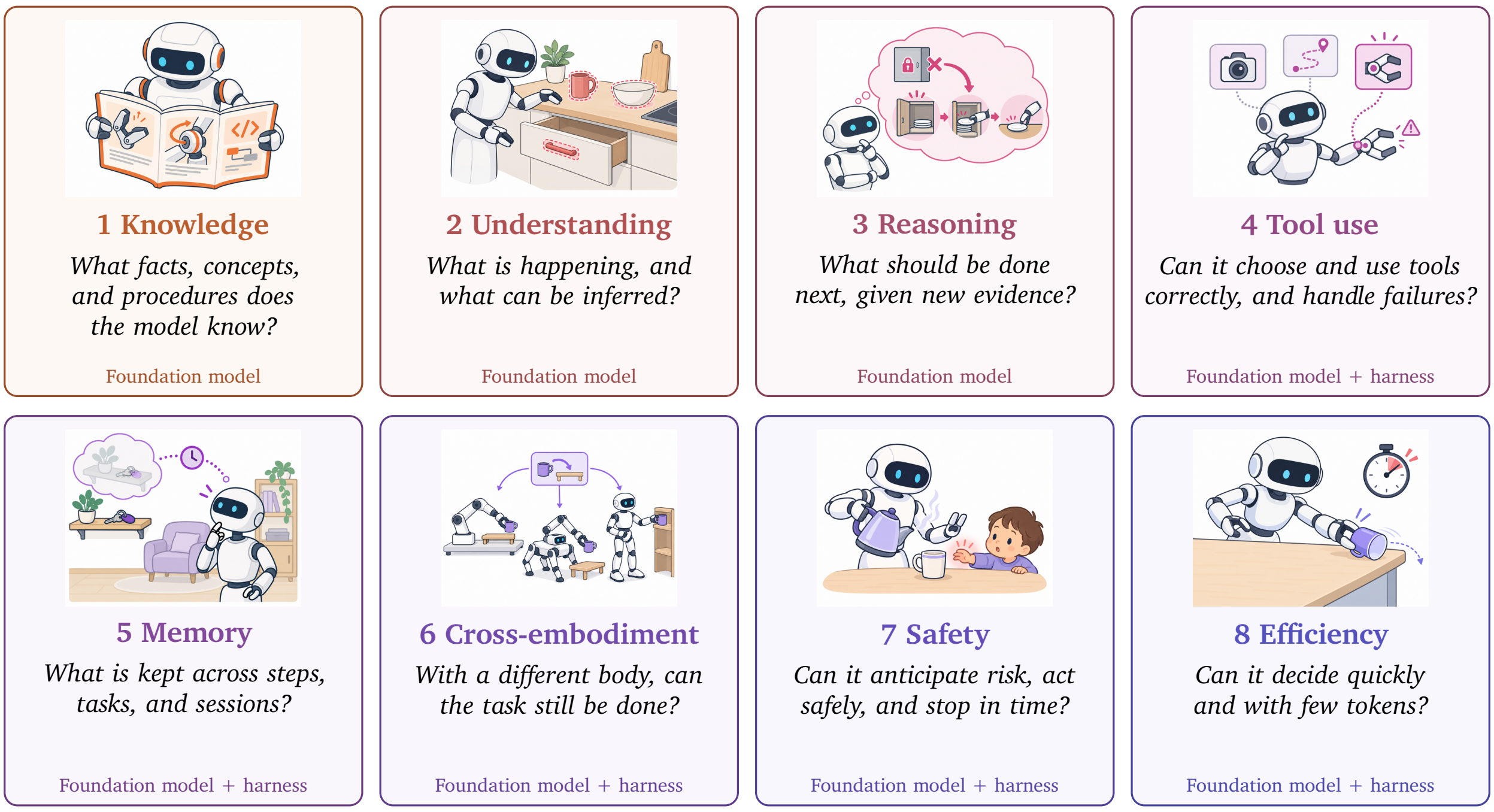}
\begin{tikzpicture}[font=\scriptsize, execute at begin node={\hyphenpenalty=10000\relax}]
\draw[evK!65!black, fill=evK!5, rounded corners=4pt, line width=0.6pt] (0,0) rectangle (3.95,-4.3);
\pic at (1.975,-1.125) {evillus={1}{evK}{\faBookOpen}{2.6cm}{1.95cm}};
\node[anchor=north, align=center, text width=3.65cm, inner sep=0pt] at (1.975,-2.27) {{\footnotesize\bfseries\color{evK!80!black}1~Knowledge}\\[3pt]{\itshape What facts, concepts, and procedures does the model know?}};
\node[anchor=south, align=center, text width=3.65cm, inner sep=0pt, font=\tiny, text=evK!70!black] at (1.975,-4.16) {Foundation model};
\draw[evU!65!black, fill=evU!5, rounded corners=4pt, line width=0.6pt] (4.15,0) rectangle (8.1,-4.3);
\pic at (6.125,-1.125) {evillus={2}{evU}{\faEye}{2.6cm}{1.95cm}};
\node[anchor=north, align=center, text width=3.65cm, inner sep=0pt] at (6.125,-2.27) {{\footnotesize\bfseries\color{evU!80!black}2~Understanding}\\[3pt]{\itshape What is happening, and what can be inferred?}};
\node[anchor=south, align=center, text width=3.65cm, inner sep=0pt, font=\tiny, text=evU!70!black] at (6.125,-4.16) {Foundation model};
\draw[evR!65!black, fill=evR!5, rounded corners=4pt, line width=0.6pt] (8.3,0) rectangle (12.25,-4.3);
\pic at (10.275,-1.125) {evillus={3}{evR}{\faProjectDiagram}{2.6cm}{1.95cm}};
\node[anchor=north, align=center, text width=3.65cm, inner sep=0pt] at (10.275,-2.27) {{\footnotesize\bfseries\color{evR!80!black}3~Reasoning}\\[3pt]{\itshape What should be done next, given new evidence?}};
\node[anchor=south, align=center, text width=3.65cm, inner sep=0pt, font=\tiny, text=evR!70!black] at (10.275,-4.16) {Foundation model};
\draw[evT!65!black, fill=evT!5, rounded corners=4pt, line width=0.6pt] (12.45,0) rectangle (16.4,-4.3);
\pic at (14.425,-1.125) {evillus={4}{evT}{\faTools}{2.6cm}{1.95cm}};
\node[anchor=north, align=center, text width=3.65cm, inner sep=0pt] at (14.425,-2.27) {{\footnotesize\bfseries\color{evT!80!black}4~Tool use}\\[3pt]{\itshape Can it choose and use tools correctly, and handle failures?}};
\node[anchor=south, align=center, text width=3.65cm, inner sep=0pt, font=\tiny, text=evT!70!black] at (14.425,-4.16) {Foundation model + harness};
\draw[evM!65!black, fill=evM!5, rounded corners=4pt, line width=0.6pt] (0,-4.52) rectangle (3.95,-8.82);
\pic at (1.975,-5.645) {evillus={5}{evM}{\faHistory}{2.6cm}{1.95cm}};
\node[anchor=north, align=center, text width=3.65cm, inner sep=0pt] at (1.975,-6.79) {{\footnotesize\bfseries\color{evM!80!black}5~Memory}\\[3pt]{\itshape What is kept across steps, tasks, and sessions?}};
\node[anchor=south, align=center, text width=3.65cm, inner sep=0pt, font=\tiny, text=evM!70!black] at (1.975,-8.68) {Foundation model + harness};
\draw[evX!65!black, fill=evX!5, rounded corners=4pt, line width=0.6pt] (4.15,-4.52) rectangle (8.1,-8.82);
\pic at (6.125,-5.645) {evillus={6}{evX}{\faRobot\,{\normalsize\faIcon{exchange-alt}}}{2.6cm}{1.95cm}};
\node[anchor=north, align=center, text width=3.65cm, inner sep=0pt] at (6.125,-6.79) {{\footnotesize\bfseries\color{evX!80!black}6~Cross-embodiment}\\[3pt]{\itshape With a different body, can the task still be done?}};
\node[anchor=south, align=center, text width=3.65cm, inner sep=0pt, font=\tiny, text=evX!70!black] at (6.125,-8.68) {Foundation model + harness};
\draw[evS!65!black, fill=evS!5, rounded corners=4pt, line width=0.6pt] (8.3,-4.52) rectangle (12.25,-8.82);
\pic at (10.275,-5.645) {evillus={7}{evS}{\faIcon{shield-alt}}{2.6cm}{1.95cm}};
\node[anchor=north, align=center, text width=3.65cm, inner sep=0pt] at (10.275,-6.79) {{\footnotesize\bfseries\color{evS!80!black}7~Safety}\\[3pt]{\itshape Can it anticipate risk, act safely, and stop in time?}};
\node[anchor=south, align=center, text width=3.65cm, inner sep=0pt, font=\tiny, text=evS!70!black] at (10.275,-8.68) {Foundation model + harness};
\draw[evE!65!black, fill=evE!5, rounded corners=4pt, line width=0.6pt] (12.45,-4.52) rectangle (16.4,-8.82);
\pic at (14.425,-5.645) {evillus={8}{evE}{\faIcon{stopwatch}}{2.6cm}{1.95cm}};
\node[anchor=north, align=center, text width=3.65cm, inner sep=0pt] at (14.425,-6.79) {{\footnotesize\bfseries\color{evE!80!black}8~Efficiency}\\[3pt]{\itshape Can it decide quickly and with few tokens?}};
\node[anchor=south, align=center, text width=3.65cm, inner sep=0pt, font=\tiny, text=evE!70!black] at (14.425,-8.68) {Foundation model + harness};
\end{tikzpicture}
\caption{\textbf{Core abilities of Embodied S2.} These abilities are necessary for Embodied S2 to perform successfully and reliably in the physical world; some reside in the foundation model, while others emerge from its integration with the embodied harness.}
\label{fig:eval-illustrated}
\end{figure}

Doing the work that the definition assigns to S2 in the physical world requires eight abilities (Figure~\ref{fig:eval-illustrated}). 
Knowledge, understanding, and reasoning reside mainly in the foundation model; tool use, memory, cross-embodiment, safety, and efficiency arise from the foundation model and the embodied harness together. 
For each ability, we describe what it is, why S2 needs it, and how the foundation model's share of it can be cultivated. 

\paragraph{Knowledge.}
Knowledge comprises the general facts, concepts, and procedures on which S2 draws. 
It matters because S2 must fill the gaps that a request leaves open, such as where a dish belongs or that a heavy box goes below a light one. 
\emph{World knowledge} covers objects, everyday relations, and culture. 
\emph{Embodied knowledge} covers materials, physical laws, tools and mechanisms, the preconditions and effects of actions, and the structure, sensing, and kinematics of robots. 
\emph{Programming knowledge} covers languages, algorithms, and software interfaces, which empowers an S2 model to compose skills with code \citep{liang2022code}. 
Knowledge requires broad coverage, which only pretraining-scale data provides. 
World and programming knowledge are therefore inherited largely from pretraining on web text, images, and code. 
Embodied knowledge is sparser in these sources but needs the same breadth. 
It should therefore also be acquired during pretraining, by adding embodied data to the corpus at scale rather than leaving it to the smaller and narrower data of post-training. 
Suitable sources include egocentric and instructional videos that show how people handle objects and what their actions achieve, and robot trajectories paired with language, which tie actions to their preconditions and effects. 
For example, Magma includes human videos and robot data in its pretraining \citep{yang2025magma}. 
Others are technical documentation, such as robot specifications, kinematic models, and controller interfaces, and simulated data on physical properties and dynamics, such as mass, friction, stability, and deformation. 

\paragraph{Understanding.}
Understanding is S2's grasp of the current situation: the state of the world, what the user wants, the state of the robot, and what cannot be inferred from the available observations. 
Understanding of the robot extends beyond its body to the capabilities beneath S2: which S1 skills exist, what their preconditions and failure modes are, and which S0 interfaces, reference frames, and limits apply. 
It differs from embodied knowledge: knowing how inverse kinematics works is knowledge, whereas judging whether \emph{this} robot, in its current state, can reach a given pose is understanding. 
Understanding matters because every decision rests on it: an understanding error could propagate into planning and execution. 
Understanding of the robot lets S2 choose the level at which to act: whether to orchestrate a skill, control S0 directly, ask, or decline. Understanding of progress and completion, in turn, makes verification possible.

\paragraph{Reasoning.}
Reasoning is how S2 decides what to do next and revises that decision as evidence arrives.
It encompasses inferring the intent behind an underspecified request, deciding when to ask for clarification, planning over long horizons, diagnosing failures, and handling interruptions.
Reasoning is therefore central to S2's role.
Two of these sub-abilities are particularly relevant to the execution paths in Figure~\ref{fig:exec-paths}: \emph{skill selection and orchestration}, which decides which existing skills and tools to use and in what order (the layered-orchestration path), and \emph{executable action inference}, which decides which action target to send directly when no suitable skill exists, without placing the model inside the real-time control loop (the direct-control path).
Over a long task, diagnosis and recovery further limit the compounding of small failure rates (Section~\ref{sec:loop}): S2 should change the conditions that caused a failure, such as the grasp point after a slip, rather than repeat the same call. 
Diagnosis can even drive learning without updating the model: by tracing each failure to its originating step, GPT-6 Astra learned manipulation skills on a real arm without demonstrations \citep{yu2026retrace}.

\paragraph{Tool Use.}
Tool use is the ability to choose, parameterize, and combine the capabilities available to S2, namely perception, planning, action, and code, and to handle their failures.
Each call involves three decisions: whether a call is needed at all; which arguments, in which reference frame and units, denote the intended physical target; and how far to rely on the result.
Tool use matters because S2 acts on the world through calls.
A sound decision executed with the wrong arguments still fails in execution, and sometimes a result can promise more than it delivers: a successful return from an inverse-kinematics solver does not guarantee that the robot can reach the solution without collision. 

\paragraph{Memory.}
Memory is the ability to retain what the current observation no longer shows, to use it in a later decision, and to replace it when it becomes stale.
A robot must remember objects and their attributes, locations, the order of completed steps, procedures demonstrated earlier, world state revealed by interaction, and the preferences of the people it serves.
Memory matters because physical work extends beyond what a camera sees at any moment and acting on an outdated memory can cause failures \citep{liang2026embodiedmemory}.
The model's role is to decide what is worth recording and whether a memory still applies to the present situation; storing and retrieving information beyond the context window is the harness's role.
The model learns these judgments from tasks in which earlier observations must be recalled and revised as the world changes.

\paragraph{Cross-Embodiment.}
Cross-embodiment is the ability to achieve a given objective when the body changes in appearance, morphology, end effector, action space, or dynamics.
The same goal may require different paths or motions on different platforms. For example, a wheeled base may need a longer route around obstacles that a legged robot can cross.
It matters because the value of a general S2 lies in serving many bodies without being rebuilt for each.
To serve a new body, the model must understand it, adapt the plans and action parameters to it, and reuse experience gained on other bodies without letting that experience interfere.

\paragraph{Safety.}
Safety is the ability to avoid harm along the entire action chain: recognizing a risk, predicting its consequences, choosing a cautious response, preventing harm before acting, monitoring during execution, recovering afterward, responding in time, and coordinating a stop.
Specifically, \emph{risk recognition} identifies hazards in the scene and in the request, such as hot liquid near a child's hand or a knife handed over blade first.
\emph{Consequence prediction} anticipates what an action would do in this particular scene. For example, pulling one box from a stack may topple the rest.
A \emph{prudent decision} chooses whether to proceed, proceed more cautiously, gather information, ask, or refuse, within the robot's validated capability limits, and prefers reversible or information-gathering actions under high uncertainty.
\emph{Prevention before action} checks a planned call against these predictions before it is dispatched.
During execution, S2 monitors whether the situation still matches its assumptions and replans when conditions change--for instance, when a person enters the workspace.
After a failure, \emph{safe recovery} first returns the robot and the scene to a safe state and only then retries.
A stop decided by S2 must be coordinated with S1 and S0 so that the running skill halts in a safe state.
Safety is thus a property of decisions in the action loop rather than of answers about hazards.
The model does not carry this responsibility alone. Hard limits on force and speed, protective stops, and permissions are enforced by the harness, S1, and S0 below the model, so that safety never rests on the model's judgment only.

\paragraph{Efficiency.}
Efficiency is the ability to make a good decision with low latency and with few tokens per decision. 
It matters because the physical world does not wait. 
A decision that arrives after the situation has changed is useless or even dangerous, and every token adds both latency and cost, which accumulate over hours of operation.
Two ways to improve decision efficiency are faster decoding and distilling capable models into smaller ones that run on or near the robot.

A growing family of embodied foundation models, often called embodied reasoning models, develops the abilities described above by post-training a general vision-language backbone on embodied data.
Examples include Gemini Robotics-ER \citep{gemini2025robotics,gemini2025robotics15}, Cosmos-Reason1 \citep{nvidia2025cosmosreason1}, Embodied-R1 and R1.5 \citep{yuan2025embodiedr1,yuan2026embodiedr15}, RynnBrain \citep{dang2026rynnbrain,li2026rynnbrain11}, MiMo-Embodied \citep{hao2025mimoembodied}, HY-Embodied \citep{tencent2026hyembodied}, Vesta \citep{bjorck2026vesta}, ACE-Brain \citep{ace2026acebrain05}, and PhysBrain \citep{deepcybo2026physbrain15}. 
Whatever model is chosen, it decides from the context it is given and acts through the interfaces it is offered.
Both are provided by the embodied harness, which we examine next.

\subsection{Embodied Harness}
\label{sec:s2-harness}

Section~\ref{sec:components} introduced the embodied harness through its five core parts; here we define its boundary. 
The harness comprises the components outside the foundation model that assemble the context for S2 and execute the decisions from S2.
A recent survey similarly describes the harness as the ``runtime control plane'' that grounds observations, coordinates executable capabilities, monitors outcomes, and manages intervention when execution diverges from the plan \citep{huang2026harnesssurvey}.
Table~\ref{tab:harness} lists the five core components and three augmentations, together with the problem that each solves.

\begin{table}[!tbp]
\centering\small
\caption{\textbf{Components of an embodied harness.} The five core components appear in
Figure~\ref{fig:architecture}; the three augmentations build on them.}
\label{tab:harness}
\begin{tabular}{@{}L{0.27\linewidth}L{0.65\linewidth}@{}}
\toprule Component & Problem addressed \\ \midrule
\rowcolor{STwoColor!7}\multicolumn{2}{@{}l}{\textbf{Core components}} \\ \midrule
Context manager & Context from the world, the user, and the body arrives continuously and
in large volume; it must be summarized and well managed. \\ \addlinespace
Tool interface & Skills and tools must be exposed to the model at a suitable level of
abstraction. \\ \addlinespace
Planner & A long task cannot be planned once and executed blindly. \\ \addlinespace
Memory & State must persist beyond the context window, and be updated. \\ \addlinespace
Verifier & A completed call does not show that a step achieved its goal. \\ \midrule
\rowcolor{STwoColor!7}\multicolumn{2}{@{}l}{\textbf{Augmentations}} \\ \midrule
Safety gate & Hazardous instructions and unsafe actions must be stopped before execution. \\ \addlinespace
Budget control & Large models are slow and expensive. \\ \addlinespace
Self-improvement & The system does not improve with use. \\
\bottomrule
\end{tabular}
\end{table}

\paragraph{Context: World, User, and Body.}
A context manager assembles the model's current context, turning a heterogeneous, continuously changing physical situation into a bounded and structured input for the model.
This context should cover at least three kinds of state. \emph{World context} records objects, places, and their relations, together with when and how each fact was observed. \emph{User context} records preferences, habits, and commitments; it changes at a different timescale and raises privacy questions that world context does not. \emph{Body context} records the robot's geometry, kinematics, calibration, and current condition, together with descriptions of the S1 skills and S0 interfaces available to it. The model's understanding of the situation rests on these records. When the body changes, an explicit description of the new body, with its frames, kinematics, sensors, tools, and validated limits, lets S2 plan for a robot that it has not used before.

\paragraph{Memory beyond the Context Window.}
The entire interaction history of an agentic robot cannot simply reside in the model's context window due to limited capacity.
For instance, a 128K-token context can hold only roughly three minutes of dense multimodal memory \citep{gopalakrishnan2026one}.
Within an episode, long context can nonetheless substitute for engineered memory, at a high cost in latency: a frozen GPT-6 Astra that kept an entire episode in context succeeded in 92.1\% of the 800 test episodes of RoboMME \citep{dai2026robomme} in a single run, averaging 18.3\,s per model call with the simulator paused during each call \citep{jedlicki2026memory}.
This result shows that in-context memory can be viable within an episode, but its latency in this setting makes it unattractive as the sole memory mechanism for time-sensitive tasks.
Across sessions, therefore, a memory module must store, retrieve, update, and invalidate information beyond the context window.
Several harnesses make this module explicit: Thea maintains a persistent scene graph \citep{wang2026thea}, HarnessVLN combines hierarchical event memory with a persistent spatiotemporal graph \citep{chen2026harnessvln}, and KnowBody maintains a revisable model of action-relevant body relations \citep{lou2026knowbody}.

\paragraph{Execution Contracts.}
Since physical actions have side effects, the harness must know precisely what each call promises. Its tool interface exposes every tool with a schema, units, reference frames, and failure semantics, so that the model's choices of tool and arguments refer to well-defined physical quantities. Recent systems point to four contracts \citep{wang2026regenharness,deng2026dynaharness,huang2026harnesssurvey}.
An \emph{observation snapshot} records when, where, and in which reference frame each observation was captured, and which calibration was in force.
A \emph{capability contract} specifies a skill's inputs and units, preconditions, expected effects, resources, time limits, behavior on cancellation, and the state it leaves for the next skill.
An \emph{invocation record} links each call to its task, subgoal, attempt, inputs, and final status, so that retries and late results can be traced.
An \emph{evidence bundle} records the observations, controller states, and verifier outputs on which a judgment of success was based.
Together, the contracts define a minimal lifecycle for every physical call: \emph{propose} $\rightarrow$ \emph{admit} $\rightarrow$ \emph{dispatch} $\rightarrow$ \emph{execute} $\rightarrow$ \emph{confirm the terminal state} $\rightarrow$ \emph{verify the effect} $\rightarrow$ \emph{commit progress}. The lifecycle has explicit branches for rejection, failure, cancellation, and unknown outcomes.
The branch for unknown outcomes is necessary since a timeout is not evidence that an action did not run. Before retrying, the harness should query the execution handle and observe the scene again; repeating a command that has in fact completed can execute a physical action twice.
Cancellation passes through three distinct states as well: requested, accepted, and completed; the return of a cancel call does not show that the robot has stopped.
Two systems illustrate this discipline: RegenHarness separates a model's proposal from a controller's termination and verified completion \citep{wang2026regenharness}, and DynaHarness records evidence under a shared physical execution contract \citep{deng2026dynaharness}.
Skill boundaries require the same care, because a grasp can succeed in a way that obstructs the subsequent insertion. 
Handoffs can be made more reliable in two ways: by specifying explicit entry and exit conditions between skills \citep{xu2026baton}, and by restricting handoffs to configurations that the next policy is known to support \citep{huang2026roboharness}.

\paragraph{Safety outside the Model.}
Safety checks are required before execution. 
A dedicated safety gate is one possible implementation; the checks may instead be integrated into the tool interface or enforced by an independent execution layer.
This check is the \emph{admit} step of the lifecycle above. 
Hard limits on force and speed are enforced below the harness, by S1 and S0, and the real-time event monitor can trigger a protective stop in S0 according to predefined safety rules. 
A stop thus takes effect even if the model or the harness fails.

\paragraph{Three Loops.}
A harness supports three feedback loops.
The \emph{sensorimotor loop} determines how actions respond to physical state; it resides in S1 and S0.
The \emph{planning loop} determines how the next subgoal is chosen after each result; it is the main loop of S2.
The \emph{evolution loop} determines which capabilities the next run will use, such as skills, prompts, context rules, tools, or model weights revised from experience.
A frozen model can still improve through the evolution loop, and a fixed program can still react through the sensorimotor loop.
Conflating the three therefore obscures where a reaction or an improvement originates.
Because the harness consists of code and data, S2 can also modify it: several recent self-improving robot agents revise tools, skills, context rules, or recovery policies rather than model weights \citep{tian2026physevo,deng2026dynaharness,lee2026emho,wang2026regenharness,lee2026robotuse}.
Such changes require the same discipline as model training, namely held-out tasks, regression checks, and versioned rollback.

\paragraph{Harness Engineering.}
Many hard problems reside in the harness: multi-rate scheduling, cancellation that actually stops motion, consistent state across asynchronous components, and accountable records of what the robot did and why.
Existing robotics infrastructure provides many building blocks for these problems, including action protocols with feedback and cancellation, middleware for real-time data flow \citep{macenski2022ros2}, and behavior trees with bounded recovery \citep{colledanchise2018behavior}.
Runtime governance layers add policy checks, capability admission, monitoring, rollback, and human override around the agent \citep{qin2026governance,liu2026phyagentos}.
The harness also carries the system's share of efficiency: it calls the model only when a decision is needed, compiles stable sub-processes into programs that run without the model, and delegates fast loops to S1 and S0.
What remains is to connect these mechanisms to a model and to measure what each mechanism contributes.

Existing measurements already show that the harness substantially influences overall performance. For example, interfaces that let the model check and retry locally outperformed plain tool calling while using fewer model calls \citep{si2026pyrualean}. 
Verification can be decisive: in one ablation, with the model fixed, removing HumanCLAW's verifier nearly eliminated navigation success \citep{siyao2026humanclaw}. The best level of abstraction for the interface also depends on the model: a skill-only interface raised GPT-5.6-Sol from 11\% to 63\% but lowered GPT-6 Astra from 92\% to 61\%; letting the model decide when to use skills recovered most of the loss \citep{han2026embodiedskillbench}.

\subsection{Constraints Imposed by the Physical World}
\label{sec:s2-requirements}

Unlike digital environments, the physical world involves embodiment, continuous dynamics, and irreversible consequences (Section~\ref{sec:digital-physical}). Observations are partial and noisy, actions are delayed and constrained by actuator limits, and contact-rich interactions are governed by friction, compliance, and wear that are difficult to model exactly.
Whatever model fills the S2 role, the physical world constrains how it may decide:
decisions face deadlines, can be interrupted at any moment, must fit the robot's compute
and cost budgets, and rest on aging, partial observations. Answer accuracy captures none
of these constraints, which the following paragraphs discuss in turn.

\begin{figure}[tb]
\centering
\begin{tikzpicture}[
  font=\scriptsize,
  execute at begin node={\hyphenpenalty=10000\relax},
  xscale=1.5, yscale=1.15,
]
\fill[red!7] (0,1.5) rectangle (5.0,4.15);
\draw[->, thick] (0,0) -- (5.2,0);
\node[anchor=north east, inner sep=1.5pt] at (5.2,-0.05) {Capability};
\draw[->, thick] (0,0) -- (0,4.3);
\node[anchor=south west, inner sep=1.5pt] at (0.05,4.25) {Reaction time};
\draw[dashed, red!70!black, thick] (0,1.5) -- (5.0,1.5);
\node[anchor=south west, red!70!black, inner sep=1.5pt] at (0.05,1.53) {Physical Deadline};
\node[anchor=north west, red!60!black, font=\scriptsize\itshape, inner sep=1.5pt] at (0.08,4.08) {Failure or danger from acting too late};
\draw[very thick, SOneColor!80!black] plot[smooth] coordinates {(0.25,0.28) (1.4,0.6) (2.45,1.16) (3.35,1.92) (4.15,2.76) (4.75,3.56)};
\node[SOneColor!80!black, align=right, anchor=east] at (3.3,2.55) {Frontier Models:\\ Monotonic Scaling};
\draw[very thick, STwoColor!85!black] plot[smooth] coordinates {(0.25,0.2) (1.3,0.46) (2.3,0.82) (3.1,1.1) (3.8,1.28) (4.9,1.35)};
\node[STwoColor!80!black, align=left, anchor=north west] at (2.85,0.95) {Embodied S2:\\ Capped Scaling};
\end{tikzpicture}
\caption{\textbf{Capped scaling.} Frontier models scale monotonically: greater capability comes with longer latency. Embodied S2 must instead become more capable \emph{within} a fixed decision deadline. A decision that arrives after the deadline is usually too late, even if it is correct: the situation has already changed, so acting on it could fail the task or create a hazard.
Note that the curves are schematic: they illustrate the qualitative relationship, not measured data.
}
\label{fig:capped-scaling}
\end{figure}
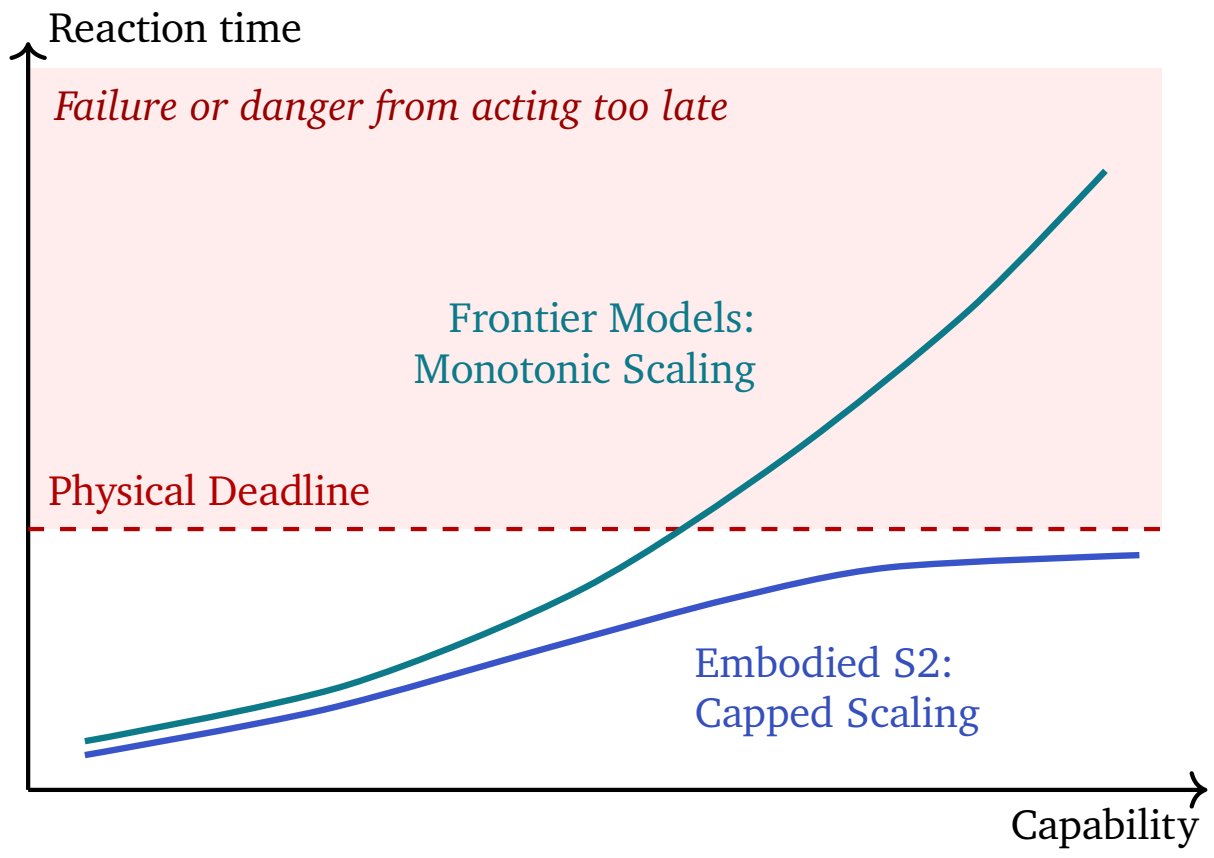

\paragraph{Capped Scaling.}
For frontier models, capability has grown with the computation: more reasoning tokens generally yield better answers at the cost of longer latency \citep{openai2024learning}.
We call this regime \emph{monotonic scaling}.
A robot cannot follow it since the physical world imposes deadlines.
A decision is useful primarily when it arrives before the situation that it addresses has changed, for instance before an object rolls off the table or a grasp slips.
Embodied S2 therefore requires a different regime, which we call \emph{capped scaling}: becoming more capable \emph{within} a fixed decision deadline (Figure~\ref{fig:capped-scaling}).
What matters is not how long the model deliberates but how well the system decides within the time available.
Humans meet physical deadlines through layered control \citep{todorov2004optimality}: grip-force corrections begin about 74\,ms after an object starts to slip \citep{johansson1987signals}, and the reflex responses that support recovery from a trip begin 60 to 140\,ms after it \citep{eng1994strategies}.
Deliberation handles only what cannot be handled faster.
An S2 should likewise match its reasoning effort to the time available and delegate fast loops to S1 and S0.
For the model, capped scaling means gaining capability within a fixed budget of tokens and time rather than by generating more tokens.

In terms of the S2 foundation model, capped scaling calls for \emph{concise reasoning}.
Much of the reasoning that people perform when acting physically is never verbalized,
such as where to put a heavy box when packing or how to approach a fragile glass.
We believe that such reasoning is under-represented in the text and chain-of-thought data
on which frontier models are trained, and that it does not require thousands of tokens.
An S2 should therefore reason briefly and in grounded terms, using points, regions,
relations, and short plans rather than long verbal deliberation. Concise reasoning helps
meet deadlines; whether it also improves correctness on physical tasks remains an open
question.

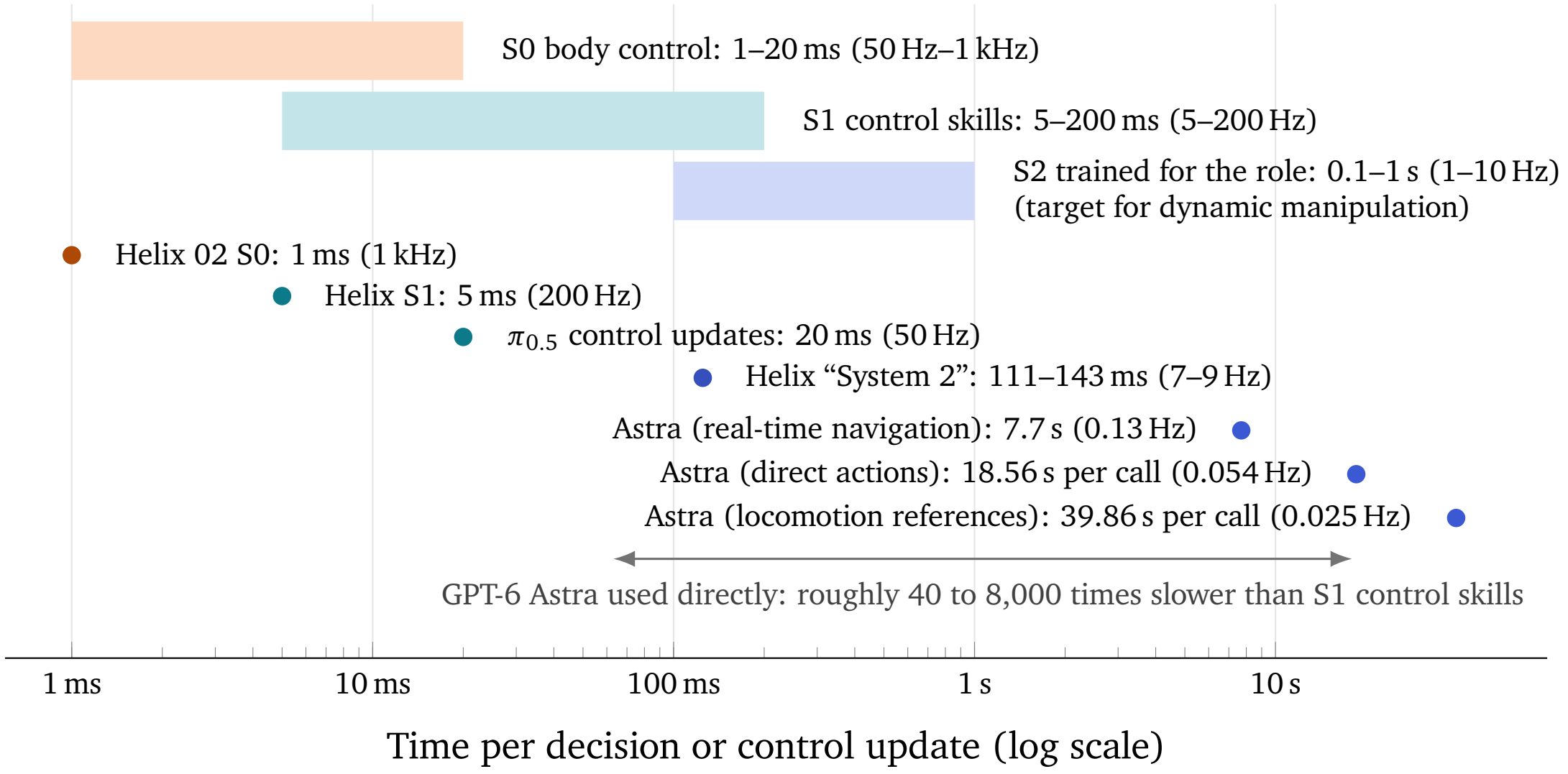
\begin{figure}[t]
\centering
\begin{tikzpicture}[font=\small]
\begin{axis}[
  xmode=log,
  width=0.95\linewidth, height=7.6cm,
  xmin=0.0006, xmax=80,
  ymin=-2.3, ymax=8.9,
  ytick=\empty,
  axis y line=none,
  axis x line*=bottom,
  xlabel={Time per decision or control update (log scale)},
  xlabel style={font=\small, yshift=-2pt},
  x tick label style={font=\scriptsize},
  xtick={0.001,0.01,0.1,1,10},
  xticklabels={1\,ms,10\,ms,100\,ms,1\,s,10\,s},
  xmajorgrids, grid style={black!10},
  clip=false,
]
\addplot[fill=SZeroColor!25, draw=none]
  coordinates {(0.001,7.6) (0.02,7.6) (0.02,8.6) (0.001,8.6)}
  \closedcycle;
\node[anchor=west, font=\scriptsize] at (axis cs:0.024,8.1)
  {S0 body control: 1--20\,ms (50\,Hz--1\,kHz)};

\addplot[fill=SOneColor!25, draw=none]
  coordinates {(0.005,6.4) (0.2,6.4) (0.2,7.4) (0.005,7.4)}
  \closedcycle;
\node[anchor=west, font=\scriptsize] at (axis cs:0.24,6.9)
  {S1 control skills: 5--200\,ms (5--200\,Hz)};

\addplot[fill=STwoColor!25, draw=none]
  coordinates {(0.1,5.2) (1,5.2) (1,6.2) (0.1,6.2)}
  \closedcycle;
\node[anchor=west, font=\scriptsize, align=left] at (axis cs:1.2,5.7)
  {S2 trained for the role: 0.1--1\,s (1--10\,Hz)\\
   (target for dynamic manipulation)};

\addplot[only marks, mark=*, mark size=2.2pt, color=SZeroColor!70!black]
  coordinates {(0.001,4.6)};
\node[anchor=west, font=\scriptsize] at (axis cs:0.00125,4.6)
  {Helix~02 S0: 1\,ms (1\,kHz)};

\addplot[only marks, mark=*, mark size=2.2pt, color=SOneColor!80!black]
  coordinates {(0.005,3.9)};
\node[anchor=west, font=\scriptsize] at (axis cs:0.0062,3.9)
  {Helix S1: 5\,ms (200\,Hz)};

\addplot[only marks, mark=*, mark size=2.2pt, color=SOneColor!80!black]
  coordinates {(0.020,3.2)};
\node[anchor=west, font=\scriptsize] at (axis cs:0.025,3.2)
  {$\pi_{0.5}$ control updates: 20\,ms (50\,Hz)};

\addplot[only marks, mark=*, mark size=2.2pt, color=STwoColor!80!black]
  coordinates {(0.125,2.5)};
\node[anchor=west, font=\scriptsize] at (axis cs:0.155,2.5)
  {Helix ``System~2'': 111--143\,ms (7--9\,Hz)};

\addplot[only marks, mark=*, mark size=2.2pt, color=HarnessColor!90!black]
  coordinates {(7.7,1.6)};
\node[anchor=east, font=\scriptsize] at (axis cs:6.2,1.6)
  {Astra (real-time navigation): 7.7\,s (0.13\,Hz)};

\addplot[only marks, mark=*, mark size=2.2pt, color=HarnessColor!90!black]
  coordinates {(18.56,0.85)};
\node[anchor=east, font=\scriptsize] at (axis cs:15,0.85)
  {Astra (direct actions): 18.56\,s per call (0.054\,Hz)};

\addplot[only marks, mark=*, mark size=2.2pt, color=HarnessColor!90!black]
  coordinates {(39.86,0.1)};
\node[anchor=east, font=\scriptsize] at (axis cs:32,0.1)
  {Astra (locomotion references): 39.86\,s per call (0.025\,Hz)};

\draw[{Latex[length=2mm]}-{Latex[length=2mm]}, black!55, line width=0.6pt]
  (axis cs:0.063,-0.6) -- (axis cs:18,-0.6)
  node[midway, below=2pt, font=\scriptsize, text=black!75]
  {GPT-6 Astra used directly: roughly 40 to 8,000 times slower than S1 control skills};
\end{axis}
\end{tikzpicture}
\caption{\textbf{How long each layer takes per decision or control update.}
All values are given as times, with the equivalent rate in parentheses (time $= 1/\text{rate}$).
The S0 and S1 bands are indicative ranges.
The S1 band covers control skills only, since the perception and planning modules of S1 are invoked on demand and have no fixed rate.
S2 is event-driven rather than periodic, so we plot time per decision; the S2 band is a target for dynamic manipulation, not a requirement for every task.
Helix's ``System~2'' is a vendor label: its output is a latent representation, so our criterion does not classify it as S2 (Section~\ref{sec:s2-definition}); we include it as the fastest upper-level module plotted.
The Astra measurements come from different tasks and settings and are not directly comparable with one another.
}
\label{fig:latency}
\end{figure}

Figure~\ref{fig:latency} places current systems on a common scale to show how they
compare with such deadlines. S0 loops run at about 50\,Hz to 1\,kHz, that is, every 1 to
20\,ms \citep{luo2026sonic,figure2026helix02}. The control skills of S1 run at about 5 to
200\,Hz, depending on the platform. The Dyna-2 policy updates its whole-body targets at about
5\,Hz \citep{dyna2026dyna21}, $\pi_{0.5}$ produces actions at 50\,Hz
\citep{pi2025pi05}, and Helix's S1 runs at 200\,Hz \citep{figure2025helix,figure2026helix02}. For
$\pi_{0.5}$, a median inference time of about 47\,ms has been reported (as quoted in
\citealp{yang2026astranomy}). The perception and motion planning modules of S1 have no
fixed rate, because S2 invokes them on demand. Used directly as the S2 foundation model,
GPT-6 Astra took roughly 40 to 8,000 times as long per decision as S1 control skills take
per update (Figure~\ref{fig:latency}). When issuing actions directly, it took 18.56\,s
per inference call \citep{yang2026astranomy}; when generating locomotion references, it
took 39.86\,s per call on average, and a 30-second walk required 250 calls with a paused
simulator \citep{galbot2026systematically}. When the world does not
pause, latency becomes a safety variable. RT-Safe \citep{xu2026rtsafe}, a simulated urban
navigation benchmark, keeps running while agents deliberate. On its hardest routes, Astra
at its default reasoning effort took 7.7\,s per decision, and models with longer response
latency collided more often.

\paragraph{Interactivity.}
Language models typically complete a response before reading the next input.
However, an S2 must be interactive: when the user says ``stop'' or the scene changes, it must abandon or revise its current line of reasoning.
This requires a model that accepts new input during generation and resumes from partial state, and a harness that cancels in-flight tool calls.
At least one commercial interface already supports such interruption: the streaming preview endpoint of Gemini Robotics-ER~2 cancels in-flight reasoning and tool calls when the user interjects \citep{google2026erstreaming}.
Real-time chunking addresses a related problem in S1 by generating the next action chunk while the current one executes \citep{black2025rtc}.

\paragraph{Deployability and Cost.}
Robots often operate with limited connectivity, and a single task may involve hundreds of decisions.
Therefore S2 must run within the robot's compute and power budget or tolerate network latency and outages.
The cost per task should be acceptable as well.
Several open embodied foundation models intended for the S2 role have 2B to 30B parameters, and some include variants designed for edge deployment \citep{tencent2026hyembodied,dang2026rynnbrain,yuan2026embodiedr15}. 
Such smaller models are attractive in part because frontier models can be expensive in agentic settings: on one simulated benchmark, an agentic Astra policy incurred nearly 19 times the API fees of direct prompting \citep{yang2026astranomy}.

\paragraph{Fresh Observations and Partial Observability.}
While S2 deliberates, S1 and S0 keep acting and the world keeps changing, so the observation behind a decision has aged by the time the decision takes effect.
S2 must therefore reason about the age of its information, request fresh observations when necessary, and express decisions in forms that remain valid under small changes.
For instance, it should issue ``place the cup on the shelf'' rather than a fixed joint configuration.
Partial observability compounds the problem: S2 must distinguish what it observes, what it infers, and what it does not know, and act on this uncertainty by looking again, asking, or abstaining.

These constraints do not imply that frontier models cannot serve as S2 foundation models; used with a harness, GPT-6 Astra already matches or exceeds trained policies on several simulated benchmarks (Section~\ref{sec:eval-results}).
Instead, the constraints imply that S2 must be judged not only by answer accuracy but also by the quality of its decisions under the deadlines, interruptions, and budgets that the physical world imposes.
This section has examined three aspects of Embodied S2: the foundation model's share of its abilities, the harness that supplies the rest, and the physical constraints on both.
An evaluation must address all three: it must test the model on its own and the complete system in closed loop, and run the closed-loop tests on the world's clock rather than with the world paused.
The next section develops such an evaluation framework.

\section{Evaluation Framework}
\label{sec:evaluation}

Two months after the release of GPT-6 Astra \citep{openai2026gpt6astra}, results in agentic robotics have become difficult to
interpret. On RoboDojo alone, success rates reported for Astra across all 42 tasks range from
22\% to 85\%, depending on harness and protocol
\citep{zhang2026unexpected,sensenova2026has,gao2026hexashellv3}.
The difficulty has three sources. First, evaluations often do not state whether they measure a
foundation model or a complete system. Second, they rarely disclose the full protocol, including
observations, budgets, and demonstrations. Third, many pause the world while the model
deliberates. This section asks how Embodied S2, a foundation model acting through an
embodied harness, should be evaluated. It addresses what to evaluate, along the nine
dimensions of Figure~\ref{fig:eval-circle}; how to evaluate; what current results
show; and where current evaluation falls short.

\input{figures/fig_eval_circle}

\subsection{What to Evaluate}
\label{sec:eval-what}

We evaluate Embodied S2 at two levels.
At the \emph{model level}, the foundation model answers on its own, typically in a single turn without tools; these tests isolate knowledge, understanding, and the model-side components of reasoning and safety.
At the \emph{system level}, the model acts through a harness and, where needed, invokes S1 skills and S0 interfaces; these tests measure closed-loop reasoning, tool use, memory, cross-embodiment transfer, safe execution, efficiency, and cost.
The two levels answer different questions and provide complementary evidence about failure. 
A failure at the model level points to the model, whereas a failure that appears only at the system level points to the model, the harness, the interfaces, or their interaction.
Comparing model-level and system-level results can help narrow down possible causes of failures.
For example, replaying the observations and skill feedback that were available to the model at the point of failure can reveal whether the model would make the same decision again under controlled conditions.
Within system-level evaluation, it also matters whether the harness is fixed across the models compared, supplied by each model's developer, or itself the object of study, because harness changes alone can move success a lot.

Across these two levels, Figure~\ref{fig:eval-circle} organizes the evaluation into nine
dimensions with 53 sub-dimensions: the eight abilities of Embodied S2
(Figure~\ref{fig:eval-illustrated}) and cost, which is not an ability but a measure of the
resources the system consumes. The 29 \emph{embodiment-related} sub-dimensions are specific to embodiment or depend strongly on it, and
standard evaluations of general language models rarely cover them. The reasoning budget
appears under efficiency rather than reasoning because it is a variable to be controlled and reported,
not an ability in itself. Fulfillment of user intent is not a separate dimension but the
end-to-end outcome that all dimensions serve.
The dimensions play different roles and are not orthogonal. Knowledge, understanding, and reasoning describe what the foundation model can do. Tool use, memory, and cross-embodiment describe what the system achieves through its harness and across bodies. Safety constrains all of them. Efficiency describes how quickly and with how few tokens the system decides, and cost the resources it consumes.
Results should therefore be reported as a profile across dimensions rather than summed into a single score.

\paragraph{Knowledge.}
\label{sec:eval-knowledge}
Knowledge tests probe the model's facts, concepts, and procedures through self-contained,
single-turn questions without tools. \emph{World knowledge} spans objects and
everyday relations, life science, matter, the Earth and the universe, and geography,
history, and culture; CommonsenseQA \citep{talmor2019commonsenseqa} is a candidate source.
\emph{Embodied knowledge} spans materials and physical properties, basic physical laws,
object functions, tools and mechanical mechanisms, action preconditions and effects, robot
structure and actuation, sensing and feedback, and coordinate systems and kinematic
fundamentals. PIQA, NEWTON, and RoboCSKBench
\citep{bisk2020piqa,wang2023newton,toberg2025robocskbench} cover parts of it.
\emph{Programming knowledge} spans language semantics, data structures and algorithms,
software interfaces, and the writing of short functions from a specification. CodeMMLU
\citep{nguyen2025codemmlu} tests the first three, and the EvalPlus versions of HumanEval
and MBPP \citep{liu2023your} test the last. In HumanEval and MBPP, the model completes a small, self-contained
function from its signature and description, and unit tests score the result. On a robot,
the same skill produces helpers such as a function that converts a detected bounding box
into a grasp point.
Results should identify the categories tested and the scoring protocol. General
physical-commonsense scores do not establish coverage of robot structure, sensing, or
kinematics, which require explicit probes.
Offline scores also characterize knowledge only under the chosen protocol; closed-loop
competence requires a separate system-level evaluation.

\paragraph{Understanding.}
\label{sec:eval-ur}
Understanding tests ask whether an S2 foundation model can extract an accurate,
task-relevant representation of the world, the user's goal, and the robot's own state from
available observations. Unlike knowledge tests, they concern a particular physical situation. They cover
multimodal perception, spatial and physical state estimation, affordance recognition,
temporal understanding, task interpretation, recognition of human behavior, awareness of the
robot's body and of the skills and interfaces available to it, progress estimation, and
uncertainty assessment. Table~\ref{tab:understanding} lists these abilities with an
evaluation target and candidate benchmarks for each. Two of them deserve particular weight
because they matter for closed-loop control and are weak in current models. Understanding
of the robot's own body and capabilities tells S2 whether to delegate to a skill or to act
directly. Understanding of task progress and completion makes verification possible, yet
success detectors are biased toward reporting success \citep{navasardyan2026failbench}.
Static question answering also misses the temporal side of understanding: because
observations age while S2 deliberates, a model must distinguish what it observed, and when,
from what still holds now. Finally, tests must respect the boundary with
reasoning: understanding determines \emph{what is happening and what can be inferred},
whereas reasoning determines \emph{what should be done next}. For example, recognizing that
an object has slipped from the gripper tests understanding; identifying the cause and
selecting a corrective action tests reasoning.

Evaluating understanding therefore requires separating state inference from downstream
decision-making and execution. At the model level, tests should specify the observations,
task instructions, robot-state information, and interaction history available to the model.
Such specification matters most for progress estimation and uncertainty assessment, where the
correct answer may depend on earlier observations or on information that remains unobservable.
At the system level, task outcomes provide complementary evidence, but they cannot by themselves
attribute failures to understanding, because planning, tool use, and execution may also
contribute. 

\begin{table}[!tbp]
\centering\footnotesize\renewcommand{\defaultaddspace}{0.35em}
\caption{\textbf{Understanding abilities, evaluation targets, and candidate benchmarks.} 
Abilities in bold are specific to embodiment or strongly depend on it, as in Figure~\ref{fig:eval-circle}.}
\label{tab:understanding}
\begin{tabular}{@{}L{0.19\linewidth}L{0.40\linewidth}L{0.35\linewidth}@{}}
\toprule Ability & Evaluation target & Candidate benchmark \\ \midrule
Multimodal perception and semantics & Recognize objects, attributes, and task-relevant entities from multimodal observations. & RoboBench \citep{luo2025robobench} \\ \addlinespace
\textbf{Spatial relations and reference frames} & Infer spatial relationships, relative positions, and geometry across viewpoints and frames. & MMSI-Bench \citep{yang2026mmsibench}, VSI-Bench \citep{yang2024vsibench} \\ \addlinespace
\textbf{Physical states and mechanisms} & Identify physical properties, object states, support and contact relationships, and mechanisms relevant to interaction. & PhysBench \citep{chow2025physbench} \\ \addlinespace
\textbf{Affordances and interaction feasibility} & Identify physically feasible interactions and action regions given the observed scene and available capabilities. & RoboBench \citep{luo2025robobench} \\ \addlinespace
Events and temporal structure & Identify events, their order, and temporal dependencies from observation sequences. & EgoTaskQA \citep{jia2022egotaskqa} \\ \addlinespace
Tasks, goals, and stated constraints & Interpret expressed goals, desired states, and task constraints. & Embodied Agent Interface \citep{li2024embodied} \\ \addlinespace
Human behavior and interaction context & Recognize human activities, intentions expressed through observable behavior, and the current stage of an interaction. & ProAct-75
\citep{zhu2026proact} \\ \addlinespace
\textbf{Embodiment, skills, and control interfaces} & Infer the robot's body state, kinematics, and reachable workspace, the available S1 skills and S0 control interfaces, and
their capabilities and constraints. & EQA-phys \citep{zhou2025physvlm}, HumanCLAW \citep{siyao2026humanclaw} \\ \addlinespace
\textbf{Task progress and outcome state} & Estimate completed steps, remaining progress, and
whether an intended physical effect has occurred. & ContextProgress-Bench
\citep{zhang2026progresscompass}, FailBench \citep{navasardyan2026failbench} \\ \addlinespace
\textbf{Observability, uncertainty, and knowledge limits} & Distinguish observed facts from inferred or unavailable information and recognize when evidence is insufficient. & AbstainEQA \citep{wu2026robots} \\
\bottomrule
\end{tabular}
\end{table}

Together, these tests characterize the information quality on which S2 decisions depend. Whether the agent can use that information to select effective actions is the subject of reasoning evaluation.

\paragraph{Reasoning.}
Reasoning tests ask whether the S2 foundation model can select and revise decisions that advance the user's goal,
given its understanding of the current state, the robot's capabilities, and the constraints
of the physical environment. They cover intent inference, causal and counterfactual
prediction, hierarchical planning, spatial exploration, skill selection and orchestration,
executable action inference, active perception and intervention decisions, failure recovery,
and handling of interruptions and rescheduling. Executable action inference, which corresponds to the direct-control
path of Figure~\ref{fig:exec-paths}, needs a precise evaluation target. We treat it as a
reasoning ability rather than as a separate policy dimension. The target is therefore
whether the model produces physically executable, goal-consistent action
decisions, not whether it can reliably control a robot as a complete policy. Reliable
control also depends on continuous closed-loop operation, latency, controllers, and the
harness, so it belongs to system-level evaluation.

Distinguishing reasoning competence from execution performance requires tests at both
levels. At the model level, \emph{controlled decision-making tests} assess whether the model
selects feasible, goal-consistent actions under specified observations, capabilities, and constraints, without physical execution. At the system level,
\emph{closed-loop tests} additionally assess whether its decisions remain effective when
actions are executed and their outcomes are observed. The distinction matters because task
success depends not only on decision quality but also on the harness, tools, controllers,
and physical environment.
Table~\ref{tab:reasoning} summarizes these abilities and lists candidate benchmarks for each.

\begin{table}[!tbp]
\centering\footnotesize\renewcommand{\defaultaddspace}{0.35em}
\caption{\textbf{Reasoning abilities, evaluation targets, and candidate benchmarks.} 
Abilities in bold are specific to embodiment or depend strongly on it, as in Figure~\ref{fig:eval-circle}.}
\label{tab:reasoning}
\begin{tabular}{@{}L{0.19\linewidth}L{0.43\linewidth}L{0.32\linewidth}@{}}
\toprule Ability & Evaluation target & Candidate benchmark \\ \midrule
Intent inference and disambiguation & Infer underspecified goals and decide whether clarification is needed before acting. & Ask-to-Act \citep{ramrakhya2025grounding} \\ \addlinespace
Causal, predictive, and counterfactual reasoning & Predict the consequences of actions, explain observed outcomes, and compare alternatives under hypothetical changes. & EgoTaskQA \citep{jia2022egotaskqa} \\ \addlinespace
Hierarchical, long-horizon, and constrained planning & Decompose goals into feasible subgoals, order actions, and satisfy dependencies and constraints. & Embodied Agent Interface \citep{li2024embodied}, LADDER \citep{petkov2026ladder} \\ \addlinespace
\textbf{Spatial search and exploration planning} & Select search locations, exploration strategies, and navigation subgoals to reach a target or reduce uncertainty. & GOAT-Bench \citep{khanna2024goatbench} \\ \addlinespace
Skill selection and orchestration & Select, sequence, and coordinate the available skills and tools according to task dependencies, capability constraints, and expected effects. & RobotWorld \citep{yang2026robotworld} \\ \addlinespace
\textbf{Executable action inference} & Generate executable action targets directly, such as end-effector poses, waypoints, trajectories, joint targets, or skill parameters,
that satisfy embodiment, geometric, and environmental constraints. & RoboDojo \citep{chen2026robodojo,zhang2026unexpected}, Robocurve \citep{robocurve2026astra}, EmbodiedBench \citep{yang2025embodiedbench} \\ \addlinespace
Active perception and intervention decisions & Decide when additional observations are needed and whether, when, and how to intervene during an ongoing task. & RoboQuest
\citep{liu2026roboquest}, ProAct-75 \citep{zhu2026proact} \\ \addlinespace
\textbf{Failure diagnosis, correction, and recovery} & Identify the cause of a failed action, choose a corrective strategy, and restore progress without unnecessarily repeating the failure. & ViFailback-Bench \citep{zeng2026vifailback}, FailBench \citep{navasardyan2026failbench}, RESCUE \citep{ahmedt2026rescue} \\ \addlinespace
\textbf{Interruption, rescheduling, and collaboration} & Revise plans after user interventions, environmental changes, or collaborator actions while preserving valid task progress. & Robix evaluation sets \citep{fang2025robix}, PARTNR \citep{chang2025partnr} \\
\bottomrule
\end{tabular}
\end{table}

\emph{Evaluation gaps.} 
Existing benchmarks provide useful measurements of reasoning-related capabilities, but several do not isolate the underlying decision.
For instance, successful navigation does not necessarily establish effective exploration planning, and successful tool-based execution does not isolate skill selection and orchestration.
Likewise, task success alone cannot show that one model generates better actions than another, because inverse kinematics, controllers, the action interface, and the observation contract also affect success.
A more direct protocol would hold the state representation, available capabilities, and execution constraints fixed while scoring candidate decisions before execution.
For failure recovery, existing benchmarks score detection, diagnosis, and correction guidance offline \citep{zeng2026vifailback}; controlled failure injection could measure the same steps during execution, together with recovery success and the additional time or actions required.
Such tests would complement closed-loop benchmarks rather than replace them.

\begin{table}[!tbp]
\centering\small
\caption{\textbf{Tool-use benchmarks, evaluation targets, and corresponding scales.}}
\label{tab:tool-benchmarks}
\begin{tabular}{@{}L{0.19\linewidth}L{0.44\linewidth}L{0.29\linewidth}@{}}
\toprule Benchmark & Evaluation target & Scale \\ \midrule
CaP-Bench \citep{fu2026capx} & Writing and running programs that compose perception and control primitives, with various abstraction, perception, and execution-feedback settings & 7 core manipulation tasks; 187 tasks in the full CaP-Gym release, from Robosuite, LIBERO-PRO \citep{zhou2025liberopro} and BEHAVIOR \citep{li2024behavior1k}\\ \addlinespace
EmbodiedBench \citep{yang2025embodiedbench} & From high-level household planning to low-level navigation and manipulation & 1,128 tasks in four
environments \\ \addlinespace
Embodied\-Skill\-Bench \citep{han2026embodiedskillbench} & Whether packaged skills help general agents: native control, skill-only execution, and agent-directed reuse & 250 skills; 100 perturbed LIBERO-PRO tasks \\ \addlinespace
RobotWorld \citep{yang2026robotworld} & Robot use across embodiments, with interaction budgets and executable success checks & 84 tasks from 20 projects in five domains \\
\bottomrule
\end{tabular}
\end{table}

\paragraph{Tool Use.}
Tool use should be evaluated with the experimental setup held fixed: the same robot, sensors, controllers, and scene configuration, prepared in advance for every model compared.
The unit of an ablation is the actual callable interface rather than a tool family, because a single grasping API may bundle perception, planning, and execution.
Three kinds of experiment answer different questions and should not be conflated.
Freezing the model and varying the tools measures the gain from provisioning.
Freezing the tools and varying the model measures how well models use them.
Freezing both and varying prompts, memory, or feedback measures the harness's effect.

The tools themselves fall into four types, each with its own evaluation target.
\emph{Perception} tools identify and segment targets, estimate spatial geometry, propose
candidate grasps, and report contact; examples are SAM~3 \citep{carion2025sam3}, RGB-D
projection and multi-view triangulation, Contact-GraspNet \citep{sundermeyer2021contact},
and contact queries. Evaluation checks whether the model requests the estimate that the
decision needs and uses the returned geometry in the correct reference frame.
\emph{Motion-planning} tools, such as planning backed by cuRobo
\citep{sundaralingam2023curobo} and waypoint execution, turn target poses or waypoints into
trajectories within kinematic limits and modeled collision constraints. Evaluation checks
whether the model supplies feasible targets and interprets planning failures.
\emph{Action} tools include learned-policy calls and direct-control commands.
Learned-policy calls delegate a subtask to a VLA or WAM, such as $\pi_{0.5}$
\citep{pi2025pi05}, GR00T \citep{nvidia2025gr00t}, or Cosmos3-Nano-Policy
\citep{nvidia2026cosmos3}. Evaluation checks whether the model delegates suitable
subtasks and monitors, accepts, corrects, or overrides the resulting actions where the
interface allows. Direct-control commands for the end effector, the joints, and the
gripper leave the target, the sequence, and the timing of observation to the model.
Evaluation checks whether these commands respect the body's units, frames, and limits.
\emph{Code} tools allow the model to compose perception, planning, and action calls into
executable programs, as in Code as Policies \citep{liang2022code}. Evaluation checks
whether the generated code uses the available APIs correctly, handles feedback and
failures, and terminates appropriately.
The underlying planners and controllers may be shared across types, so a test must name the
interface that the agent actually called.

The ability to use tools has three aspects, each scored separately.
\emph{Effectiveness and reliability} concern whether the model uses tools correctly,
verifies their actual results, and adjusts its actions after failures.
\emph{Speed and efficiency} concern how long a task takes, how many calls it needs, and how
quickly the model corrects a problem once it is detected. They matter because each call to
a frontier model can take seconds (Figure~\ref{fig:latency}). They should be
reported even when success does not change, together with how tools are organized and
scheduled, for example whether the agent keeps reasoning while a tool runs.
\emph{Learning and transfer} concern whether the model distills what it has learned about
tools into skills that improve later tasks, as when frontier models author skills that
later run without them \citep{bai2026compile}. Measuring these aspects apart from
provisioning gains requires paired tests that change one condition at a time, with the
correct behavior and an independent scoring rule fixed in advance. Table~\ref{tab:tool-benchmarks} lists benchmarks that test
tool use and the tool types that each exposes.

\begin{table}[!tbp]
\centering\small
\caption{\textbf{Memory benchmarks, evaluation targets, and corresponding scales.}}
\label{tab:memory-benchmarks}
\begin{tabular}{@{}L{0.22\linewidth}L{0.46\linewidth}L{0.24\linewidth}@{}}
\toprule Benchmark & Evaluation target & Scale \\ \midrule
MIKASA-Robo \citep{cherepanov2026mikasarobovla} & Object, spatial, sequential, and capacity memory & 90 tasks in three horizon splits of 25 to 2,160 steps \\ \addlinespace
RoboMME \citep{dai2026robomme} & Temporal (counting), spatial (permanence), object (reference), and procedural (imitation) memory & 16 tasks $\times$ 50 test episodes; 1,600 training demonstrations \\ \addlinespace
RoboMemArena \citep{lei2026robomemarena} & Multi-object transfer, occlusion, counting, and sequence, in long multi-stage manipulation & 26 simulated and 5 real-world tasks \\ \addlinespace
EMem-Bench \citep{liang2026embodiedmemory} & Passive observation, dynamic tracking, interaction failure, and experience generalization & 2,554 episodes in 1,118 scenes \\ \addlinespace
OpenEQA \citep{majumdar2024openeqa} & Questions about a recorded episode and questions answered by active exploration & More than 1,600 questions from 180 real-world environments \\
\bottomrule
\end{tabular}
\end{table}

\paragraph{Memory.}
\label{sec:eval-memory}
Memory evaluation asks whether a robot uses information that its current observation no longer contains. 
Inspired by MIKASA-Robo \citep{cherepanov2026memory}, we divide memory-intensive tasks into six classes. 
\emph{Object memory} keeps information about objects that become temporarily unobservable, and \emph{spatial memory} keeps object locations and a map of the environment. \emph{Sequential memory} keeps temporally ordered information, and \emph{memory capacity} concerns keeping several pieces of information at once. 
\emph{Body memory} keeps information about the robot's own body, such as its kinematic configuration and action-relevant body relations; \emph{user memory} keeps persistent information about the user, such as preferences and prior instructions.
Table~\ref{tab:memory-benchmarks} lists five benchmarks that together cover tabletop manipulation, household interaction, and question answering. They score different outcomes: manipulation success (MIKASA-Robo \citep{cherepanov2026memory}, RoboMME \citep{dai2026robomme}, and RoboMemArena \citep{lei2026robomemarena}), success on interactive tasks that require building and updating memory (EMem-Bench \citep{liang2026embodiedmemory}), or answers to questions about what the robot has seen (OpenEQA \citep{majumdar2024openeqa}). None of these benchmarks tests body or user memory.

\paragraph{Cross-Embodiment.}
\label{sec:eval-cross}
Cross-embodiment evaluation asks whether a fixed objective remains achievable when the body changes. It has three aspects. \emph{Task transfer performance} compares success rate, execution efficiency, and stability across the change of body, with the same objectives and matched scenes. \emph{Body understanding} concerns whether S2 understands the new body's perception, motion, and manipulation capabilities and adjusts its action parameters and execution plans to them. For example, when a task moves from a small arm to a larger one such as a Franka Panda, aligning the end-effector target pose is not enough. Longer links change the trajectory of the whole arm, so S2 must check again for collisions with the environment and with the robot itself.
\emph{Experience reuse} concerns whether S2 carries over what still applies from its existing skills, revises the parts specific to the old body, and keeps outdated experience from interfering with execution on the new one. 
A transfer protocol should fix the S2 foundation model, pair the scenes across bodies, and report each embodiment separately. It should also disclose all exposure to the target body (demonstrations, adaptation data, tools, and controller changes) and compare only tasks that are feasible for both bodies. Each robot should be described by a versioned \emph{capability card} that lists its frames and units, kinematic and geometric parameters, sensors, tools with their preconditions and postconditions, controller version, and validated capability limits. Table~\ref{tab:cross-embodiment} summarizes five benchmarks evaluating multiple embodiments on tasks including navigation and manipulation.

\begin{table}[!tbp]
\centering\small
\caption{\textbf{Cross-embodiment benchmarks and what each can and cannot cover.}}
\label{tab:cross-embodiment}
\begin{tabular}{@{}L{0.26\linewidth}L{0.24\linewidth}L{0.40\linewidth}@{}}
\toprule Benchmark & Focus & Coverage \\ \midrule
RobotWorld \citep{yang2026robotworld} & Multi-body execution & Execution on many bodies;
cannot isolate morphology, because objectives are not paired across bodies \\
\addlinespace
Habitat-MAS \citep{chen2025emos} & Heterogeneous capabilities and planning &
Planning with robot capability descriptions; tests cooperation, not transfer \\
\addlinespace
OmniNavBench \citep{sun2026beyond} & Navigation across bodies & The same instructions on
wheeled, quadruped, and humanoid agents; navigation only \\
\addlinespace
ZETA \citep{yan2026zeta} & Tabletop manipulation transfer & Transfer to held-out target
bodies; target-data exposure must be disclosed \\
\addlinespace
DexJoCo-X \citep{jiang2026dexjocox} & Dexterous-hand manipulation & Joint training across
seven hands; transfer to an unseen hand is left to future work \\
\bottomrule
\end{tabular}
\end{table}

\paragraph{Safety.}
Safety evaluation should cover the entire action chain, from risk recognition to coordinated interruption. The eight axes in Table~\ref{tab:safety} turn that chain into tests, each with candidate benchmarks. Existing benchmarks test three abilities directly (risk recognition, prudent decisions within capability limits, and prevention before action) and whether the postconditions of post-action handling hold.
For example, IS-Bench \citep{lu2025isbench} checks whether mitigation is performed before and after risk-prone steps. The other four are covered only partially, and executable protocols are still missing for delayed consequences, dynamic disturbances, recovery, cancellation, and takeover. Refusal, safe task completion, serious violations, and excessive refusal must be reported separately, because a refusal rate alone does not measure safe completion. Refusal must also be measured in the action loop, because refusal learned in text transfers only partially to action \citep{sun2026roboharm}.
LLM-based safety judges alone do not suffice: in one study, an LLM judge missed nearly 83\% of unsafe trajectories, labeling them safe or misidentifying the unsafe step \citep{zhan2025sentinel}.
Edits to language models can introduce misinformation and bias with little effect on general capability scores \citep{chen2026editattack}. Robot safety should therefore be retested in the action loop after model updates.

\begin{table}[!tbp]
\centering\small
\caption{\textbf{Safety abilities, from risk recognition to coordinated interruption, with candidate benchmarks and the coverage each provides.} Abilities in bold are specific to embodiment or depend strongly on it, as in Figure~\ref{fig:eval-circle}.}
\label{tab:safety}
\begin{tabular}{@{}L{0.30\linewidth}L{0.28\linewidth}L{0.34\linewidth}@{}}
\toprule Ability & Candidate benchmark & Coverage \\ \midrule
Risk recognition: identify risks involving objects, environments, and people & IS-Bench \citep{lu2025isbench}; HomeSafeBench \citep{yao2025homesafebench} & Risks that emerge during household tasks; hazards reported during free exploration of a home \\ \addlinespace
\textbf{Consequence prediction}: anticipate immediate, delayed, and counterfactual
consequences & SafeMindBench \citep{chen2025safemind}; RoboHarm \citep{sun2026roboharm};
ASIMOV-2.0-Injury \citep{jindal2025can} & Factual, causal, and temporal constraints
in planning; refusal on real arms; risk type and severity in text scenarios from injury
reports, the safety effect of a candidate action, and the severity after it \\ \addlinespace
Prudent decisions within capability limits: choose execution, verification, assistance, or refusal
& ASIMOV-Agentic \citep{deepmind2026gr2safety}; SafeAgentBench
\citep{yin2024safeagentbench} & Decisions under payload and reach limits and under uncertainty; refusal of hazardous tasks in planning \\ \addlinespace
\textbf{Prevention before action}: check preconditions and use a safe action order & IS-Bench,
mitigation before risky steps \citep{lu2025isbench}; DESPITE \citep{zhang2026using} & Mitigation performed in the correct order; dangerous plans among valid ones in symbolic planning \\ \addlinespace
\textbf{Monitoring and replanning}: observe execution feedback and revise the plan after a
deviation & SENTINEL \citep{zhan2025sentinel}; RT-Safe \citep{xu2026rtsafe} & Formal checks of plans and trajectories; collisions in a world that keeps running. Timed faults and dynamic disturbances are missing \\ \addlinespace
Post-action handling and recovery: remove residual hazards, recover, or terminate
safely & IS-Bench, mitigation after risky steps \citep{lu2025isbench} & Postconditions; protocols for incidents and recovery from failure are missing \\
\addlinespace
\textbf{Timely safety response}: intervene effectively within a specified safety window & RT-Safe \citep{xu2026rtsafe}; ASIMOV-2.0-Video \citep{jindal2025can} & Safety events when the world keeps moving during inference, in simulation; predicted intervention times in videos. End-to-end timing on robots is missing \\ \addlinespace
\textbf{S2/S1 coordination and interruption}: propagate constraints, cancel commands, stop, or
transfer control & ASIMOV-Agentic \citep{deepmind2026gr2safety} & Stop-or-continue decisions and the feasibility of VLA calls \\
\bottomrule
\end{tabular}
\end{table}

\paragraph{Efficiency and Cost.}
Efficiency evaluation measures the latency from observation to action, the number of model calls, the freshness of the observations on which each decision rests (the observation age defined in Equation~\ref{eq:obs-age}), and the rate of timeouts and missed deadlines, under a stated reasoning budget and response deadline.
Output tokens per second measure decoding speed, not robot responsiveness. Latency should therefore include perception, context preparation, inference, network delay, tool execution, and controller response whenever they are on the critical path.
Cost should be reported as tokens, API charges, total task time, and robot occupancy, per attempt and per success, including failures and retries.
Reported costs differ by orders of magnitude, and cost rankings need not follow success rankings. In Astranomy, an agentic Astra session used 30.9M tokens (\$36.40) per successful episode, against 160k tokens (\$1.96) for direct prompting, although most of its input was served from cache \citep{yang2026astranomy}.
On RoboQuest, an average episode with Astra took 10.4 minutes of simulated time but 53.8 minutes of wall-clock time and cost \$11.29 \citep{liu2026roboquest}.
On a single real six-axis arm, frontier coding agents took about 30 minutes per task that a person completes in 20 seconds \citep{artificialanalysis2026aarobotics}.

\paragraph{User Intent.}
\label{sec:eval-intent}
For user intent, explicit instructions are the easy case; harder cases include vague requests, unstated preferences, learned habits, and changes of mind during a task. Evaluation should therefore score the final state against the user's intent rather than the literal instruction. It should also report the number of clarifying questions, adherence to stated and learned preferences, the handling of corrections and interruptions, and the number of human interventions.
Existing benchmarks cover only parts of intent fulfillment.
PRISM isolates implicit intent resolution, which was ``a significant bottleneck for all model families''
\citep{lim2026prism}. TEACh requires acting on intent conveyed over several dialogue turns
\citep{padmakumar2022teach}, and Ask-to-Act scores clarification of ambiguous requests
\citep{ramrakhya2025grounding}. MEMENTO finds that agents ``can recall simple object
semantics but struggle to apply sequential user patterns to planning''
\citep{kwon2025embodied}. Two devices can be borrowed from digital agents: a hidden supervisor
with a simulated user who gives multi-turn feedback ``without leaking grading criteria''
\citep{chen2026uniclawbench}, and the pass$^k$ reliability metric \citep{yao2024taubench}.
To our knowledge, no benchmark yet scores end-to-end intent fulfillment on real robots with real users across sessions.

\subsection{How to Evaluate}
\label{sec:eval-how}

The dimensions above specify what to measure. Comparable results also require a common way
of measuring, which the same benchmark name does not guarantee. We therefore propose the following principles, several of which are adapted from the evaluation of digital agents
\citep{yao2024taubench,kwa2025measuring}.
\begin{enumerate}[label=\roman*., leftmargin=*, itemsep=2pt, topsep=3pt]
  \item \textbf{Separate the model from the system.} Fix the harness when comparing
        foundation models, and fix the model when comparing harnesses. In both cases, match the
        embodiment, interfaces, observations, task instances, and budgets. Otherwise,
        label the comparison as one between complete systems, as CodeActionBench does
        \citep{lyu2026codeactionbench}. What counts as ``no tools'' must be defined by the
        interface contract rather than by name: in EmbodiedSkillBench, for example, the ``no skill''
        condition still acts through registered native end-effector and gripper tools
        \citep{han2026embodiedskillbench}.
  \item \textbf{Treat the action interface as a controlled variable.} The same foundation
        model may call skills, emit end-effector targets, or output joint targets
        (Figure~\ref{fig:exec-paths}). These interfaces do not form a scale from easy to
        hard: each brings different execution priors, control burdens, failure modes, and
        degrees of freedom. Model comparisons should therefore fix the action abstraction
        level and report it. Its effect can be measured directly by holding the model,
        tasks, robot, sensors, environment, and budgets fixed while varying the interface.
        The comparison should include a skill-only interface, a direct end-effector
        interface, and a hybrid in which the model decides
        when to call a skill and when to act directly. Such comparisons should report task
        success separately from action validity, collisions and constraint violations, the
        number of calls, physical execution time, inference latency, recovery, and total
        cost.
  \item \textbf{Run on a world clock.} Many simulated evaluations pause the world while the
        model deliberates \citep{jedlicki2026memory,galbot2026systematically,yang2026robotworld}.
        Paused and live execution measure different abilities and must be reported
        separately. On matched hard routes of RT-Safe, live execution raised collisions by a
        factor of 12.3 but left task completion essentially unchanged. Safe
        success fell from 19.8\% to 0.7\%, and mean latency was positively associated with collisions
        across models \citep{xu2026rtsafe}. 
        Embodied First Exam separates a Standard setting, in which the agent receives sensor observations and the simulation pauses while it deliberates, from a Realtime setting, in which the environment continues to advance during decision-making. On 54 dynamic core tasks evaluated in both settings, GPT-6 Astra solved 23 Standard tests and 17 Realtime tests \citep{wang2026embodiedexam}.
        Live evaluations should report the rate of
        missed deadlines and the distribution of the \emph{observation age}. The age is defined as
        \begin{equation}
          \Delta t_{\mathrm{obs}} = t_{\mathrm{effect}} - t_{\mathrm{capture}},
          \label{eq:obs-age}
        \end{equation}
        where $t_{\mathrm{capture}}$ is when the observation behind a decision was
        captured and $t_{\mathrm{effect}}$ when the
        resulting action takes effect. Because it includes perception, communication,
        queuing, and execution delays, observation age reflects the physical response more
        closely than model inference latency does. When sensors, computation, and
        actuators run on different devices, both times must be measured on synchronized
        clocks.
  \item \textbf{Disclose the observation contract.} State what the model and its tools
        receive: cameras and their history, the source of depth (a sensor, a prediction,
        or triangulation), privileged object state, success or progress signals, and any
        demonstrations or scene assets. Give the agent only information that a real
        robot could also obtain \citep{openrua2026revisiting}. Without this disclosure,
        numbers from different studies cannot be compared.
  \item \textbf{Verify completion independently.} Decide success with a checker that has
        access to ground truth that the system cannot observe, and never accept the
        model's own claim or a successful tool return as proof. Report the rate of false
        completion claims, because agents often declare success prematurely
        \citep{jedlicki2026memory,liu2026roboquest} and VLM judges are biased toward success
        \citep{navasardyan2026failbench}. The checker must also be inaccessible to the
        agent. In one study, agents read static object files that a physical robot could
        not read and located a hidden success check within a single session
        \citep{wang2026sear}. Hiding the verifier changes results substantially. When tool
        returns no longer revealed success, agents had to decide for themselves when an
        episode ended. Under this condition, the mean scores of two systems over 50 episodes on 10 tasks of RoboDojo fell from 83.2 to 63.5 and from 42.0 to 26.1 \citep{openrua2026revisiting}.
  \item \textbf{Quantify uncertainty and reliability.} Report per-task results, the number
        of repetitions, and confidence intervals. A difference of a few percentage points on a handful of real-robot trials rarely
        supports a conclusion \citep{team2025careful,kressgazit2024robot}. Reliability
        across repeated runs, such as the probability that all $k$ trials succeed, should
        complement the chance of at least one success \citep{yao2024taubench}. Rankings
        that aggregate benchmarks need the same care. The Robot Intelligence Index, for
        example, reports for each model the range within which its rank falls in 90\% of
        resampled re-runs \citep{playground2026rii}.
  \item \textbf{Control and sweep budgets.} Fix and report the reasoning effort, the
        budgets for calls, steps, and wall-clock time, and any retry rules, and sweep them
        into success--cost and success--latency curves where possible. In Astranomy,
        raising the effort from medium to extra-high improved success only from 72.7\% to
        74.5\% at 2.6 times the latency and 2.2 times the cost \citep{yang2026astranomy}.
        Budgets must also match the benchmark's own limits: recounted under RoboDojo's native
        step limit, one system's success rate fell from 47.1\% to 1.0\%, because 97 of its 99
        successes came after that limit \citep{openrua2026revisiting}. Improvement methods
        should be compared with retry baselines under the same budget, with development and
        test tasks kept apart \citep{wang2026rethinking,wang2026sear}. For example, a skill developed on
        one task with the benchmark's verifier raised that task's score from 0 to 80 yet left two
        of four other tasks at baseline \citep{openrua2026revisiting}. Results that use context about the test tasks
        should not be compared with zero-shot results \citep{openrua2026revisiting}.
  \item \textbf{Validate on real robots.} Treat simulation as a screen rather than a
        verdict, and audit the shortcuts of shared simulated suites. In SIMPLER, simulated
        success tracked real success closely once control and visual gaps to the real setup were
        mitigated, and less closely when they were only partly mitigated \citep{li2024simpler}.
        Policies trained in simulation can also fail badly on real robots \citep{ibarz2021train}. Real-robot
        evaluations should pair seeds and scenes and report the human assistance and
        environment resets they required.
  \item \textbf{Fix the conditions of a level claim in advance.} Some results are offered as evidence
        that a robot carries a level of delegated responsibility (Section~\ref{sec:ladder}).
        Such a result should fix beforehand the initial conditions, the permitted assistance and
        interruptions and how they are counted, and the evidence of completion. Completing a
        long script does not demonstrate delegation if a person must keep repairing the
        plan. The duration of the work should be measured in the time that a skilled person
        would need for it, as METR \citep{kwa2025measuring} does for software agents. The
        robot's own wall-clock time and the human assistance it received should be reported
        separately, because a robot may need far longer than a person for the same work.
\end{enumerate}

The checklist below collects the information that these principles require every report
to disclose.
\begin{tcolorbox}[enhanced, colback=STwoColor!4, colframe=STwoColor!60, boxrule=0.8pt, arc=4pt, left=8pt, right=8pt, top=6pt, bottom=6pt]
\textbf{Reporting Checklist: The Harness Card.} Every reported S2 result should be
accompanied by a \emph{harness card}. The card records (i)~the foundation model, its version,
and its reasoning budget, and (ii)~the harness version, prompts, skill library, memory, and
verifiers. It records (iii)~the embodiment and the S1 and S0 interfaces, with the action abstraction
and control rates, and (iv)~the observation contract and any privileged information. It also records (v)~the
time semantics, that is, whether the world pauses during inference, and (vi)~the number of runs and seeds, with confidence
intervals. Finally, it records (vii)~the rate and definition of human or rule-based interventions and
the paths for cancellation and human takeover.
\end{tcolorbox}

\subsection{What Current Results Show}
\label{sec:eval-results}

Table~\ref{tab:results} collects published results for general models serving as the S2
foundation model. Most come from preprints posted within two months of Astra's release; we report them as published and have not reproduced them. The rows are grouped by what they show: competitiveness on narrower simulated suites, low success on broad or exploratory benchmarks, and
sensitivity to the interface, live execution, and physical action. Read in light of the principles above, these and other results support seven conclusions, which we qualify by the strength of the evidence.

\begin{table}[!tbp]
\centering\small\renewcommand{\defaultaddspace}{0.3em}
\caption{\textbf{A snapshot of published results for general models serving as the S2
foundation model (October 2026).} Rows are grouped by the conclusions of
Section~\ref{sec:eval-results} that they support. S1 lists the skills and task tools that the
model invoked; \emph{none} means that it sent end-effector, joint, or motion targets directly to
low-level control. Settings differ, so rows are not comparable; several of the simulations pause while the model deliberates.}
\label{tab:results}
\begin{tabular}{@{}L{0.24\linewidth}L{0.17\linewidth}L{0.25\linewidth}L{0.24\linewidth}@{}}
\toprule Benchmark & S2 foundation model & S1 components & Comparison \\ \midrule
\rowcolor{STwoColor!7}\multicolumn{4}{@{}l}{\textbf{Competitive with trained policies on narrower simulated suites}} \\ \midrule
RoboDojo-Sim \citep{chen2026robodojo,zhang2026unexpected} & GPT-6 Astra & None & 22.48\%; above all 40 public policies at the time \\ \addlinespace
RoboDojo-Sim, official evaluation \citep{gao2026hexashell} & GPT-6 Astra & Agent-written tools, mostly task-specific & 46.9\%; first on the leaderboard when posted \\ \addlinespace
RoboLab-120 \citep{yang2026robolab,yang2026astranomy} & GPT-6 Astra, direct / agentic session & None & 66.3\% / 74.8\%; best leaderboard policy: 51.0\% \\ \addlinespace
LIBERO-PRO \citep{zhou2025liberopro,si2026pyrualean} & GPT-6 Astra & Motion primitives, segmentation, and $\pi_{0.5}$, composed in code & 94.0\%; $\pi_{0.5}$: 53.0\% (different evaluation setup)  \\ \addlinespace
RoboMME \citep{dai2026robomme,jedlicki2026memory} & GPT-6 Astra & None & 92.1\%; best leaderboard entry: 60.8\% \\ \addlinespace
LADDER \citep{petkov2026ladder} & GPT-6 Astra & None & 62.0\%; learned-policy systems: at most 29.6\% \\ \midrule
\rowcolor{STwoColor!7}\multicolumn{4}{@{}l}{\textbf{Far from solved on broad or exploratory benchmarks}} \\ \midrule
RobotWorld \citep{yang2026robotworld} & GPT-6 Astra & None & 19.0\%; 63 tasks solved by none of five models \\ \addlinespace
RoboQuest \citep{liu2026roboquest} & GPT-6 Astra & None & 23.2\%; fine-tuned $\pi_{0.5}$ almost never succeeds \\ \addlinespace
HumanCLAW \citep{siyao2026humanclaw} & Best of nine VLMs & Whole-body skills such as walking, turning, and climbing stairs & 16.8\%; failures arise from losing track of the body \\ \midrule
\rowcolor{STwoColor!7}\multicolumn{4}{@{}l}{\textbf{Sensitive to the interface, live execution, and physical action}} \\ \midrule
EmbodiedSkillBench \citep{han2026embodiedskillbench} & GPT-6 Astra & None / selected skills only / skills on demand & 92\% / 61\% / 85\%; the interface can cap the model \\ \addlinespace
RT-Safe, hard routes \citep{xu2026rtsafe} & Eight VLMs & None & Safe success 19.8\% (paused) vs.\ 0.7\% (live); collisions rise 12.3-fold in the live world \\ \addlinespace
RoboHarm, real robot \citep{sun2026roboharm} & Six frontier models & None & Refusal 72\% (text) vs.\ 18\% (robot); refusal in text transfers only partly to action \\
\bottomrule
\end{tabular}
\end{table}

\paragraph{General models with harnesses are competitive on narrower simulated suites.}
On several simulated benchmarks, a general model with a harness matches or exceeds trained
policies, particularly on tasks that reward semantic understanding, memory, or flexible
composition
\citep{gao2026hexashell,yang2026astranomy,si2026pyrualean,jedlicki2026memory,petkov2026ladder}.
On LADDER, for example, orchestration recovers part of the success that learned policies lose
as tasks lengthen \citep{petkov2026ladder}. In several comparisons, however, the model received information that the policies did not,
such as a written task recipe \citep{zhang2026unexpected,li2026embodiedevo}. The result is also model-specific: under
the same action interface, a previous-generation model reached 0.88\% on RoboDojo, whereas
Astra reached 22.48\% \citep{zhang2026unexpected}.

\paragraph{Broad benchmarks remain far from solved.}
When tasks span domains or demand active exploration, success remains low: the best model
reached 19.0\% on RobotWorld \citep{yang2026robotworld}, 23.2\% on RoboQuest
\citep{liu2026roboquest}, and 16.8\% on HumanCLAW \citep{siyao2026humanclaw}. The high
scores in Table~\ref{tab:results} come from narrower suites. Recognition also does not imply
completion: on HumanCLAW, the best model found the target in 64.9\% of episodes but
completed only 16.8\% \citep{siyao2026humanclaw}.

\paragraph{Current weak points are physical and evidential.}
Failures cluster on two sides: physical action and perception, and the gathering and judging of evidence. On the physical side, major challenges remain in precise manipulation and contact-rich interaction \citep{gao2026hexashell,yu2026physladder}, as well as dynamic control and locomotion; Astra, for example, solved none of the 11 locomotion tasks in RobotWorld \citep{yang2026robotworld}. Spatial understanding has improved but remains below human level. In the evaluation of \citet{deepcybo2026physbrain15}, GPT-6 Astra at low reasoning effort scored 59.8 on VSI-Bench, against 45.4 for the best model in the original VSI-Bench evaluation and 79.2 for human evaluators on a 400-question subset \citep{yang2024vsibench}. On MMSI-Bench, the strongest proprietary model at publication answered about 40\% of multi-image spatial questions correctly, against 97\% for humans \citep{yang2026mmsibench}. Body understanding is weak as well: in HumanCLAW, failures stem not from misrecognizing the target but from the model losing track of its own body \citep{siyao2026humanclaw}. On the evidential side, agents struggle to explore actively for task-relevant evidence \citep{liu2026roboquest}, to recognize what cannot be answered from observation, and to judge completion. On AbstainEQA, for example, the best frontier model recognized 42.79\% of unanswerable questions, against 91.17\% for humans \citep{wu2026robots}. Agents also declare success prematurely \citep{jedlicki2026memory,liu2026roboquest}; this was the most common failure of a frozen Astra policy on memory-dependent manipulation \citep{jedlicki2026memory}. Automatic judges share this bias toward reporting success \citep{navasardyan2026failbench}.

\paragraph{Harness and protocol move results substantially.}
With the model held fixed, changes to verifiers, memory, interfaces, and post-call decision-making moved success rates by about 9 to 35 percentage points in four studies \citep{siyao2026humanclaw,wei2026scoper,si2026pyrualean,ge2026urai}. In a fifth, restricting agents to skills moved them by up to 52 points \citep{han2026embodiedskillbench}. Changing only the interface illustrates the effect: with the same Astra planner, primitives, and call budget, Python cells raised success over tool calling from 63.1\% to 71.7\% \citep{si2026pyrualean}. Changing the model can have effects of similar size, as the RoboDojo comparison above shows. Studies that cross models with harness conditions remain small, with at most four models and a few conditions on the same tasks \citep{ge2026urai,wei2026scoper,han2026embodiedskillbench}, so these figures show that both factors matter, but not which is generally larger. Combined differences in harness and protocol widen the spread further. For Astra on RoboDojo, one report gives 22.48\% with one seed, a task recipe from the RoboDojo wiki, and direct end-effector control \citep{zhang2026unexpected}. Another gives 37.82\% with the same wiki recipes plus experience reused from practice scenes and a higher reasoning effort \citep{li2026embodiedevo}. A third gives 46.9\% in the official evaluation with a tool library built on the benchmark's own tasks \citep{gao2026hexashell}; its later version, evaluated by its authors on public test layouts, gives 69.9\% with a small set of general tools and no task-specific input, and 85.0\% with about a page of guidance per task \citep{gao2026hexashellv3}. A fourth gives 81.9\% over five episodes per task, with in-context demonstrations and up to ten times the step budget \citep{sensenova2026has}. 
Embodied First Exam also finds substantial harness effects on a common set of 456 tasks: GPT-6 Astra solved 34\% in Codex and 22\% in Claude Code under its Standard setting \citep{wang2026embodiedexam}.
Without a harness card, such numbers cannot be compared.

\paragraph{Tools add capability only beyond what the model can already do.}
More tools do not necessarily give better results. In EmbodiedSkillBench, restricting
agents to skills selected for each task from a repository of 250 action, perception, and experience skills lifted GPT-5.6-Sol
from 11\% to 63\% success but lowered GPT-6 Astra from 92\% under native control to 61\%.
When Astra could decide for each skill whether to invoke, modify, or ignore it, it recovered
to 85\% \citep{han2026embodiedskillbench}. Skills thus help an agent only as far as they
exceed what it can already do. They can also reduce cost, even where they add no success.
In the same benchmark, skill-only execution cut the token usage of the weaker agents by 58\%
to 85\%. For Astra, which gained no success from skills, selective use still saved 14.1\% of
its tokens and 11.9\% of its time relative to native control \citep{han2026embodiedskillbench}.
How tools are organized and scheduled matters as well. The motion, perception, skill, and
episode tools of the uni-robot-agent harness share arguments, units, and feedback. In this
harness, letting the agent keep reasoning while a tool runs shortened each of the five tasks
tested, in one episode each, by 4.8\% to 73.7\%. Swapping two T-shaped blocks, for example, took 12.5 instead of
34.7 minutes, although the authors do not attribute the whole reduction to scheduling.
Removing the harness's predefined skill library, by contrast, left an open-weight Qwen model with 96.6\% of its score, so
effective execution did not depend mainly on that library \citep{openrua2026unirobot}.

\paragraph{Paused worlds hide latency, cost, and risk.}
Frontier models take seconds to tens of seconds per decision (Figure~\ref{fig:latency}),
and many evaluations hide this delay by pausing the simulation. In a world that keeps
running, latency is associated with collisions \citep{xu2026rtsafe}; in one real-robot evaluation,
deliberation dominated task time, which was about 100 times a person's
\citep{artificialanalysis2026aarobotics}. Several designs therefore move work out of the
model loop: program-level control \citep{si2026pyrualean}, VLA-first execution that consults
the S2 foundation model only when needed \citep{sun2026call}, and skills written offline by a
frontier model that then run without it \citep{bai2026compile}.

\paragraph{Safety learned in text transfers only partially to action.}
Pooled over the same six models, refusal fell from 72\% of text requests to 18\% of
real-robot trials with end-effector control \citep{sun2026roboharm}. This drop highlights a substantial text-to-action refusal gap under the tested conditions, and a second study corroborates it
\citep{grover2026runtime}. Safety must therefore be measured in the action loop, along the
full chain from recognizing a risk to stopping in time.

\paragraph{What Is Not Yet Supported.}
Current evidence does not support cross-paper rankings, because interfaces, privileged information, budgets, and time semantics differ. Nor does it support claims of reliability, because many headline results come from single runs or few seeds. Aggregate rankings have begun to address the first problem. The Robot Intelligence Index, launched with 16 public simulated and real-robot benchmarks and extended to 22 by November 4, 2026, combines them into one rating per model using a weighted Bradley--Terry model. In this model, a benchmark's weight is the independence of its evaluators multiplied by the repeatability of its scores \citep{playground2026rii}. The index itself shows, however, why a single rating cannot replace the profile we recommend: 13 of the 23 models that rank in the top three on some benchmark rank in the bottom half on another. Performance in the real world, in continuous time, and over long horizons also remains largely unmeasured, including memory across sessions and the fulfillment of user intent on real robots. Several further findings rest on a single study so far. One is that common embodied-reasoning benchmarks did not track closed-loop control \citep{yang2025vlaser}; another is that, when the world changes, memory fails mainly at replacing outdated state rather than at retaining history \citep{liang2026embodiedmemory}. Finally, benchmarks saturate quickly. VLAs that exceeded 90\% on LIBERO \citep{liu2023libero} dropped to around 50\% on LIBERO-PRO \citep{zhou2025liberopro}, yet an agentic system reached 94.0\% on LIBERO-PRO's perturbed suites within weeks of Astra's release \citep{si2026pyrualean}.
Evaluation must therefore move toward live execution, real robots, longer horizons, and the embodiment-specific sub-dimensions of Figure~\ref{fig:eval-circle}.

\subsection{Limitations of Current Evaluation}
\label{sec:op-eval}

Moving evaluation in these directions requires more than the principles above, because
current evaluation remains limited even when they are followed. The most fundamental
limitation is coverage: current evaluation spans too little responsibility, space, and time
to test agentic robotics as we envision it. The others concern the validity of simulated results, comparability in continuous time,
credible acceptance of complex tasks, and the cost of scaling real-world evaluation.

\paragraph{Coverage of Responsibility, Space, and Time.}
Table~\ref{tab:eval-coverage} lists the spatial scope and time span of benchmarks reviewed in this paper. Most of them test primitives and single tasks in a workcell or a
room, within one episode. Building-scale scenes appear mainly in navigation and memory benchmarks, and site-scale settings in driving, aerial control, and city navigation. Evaluations beyond one episode use a
continuous three-hour simulation, histories of earlier sessions, or learning across tasks. Responsibility is covered least: no benchmark we
reviewed tests a workflow of interdependent tasks that share resources and deadlines, or a
role sustained over days.
How to construct benchmarks for workflows and roles at building and site scale, over days rather than minutes, remains open.

\begin{table}[!tbp]
\centering\small
\caption{\textbf{Spatial scope and time span of benchmarks reviewed in this paper.}}
\label{tab:eval-coverage}
\begin{tabular}{@{}L{0.42\linewidth}L{0.25\linewidth}L{0.25\linewidth}@{}}
\toprule Benchmark & Spatial scope & Time span \\ \midrule
LIBERO \citep{liu2023libero} & Workcell & Single episodes \\
LIBERO-PRO \citep{zhou2025liberopro} & Workcell & Single episodes \\
CALVIN \citep{mees2022calvin} & Workcell & Single episodes$^{a}$ \\
RoboDojo \citep{chen2026robodojo} & Workcell & Single episodes$^{b}$ \\
RoboTwin 2.0 \citep{chen2025robotwin} & Workcell (dual-arm) & Single episodes \\
RoboLab \citep{yang2026robolab} & Workcell & Single episodes \\
VLABench \citep{zhang2025vlabench} & Workcell & Single episodes \\
UniVTAC \citep{chen2026univtac} & Workcell (contact) & Single episodes \\
Bench2Dex \citep{yang2026bench2dex} & Workcell (two hands) & Single episodes \\
DexJoCo \citep{wang2026dexjoco} & Workcell (hands) & Single episodes \\
LADDER \citep{petkov2026ladder} & Workcell & Single episodes \\
RoboMME \citep{dai2026robomme} & Workcell & Single episodes \\
RoboMME-Interference \citep{rathi2026robomme} & Workcell & After unrelated sessions$^{c}$ \\
RobotWorld \citep{yang2026robotworld} & Workcell to outdoors & Interaction budgets \\
ALFRED \citep{shridhar2020alfred} & Room (household) & Single episodes \\
RoboCasa365 \citep{nasiriany2026robocasa365} & Room (kitchens) & Single episodes \\
RoboQuest \citep{liu2026roboquest} & Room (kitchens) & Single episodes \\
MessyMem \citep{banwasi2026messymem} & Room (kitchen) & Over three hours$^{d}$ \\
R2R \citep{anderson2018vln} & Building & Single episodes \\
R2R-CE \citep{krantz2020vlnce} & Building & Single episodes \\
HumanCLAW \citep{siyao2026humanclaw} & Building (houses) & Single episodes \\
OmniNavBench \citep{sun2026beyond} & Building & Single episodes$^{e}$ \\
BEHAVIOR-1K \citep{li2024behavior1k} & Building (homes, offices) & Single long episodes \\
EMem-Bench \citep{liang2026embodiedmemory} & Building (homes) & One task after a history \\
EmbodiedCLBench \citep{huang2026embodiedclbench} & Household scenes & Learning, then test \\
RT-Safe \citep{xu2026rtsafe} & Site (city routes) & Continuous time \\
\bottomrule
\multicolumn{3}{@{}p{0.94\linewidth}@{}}{\footnotesize $^{a}$\,Chains of five consecutive instructions. $^{b}$\,8 to 76~s of
commanded motion \citep{sensenova2026has}. $^{c}$\,Each task's memory holds its relevant prior
demonstration followed by a controlled number of unrelated sessions. $^{d}$\,25 consecutive tasks without resetting memory or the environment.
$^{e}$\,About 49~s on average.}
\end{tabular}
\end{table}

\paragraph{Validity of Simulated Results.}
Simulation is useful for S2 only to the extent that it preserves what S2 must respond to:
realistic observations, informative feedback, failures, and changes in the environment.
Simulators may have limited fidelity in modeling complex physical interactions like deformation, gear backlash, wear and fluid dynamics
\citep{tobin2017domain}, and on real robots, physical parameters such as motor strength, latency, and battery voltage change over time
\citep{tan2018simtoreal}. Real-to-sim evaluation can track real-world rankings for some
policies \citep{li2024simpler}, but its effectiveness depends on the policy
\citep{zhou2025autoeval}, and we know of no comparable validation for Embodied S2. Whether
rankings of S2 foundation models and harnesses obtained in simulation hold on real robots,
and which properties of a simulator decide this, remain open questions.

\paragraph{Comparability in Continuous Time.}
Many agentic evaluations pause the world while the model deliberates \citep{sensenova2026has}. One report estimates
that the budgets of RoboDojo \citep{chen2026robodojo} correspond to 8 to 76 seconds of
commanded motion, against an average of 55 minutes of computation. It summarizes the
effect: ``The model reasons as though the world waits for it'' \citep{sensenova2026has}. A
fair comparison in a world that keeps moving must count each system's own reaction latency
as part of its performance. Yet such a world does not repeat, so runs differ in timing
as well as in decisions. How to control for timing variation across runs at a cost that
real robots can bear is an open question.

\paragraph{Credible Acceptance of Complex Tasks.}
A robot's own judgment that a task is complete differs from external acceptance of the
result. Automatic judges cannot yet provide such acceptance: VLM judges of robot task success have shown a systematic bias toward predicting success when the evidence is ambiguous
\citep{navasardyan2026failbench}. False alarms are also costly: under a stop-and-reset
model of intervention, even an idealized failure detector can afford a false-alarm rate of
only 2.0\% to 3.3\% before unnecessary resets outweigh the savings from stopping failures
early \citep{ahmedt2026rescue}. Even an accepted result is hard to attribute to specific design choices. RobotWorld, for example, cautions that its
five-model snapshot supports ``matched-task observational comparisons, not controlled
ablations'' of history, code control, or reasoning budget \citep{yang2026robotworld}. Workflows and roles make acceptance
harder still, because their acceptance criteria span many subgoals and depend on the user's
intent. How to obtain acceptance judgments with known error rates for such work is open.

\paragraph{Scaling Real-World Evaluation.}
Because real-robot trials are usually slow and expensive, many reported real-robot results rest on few trials per task.
This scarcity is especially consequential because results for the same task vary across runs and across the testers who prepare each trial \citep{yakefu2025robochallenge}.
Distributed real-robot evaluation \citep{atreya2025roboarenax}, autonomous real-world evaluation \citep{zhou2025autoeval}, remote access to standardized hardware \citep{chen2026robodojo,yakefu2025robochallenge}, and third-party real-robot leaderboards \citep{robocurve2026site} are promising steps toward evaluation at scale.
For evaluating Embodied S2, one reported protocol offers a reasonable minimum standard \citep{sensenova2026has}: it sets a fixed wall-clock budget, stations an operator at the emergency stop for the entire trial, and requires a human to judge success from the physical outcome rather than from the model's own completion claim.
The open question is how to obtain statistically reliable evidence as evaluation extends to larger spaces, longer time periods, and broader task responsibilities—settings in which each trial costs more.

\section{Digital Agents versus Physical Agents}
\label{sec:digital-physical}

Agentic robotics inherits a central principle from digital agents: a foundation model can solve complex tasks by repeatedly observing an environment, invoking tools, and verifying the results.
Its architecture, a foundation model in a loop with tools, context, and memory, draws on two lineages. One is digital agents, which have advanced rapidly on coding, computer use, and research tasks; the other is earlier work on robots driven by language models.
The success of digital agents, however, rests on favorable properties of the digital world: interfaces that the model reads and writes directly, outcomes that can be checked automatically, abundant data, and environments that are cheap to copy and reset.

Physical environments weaken each of these properties and add constraints of time,
irreversibility, and embodiment. The challenge is therefore not to give a digital agent
cameras and robotic tools, but to determine which assumptions no longer hold, which
principles still transfer, and what must be rebuilt for reliable physical execution. Our
answer is that the principles of digital agents largely transfer, but the assumptions that
make them reliable do not.
We first review what digital agents can do and why, and then take up these three questions in turn.

\subsection{What Digital Agents Can Do and Why}
\label{sec:digital-can}

The length of software tasks that frontier agents can complete has grown steadily.
METR \citep{kwa2025measuring} measures this progress as a \emph{time horizon}: the duration, in human working time, of tasks that an agent completes with a given probability.
The horizon has doubled roughly every seven months since 2019 \citep{kwa2025measuring,metr2026th11}, and about every four months since 2023 on METR's revised task suite \citep{metr2026th11}.
By 2026, the best measured 50\% horizons exceeded ten hours, and METR cautions that measurements above 16 hours are unreliable with its current task suite \citep{metr2026timehorizons}.
These results measure capability on particular distributions of software tasks with checkable specifications, not reliable long-horizon autonomy in general.
Reliable completion is harder: time horizons at 80\% success are several times shorter than at 50\% \citep{kwa2025measuring,metr2026timehorizons}.

A digital agent pairs a foundation model with a harness of tools, context management, memory, and an execution sandbox. Its results benefit from four conditions that software happens to satisfy.
We call them \emph{enabling conditions}: they recur across the systems behind much of the recent progress, and we treat them as design guidance.
First, \emph{the interface is native to the model}: state and actions take the form of the model's own inputs and outputs. State is text that the agent reads, and actions are text or code that it emits directly through an agent--computer interface \citep{yang2024sweagent}.
\emph{Outcomes are generally verifiable}: compilers, tests, and program execution return objective signals quickly and at scale, signals that support iteration at test time \citep{yang2024sweagent} and reinforcement learning with verifiable rewards \citep{lambert2024tulu3,deepseek2025r1}.
\emph{Data are abundant}: public code, documentation, and discussion accumulated over decades give pretrained models broad knowledge of the domain; Codex, for example, was trained on code from 54 million public GitHub repositories \citep{chen2021evaluating}.
\emph{Environments are cheap}: a sandbox with an executable runtime and unit tests can be copied, run in parallel, reset, and replayed, so that agents can be trained and evaluated on thousands of real tasks \citep{pan2024swegym}.
The first two conditions shape every step of the agent's loop; the last two determine how much the agent can learn before deployment and how widely it can be tested.
\citet{wei2025verifier} links verification to learning through a \emph{verifier's rule}: the ease of training AI to solve a task is proportional to how verifiable the task is.

\begin{figure}[t]
\centering
\includegraphics[width=\linewidth]{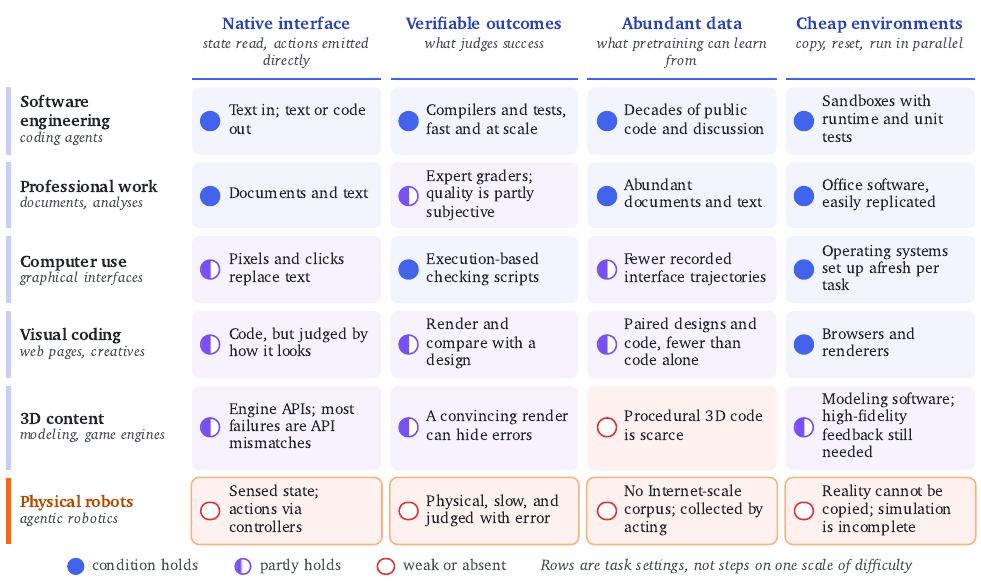}
\caption{\textbf{The four enabling conditions of digital agents, and how far each domain satisfies them.} The domains illustrate how the conditions vary across task settings rather than forming a strict progression of capability or difficulty.}
\label{fig:digital-conditions}
\end{figure}

Because these conditions are properties of a task setting rather than of programming per se, capability has spread to settings in which they could be established. These settings include software engineering \citep{jimenez2024swebench} and professional work across many occupations \citep{patwardhan2025gdpval}. They also include computer use, in which agents operate graphical interfaces in real operating systems \citep{xie2024osworld}, and visual coding, in which generated code is rendered and compared with a design or specification \citep{si2024design2code,xu2026commercevibe}. Most recently, they include 3D content, in which agents write programs for modeling software and game engines \citep{gao2026threedcodebench,ye2026code4scene}.
Most of these extensions weaken some of the conditions, and capability lags where they
weaken. Computer use weakens the native interface by replacing text with pixels and
clicks. Its time horizons are 40 to 100 times shorter than those in software and reasoning
domains, although they are increasing at similar rates \citep{kwa2025domains}. In 3D
content, verification weakens, because a convincing render can hide incorrect spatial
relations \citep{ye2026code4scene}. Progress there depends on execution environments that
return high-fidelity feedback and on more procedural data \citep{gao2026threedcodebench}.
Figure~\ref{fig:digital-conditions} summarizes our assessment for each domain. The domains differ in inputs, outputs, and verification requirements and do not form a strict progression of capability or difficulty.
Physical robots weaken all four conditions at once, as the next subsection shows.

\subsection{Which Assumptions Fail in the Physical World}
\label{sec:digital-different}
\label{sec:digital-cannot}

Each enabling condition enters digital-agent design as an assumption. This subsection asks which
of these assumptions, and which further ones, fail in the physical world, and what problem
each failure creates.
Table~\ref{tab:digital-physical} contrasts typical sandboxed coding agents, the setting in
which digital agents are most mature, with physical agents along eleven aspects, grouped by
the four enabling conditions and by the constraints that the physical world adds. Many of
the differences trace back to two abilities that a coding sandbox provides and the physical
world does not: reading the state of the world and judging the outcome of an action
\citep{wang2026thea}. Not every digital task offers these abilities either: computer-use and
web agents also face environments that change, hide state, or cannot be rolled back, and
their limits foreshadow the physical setting.

\begin{table}[t]
\centering\small
\caption{\textbf{Typical coding agents and physical agents compared.} Rows are grouped by the four enabling conditions and by the constraints that the physical world adds.}
\label{tab:digital-physical}
\begin{tabular}{@{}L{0.15\linewidth}L{0.36\linewidth}L{0.41\linewidth}@{}}
\toprule & Coding agent & Physical agent \\ \midrule
Environment & A designed artifact: files, code, APIs & Objects, people, and spaces \\ \midrule
\rowcolor{STwoColor!7}\multicolumn{3}{@{}l}{\textbf{Native interface}} \\ \midrule
Input and state & Directly readable files, logs, and tool outputs & Partial, noisy, occluded,
and delayed sensing; contact and force are invisible to cameras; state must be
constructed \\ \addlinespace
Action interface & Tool calls, file edits, commands, tests & Skills, trajectories, and actuators \\ \midrule
\rowcolor{STwoColor!7}\multicolumn{3}{@{}l}{\textbf{Verifiable outcomes}} \\ \midrule
Verification & Fast, reproducible automated tests & Costly, slow, hard to reproduce \\ \midrule
\rowcolor{STwoColor!7}\multicolumn{3}{@{}l}{\textbf{Abundant data}} \\ \midrule
Data & Decades of public code and text; trajectories generated in sandboxes & No
Internet-scale corpus; collected by operating robots; force and touch data scarcest \\ \midrule
\rowcolor{STwoColor!7}\multicolumn{3}{@{}l}{\textbf{Cheap environments}} \\ \midrule
Training environment & Sandboxes that can be copied, reset, and run in parallel &
Simulators faithful for rigid-body motion but not for all contact physics; real resets
need time and often a person \\ \midrule
\rowcolor{STwoColor!7}\multicolumn{3}{@{}l}{\textbf{Constraints added by the physical world}} \\ \midrule
Loop & Call a tool, then check the result & Continuous sensing and motion with online
correction \\ \addlinespace
Time & Can wait, pause, retry, or run faster & The world does not pause or speed up; low-level
control must stay real-time \\ \addlinespace
Failure and recovery & Version control, snapshots, and retries & Collisions and damage;
physical state cannot be rolled back \\ \addlinespace
Body and transfer & No body; easy reuse across environments & Bodies heat, wear, drift,
and break; new embodiments need re-adaptation and re-validation \\ \midrule
Main difficulty & Managing context, tools, dependencies, and progress over long tasks &
Acting reliably under partial observability and physical constraints \\
\bottomrule
\end{tabular}
\end{table}

\paragraph{State cannot be read directly, and actions do not take effect directly.}
Digital-agent design assumes that the agent reads the environment state, and acts on it, through documented interfaces. 
A coding sandbox offers such interfaces but a physical environment does not \citep{wang2026thea}.
Digital agents already falter when their environment is less transparent: they perform
worse on tasks without clear feedback loops \citep{kwa2025measuring}. 
For a robot, partial observation is the usual condition: it observes only a local, momentary
slice of the world through noisy sensors and must assemble and maintain a model of the rest
\citep{wang2026thea}. Physical outcomes also depend on quantities that cameras do not see,
such as mass, friction, compliance, and contact force. Tasks such as insertion are difficult
to complete robustly with vision alone \citep{chen2026univtac}. Touch carries much of this
information: people correct their grip from touch within tens of milliseconds
(Section~\ref{sec:s2-requirements}), and in simulated tasks a frontier multimodal model
interpreted vision-based tactile images when given them \citep{yang2026visuotactile}. Nor do actions take effect
directly: they are skills, trajectories, and actuator commands whose meaning is set by
controllers, reference frames, and units, and whose effects arrive with delay. The resulting
problem is one of grounding: state must be estimated, together with its source, age, and
uncertainty, and actions must carry explicit physical semantics.

\paragraph{Outcomes are not automatically verifiable.}
Digital-agent design usually assumes that outcomes can be checked automatically, quickly, and
reproducibly. Even with such checks, verification is imperfect: agents asked to evaluate
their own work tend to praise it confidently \citep{anthropic2026harnessdesign}, and
benchmark oracles can themselves be flawed \citep{openai2026swebenchverified}. Physical
outcomes must instead be observed, judged, and sometimes re-measured, and both false
successes and false failures are costly. A false success leaves a task
unfinished or a hazard in place, and VLM-based judges are biased toward this error when the visual evidence is ambiguous
\citep{navasardyan2026failbench}.
By the verifier's rule \citep{wei2025verifier}, this scarcity of credible verification is itself a central
obstacle to training physical agents. The resulting problem is to verify outcomes using evidence:
each verdict must state which of the four questions of the agentic loop it
answers (acceptance, termination, physical effect, or intent) and come from a calibrated verifier.

\paragraph{Data are not abundant.}
Digital-agent design assumes that pretraining has already covered the domain. Unlike natural
language and two-dimensional vision, dexterous manipulation has no Internet-scale corpus
\citep{hoque2025egodex}. Robot data are produced by operating robots, at the pace of the
physical world. The 76,000 trajectories of DROID, about 350 hours of interaction, took 50
data collectors twelve months on three continents \citep{khazatsky2024droid}. Teleoperation
of this kind lacks scale and diversity, and synthetic data bring a gap to reality
\citep{luo2025beingh0}. Human video scales more easily, but large collections such as Ego4D
\citep{grauman2022ego4d} lack native hand-pose annotations and action labels \citep{hoque2025egodex}, and video
records no contact forces. Touch is scarcest: acquiring large-scale and reliable
tactile data in the physical world remains costly \citep{chen2026univtac}. The resulting
problem is one of supply: abundant data must be manufactured by a data engine that turns
the operation of robots into training data.

\paragraph{Environments are not cheap to copy and reset.}
Digital-agent design assumes that environments can be copied, reset, and run in parallel.
Where simulation is faithful, the digital recipe returns: with thousands of simulated robots trained in parallel, a quadruped learns to walk on uneven terrain in about twenty minutes
\citep{rudin2021walk}. Where it is not, policies trained in simulation often fail to
transfer to the real world \citep{tan2018simtoreal}, and simulation pipelines have only
begun to cover complex physics such as deformable objects and fluids
\citep{qin2026embodiedsmith}. 
The alternative, practice in the real world, requires the scene to be reset after each attempt; this is typically handled by automated reset skills, with manual reset as a fallback when automatic reset fails \citep{li2026roboclaw}.
The cheap, parallel, and resettable environment of digital agents therefore exists only in part. The
resulting problem is twofold: every physical experiment has a cost that must be budgeted,
and results obtained in simulation must be validated against reality.

\paragraph{Assumptions beyond the Four Conditions.}
Three further assumptions of digital-agent design fail for reasons other than a missing
condition: the physical world adds constraints of time, irreversibility, and embodiment.
Web and computer-use agents meet milder forms of the first two, but the third has no close
digital counterpart. Embodiment amplifies the first two and makes
all three matters of safety and control.
\begin{itemize}[leftmargin=*, itemsep=2pt, topsep=3pt]
  \item \emph{Deliberation is not free.} While a robot deliberates, the world continues to evolve
        \citep{black2025rtc}. A robot that takes a minute may act on a scene that no longer exists, with
        consequences for both performance and safety.
  \item \emph{Errors cannot be undone.} Coding agents undo errors with version control,
        snapshots, and sandboxes \citep{anthropic2025harnesses}. Many physical actions cannot be undone \citep{wang2026thea}, and the
        world offers no general counterpart to restoring a software snapshot \citep{zhu2026language}.
        Safety must therefore be enforced before an action is taken, and below the model.
  \item \emph{Bodies do not stay fixed.} Coding agents have no body to manage \citep{wang2026thea}. A
        robot's capabilities depend on its body, and that body's condition changes
        with use. As \citet{sagar2026tear} note, over hours of operation motors gradually heat up,
        current saturates near contact, and voltage sags under load, so that the same command
        can produce a weaker, delayed, or noisier motion. Motor drivers reduce current near thermal limits and
        can leave closed-loop control beyond them \citep{odrive2026hardware}. Actuators can
        also switch off torque after overheating until they are rebooted
        \citep{robotis2026mx106}. In simulated LIBERO \citep{liu2023libero} tasks, one study perturbed only hardware and setup factors, such as camera mounting, calibration,
        actuation gain, control rate, and gripper pads. At the highest severity, the pooled success of
        the eight VLA models tested on all four suites fell from 93.2\% to 50.9\%
        \citep{cui2026liberorig}.
\end{itemize}

\subsection{What Transfers}
\label{sec:digital-transfer}

Beneath the practices that make digital agents work lie principles about how an agent
organizes its work, not about what its environment provides. Because they do not depend on
the enabling conditions, five such principles carry over to physical agents and already
shape our architecture.
\begin{itemize}[leftmargin=*, itemsep=2pt, topsep=3pt]
  \item \emph{Close the loop on evidence.} Agents are language models ``using tools based on
        environmental feedback in a loop'' and need ground truth from the environment at
        each step \citep{anthropic2024agents}. 
        For a robot, the principle requires the agent to observe and assess an action's relevant effects before treating its intended outcome as established or making a decision that depends on that outcome. The agent may continue planning or monitoring while an action is in progress, provided that subsequent actions respect the execution and safety constraints.
  \item \emph{Treat tools as contracts.} Tools are contracts between deterministic systems
        and non-deterministic agents \citep{anthropic2025tools}. Execution contracts
        extend this principle to physical calls. A
        capability contract, for example, states a skill's inputs and units,
        preconditions, expected effects, time limits, and behavior on cancellation, so that
        S2 can reason about a call before making it and interpret its result afterward.
  \item \emph{Manage context explicitly, and keep durable state outside it.} Context is a
        finite resource with diminishing returns, managed through compaction, structured
        notes, and sub-agents \citep{anthropic2025context}. Long-running agents also keep
        progress files \citep{anthropic2025harnesses} and a session log outside the model's
        context window \citep{anthropic2026managedagents}. The embodied harness applies
        the same discipline to the world, the user, and the body: it keeps task progress,
        maps, and memories outside the model and supplies them when a decision requires
        them.
  \item \emph{Separate doing from judging.} Long-running agents benefit from end-to-end tests,
        because models tend to mark work as complete without proper testing
        \citep{anthropic2025harnesses}. Separating the agent that performs the work from
        the agent that judges it also helps \citep{anthropic2026harnessdesign}. In the
        embodied harness, the verifier is the judge: it assesses each step and the
        task from evidence rather than from the model's claim that it has
        finished.
  \item \emph{Package capabilities as reusable skills.} Digital agents load packaged
        skills when a task needs them \citep{anthropic2025skills}. An agent in Minecraft accumulated a library of
        skills written as code \citep{wang2024voyager}. For robots, S1 skills and the
        programs that S2 writes play this role (Figure~\ref{fig:exec-paths}): a capability
        validated once can be invoked many times, and reuse also reduces the number of
        decisions that S2 must make at run time.
\end{itemize}
These principles transfer intact, but their digital implementations relied on the enabling
conditions. The next subsection therefore turns to what physical agents must rebuild.

\subsection{What Must Be Rebuilt}
\label{sec:digital-add}

\begin{figure}[t]
\centering
\includegraphics[width=\linewidth]{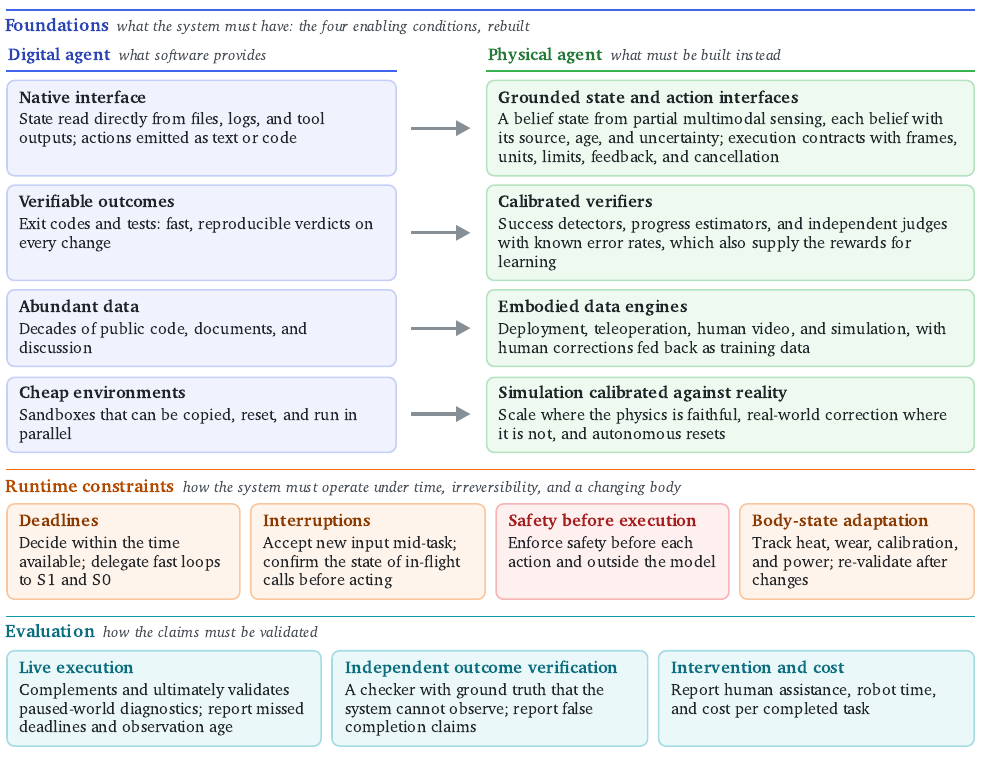}
\caption{\textbf{What physical agents must rebuild.} 
Foundations replace the four enabling conditions of Figure~\ref{fig:digital-conditions}; runtime constraints, which follow from time, irreversibility, and a changing body, govern how the system operates; and evaluation in the running world validates its claims.}
\label{fig:digital-add}
\end{figure}

Where an enabling condition fails, a physical agent must rebuild it. Where the physical
world adds a constraint, the agent must operate within it. Success at either cannot be
assumed, so claims about the agent must be validated in the running world.
Figure~\ref{fig:digital-add} organizes these requirements in three layers: the foundations
that the system must have, the constraints under which it must operate, and the evaluation
that validates its claims.

\paragraph{Foundations.}
Four foundations replace the four enabling conditions.
\begin{itemize}[leftmargin=*, itemsep=2pt, topsep=3pt]
  \item \emph{Grounded state and action interfaces in place of a native interface.} On the
        state side, the harness maintains a belief state: a persistent and revisable
        estimate of world, user, and body state. The estimate is built from partial
        multimodal sensing, including touch and force when a task depends on contact. On the action
        side, each call passes through an execution contract. The contract states the call's
        reference frame, units, preconditions, capability limits, and expected effects. It
        defines the feedback reported during execution and the behavior on cancellation,
        and it enforces safety checks, permissions, and postconditions that
        the model cannot bypass. Because a physical call can be executed exactly as
        commanded and still fail to produce its effect, the contract separates the
        termination of a call from the verification of its effect.
  \item \emph{Calibrated verifiers in place of exit codes.} Success detectors, progress
        estimators, and independent judges with known error rates make physical outcomes
        checkable. They serve both the loop at run time and the rewards for learning, so
        the quality of verification can become a bottleneck for how far a physical agent improves
        \citep{wan2026nofreechecker,zhou2026verifine}.
  \item \emph{Embodied data engines in place of Internet-scale corpora.} Physical agents
        must produce their own data from teleoperation, human video, and simulation. Above
        all, they must produce it from deployment, in which autonomous attempts and human
        corrections across a fleet become training data \citep{wang2026lwd}.
  \item \emph{Simulation calibrated against reality in place of sandboxes.} Simulators and
        digital twins \citep{todorov2012mujoco} cover part of the need but not the deployed
        world \citep{wang2026thea}. They provide scale where their
        physics is faithful, and real-world experience corrects them where it is not.
        Autonomous resets make practice in the real world more affordable
        \citep{li2026roboclaw}.
\end{itemize}

\paragraph{Runtime Constraints.}
The three constraints that the physical world adds govern how the system operates: time imposes deadlines and
interruptions, irreversibility requires safety before execution, and a changing body
requires adaptation.
\begin{itemize}[leftmargin=*, itemsep=2pt, topsep=3pt]
  \item \emph{Deadlines.} The agent treats latency as a hazard, decides within deadlines
        rather than deliberating at length, and delegates fast loops to S1 and S0. This division of labor mirrors the separation
        of fast control from slow reasoning in Helix \citep{figure2025helix} and of fast
        interaction from background reasoning in Thinking Machines' interaction models \citep{thinkingmachines2026interaction}.
  \item \emph{Interruptions.} The agent accepts new input while it reasons or acts. Before
        dispatching a conflicting action, it confirms the state of the calls in flight and
        of the body, because a cancellation request alone does not establish that
        execution has stopped.
  \item \emph{Safety before execution.} Because many actions cannot be undone, the agent
        anticipates consequences, prefers reversible or information-gathering actions under
        uncertainty, and asks for help or refuses when needed.
  \item \emph{Body-state adaptation.} The agent tracks heat, wear, and power,
        and adapts to them, for example by conditioning its actions on actuator telemetry
        \citep{sagar2026tear}. It re-validates its skills after maintenance or a change of
        embodiment.
\end{itemize}

\paragraph{Evaluation.}
Claims about a physical agent must be validated under the conditions in which it operates.
\begin{itemize}[leftmargin=*, itemsep=2pt, topsep=3pt]
  \item \emph{Live execution.} Paused-world evaluations remain useful diagnostics because
        they separate the quality of a decision from its timing. Paused and live execution,
        however, measure different abilities, so live evaluation complements, and ultimately
        validates, paused-world results.
  \item \emph{Independent outcome verification.} Success should be decided by a checker with
        access to ground truth that the system cannot observe.
  \item \emph{Intervention and cost.} Each result should report the human assistance and
        interventions it required, the robot's time, and the cost per completed task.
\end{itemize}

These requirements distinguish Embodied S2 from a digital agent: they reshape both the capabilities required of S2 and the conditions under which those capabilities must be evaluated.
Progress on digital-agent benchmarks, however rapid, therefore does not translate directly into the ability to delegate physical work; that ability must be measured and built in the physical world.

\section{Road Ahead}
\label{sec:road-ahead}

Agentic robotics broadens the question by which progress in robotics is judged: from what a
robot can do on command to what it can be entrusted to do on its own. General foundation
models already supply much of the knowledge and reasoning that such trust requires. Turning
these capabilities into dependable physical work, however, raises problems of timing,
persistent state, verification, adaptation, and learning from experience that no single
model release will resolve. Where the field goes next will therefore depend on how it
measures progress and which of these problems it addresses. This section proposes a ladder
of delegable work as a measure of progress and places current systems on it. It then
identifies research directions that climbing the ladder requires and closes with open
questions about how agentic robotic systems should be built.

\subsection{Ladder of Delegable Work}
\label{sec:ladder}

\begin{figure}[htbp]
\centering
\colorlet{rungA}{SOneColor!85}
\colorlet{rungB}{SOneColor!40!STwoColor!85}
\colorlet{rungC}{STwoColor!70!FMColor!85}
\colorlet{rungD}{FMColor!85}
\begin{tikzpicture}[
  font=\footnotesize,
  execute at begin node={\hyphenpenalty=10000\relax},
  >={Latex[length=2mm]},
  lvl/.style={font=\footnotesize\bfseries},
  ex/.style={font=\scriptsize\itshape, black!65, align=center, text width=3.3cm},
  dir/.style={draw=#1!75!black, fill=#1!9, rounded corners=3pt, line width=0.6pt,
    minimum height=0.6cm, align=left, inner sep=4pt, font=\scriptsize},
]
\foreach \i/\id/\name/\dur/\ex/\col in {0/L1/Primitive/{sub-second to seconds}/{reach a handle,\\pick up a cup}/rungA, 1/L2/Task/{minutes}/{make a cup of coffee and bring it to the user}/rungB, 2/L3/Workflow/{hours}/{prepare breakfast for four, all ready on time}/rungC, 3/L4/Role/{days and longer}/{run the household's daily breakfast}/rungD}{
  \pgfmathsetmacro{\x}{\i*3.5}
  \pgfmathsetmacro{\y}{\i*0.7}
  \fill[\col, rounded corners=1.5pt] (\x+0.04,0) rectangle (\x+3.46,\y+0.85);
  \node[lvl, anchor=south, text=\col!70!black] at (\x+1.75,\y+0.88) {\id\ \name};
  \node[anchor=north, font=\scriptsize\bfseries, text=white, align=center] at (\x+1.75,\y+0.8) {\dur};
  \node[ex, anchor=south] at (\x+1.75,\y+1.25) {\ex};
}
\draw[->, thick, black!55] (0,-0.3) -- (14.3,-0.3) node[below left, font=\scriptsize, align=right]
  {Responsibility a robot can be trusted to carry alone};
\foreach \i/\acc in {0/{the result of\\one operation}, 1/{delivery of\\one task}, 2/{the coordinated result\\of several tasks}, 3/{sustained performance\\of a standing role}}{
  \pgfmathsetmacro{\x}{\i*3.5}
  \node[font=\scriptsize, align=center, text width=3.3cm, anchor=north] at (\x+1.75,-0.8) {\acc};
}
\node[font=\scriptsize\bfseries, anchor=east, align=right] at (0.1,-1.1) {Responsibility};
\end{tikzpicture}
\caption{\textbf{Ladder of delegable work.} Progress is measured by the responsibility that a robot can be
trusted to carry alone. Each of the four levels includes those below it; typical durations are illustrative.}
\label{fig:roadmap}
\end{figure}
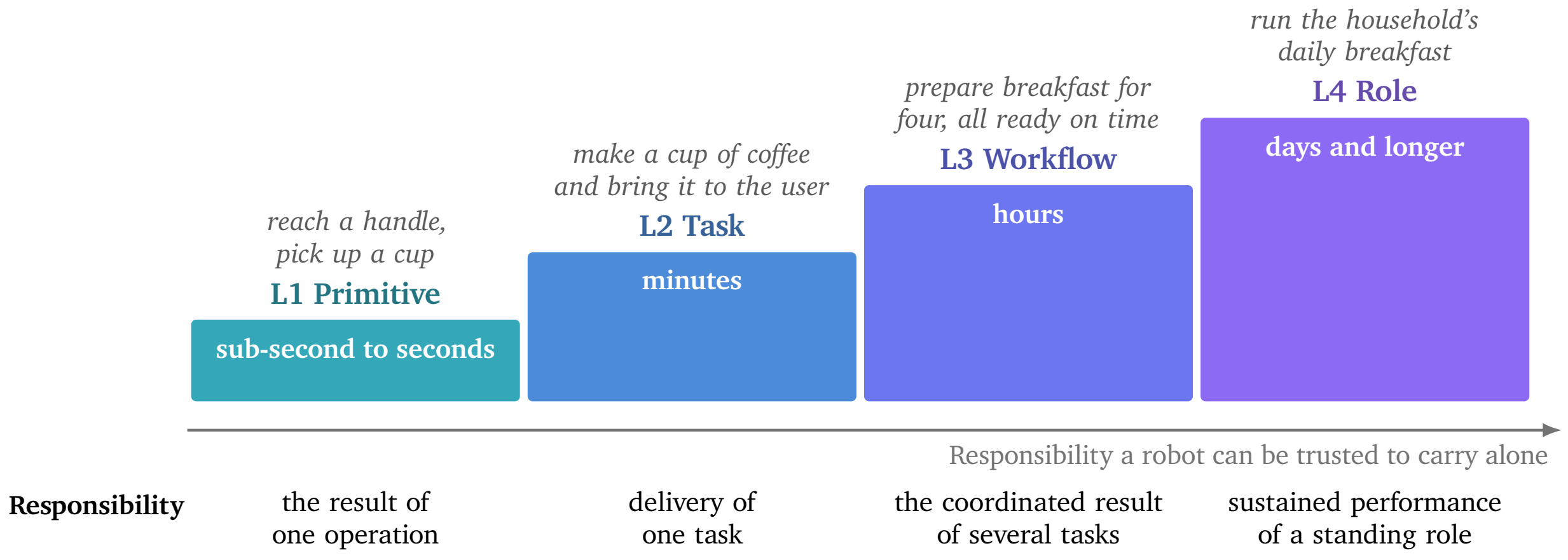

Reaching embodied intelligence that is genuinely useful calls for a high-level ladder that
guides progress toward it. We propose such a ladder for agentic
robotics: it measures progress by the scope of responsibility that a robot can reliably
carry alone, independently of its body, domain, or implementation. The ladder has four
levels (Figure~\ref{fig:roadmap}).
At L1, the \emph{primitive} level, the robot executes a local skill, such as picking up a cup, and checks whether the intended result has been achieved.
At L2, the \emph{task} level, the robot coordinates skills to deliver a complete task, such as making and serving coffee: Embodied S2 decomposes the goal, selects actions, monitors execution, and responds to failures until the acceptance criteria are met.
At L3, the \emph{workflow} level, the robot coordinates interdependent tasks toward an overall goal, such as preparing breakfast for four: Embodied S2 manages timing, priorities, and shared resources, and revises its plan as needed.
At L4, the \emph{role} level, the robot sustains an agreed responsibility across successive tasks and workflows, such as managing daily breakfast around household schedules and preferences. 
This responsibility continues after individual workflows end, until it expires or is withdrawn.
Each level builds on those below it, extending the scope over which the robot must connect reasoning, execution, and verification.

\begin{figure}[htbp]
\centering
\includegraphics[width=0.92\linewidth]{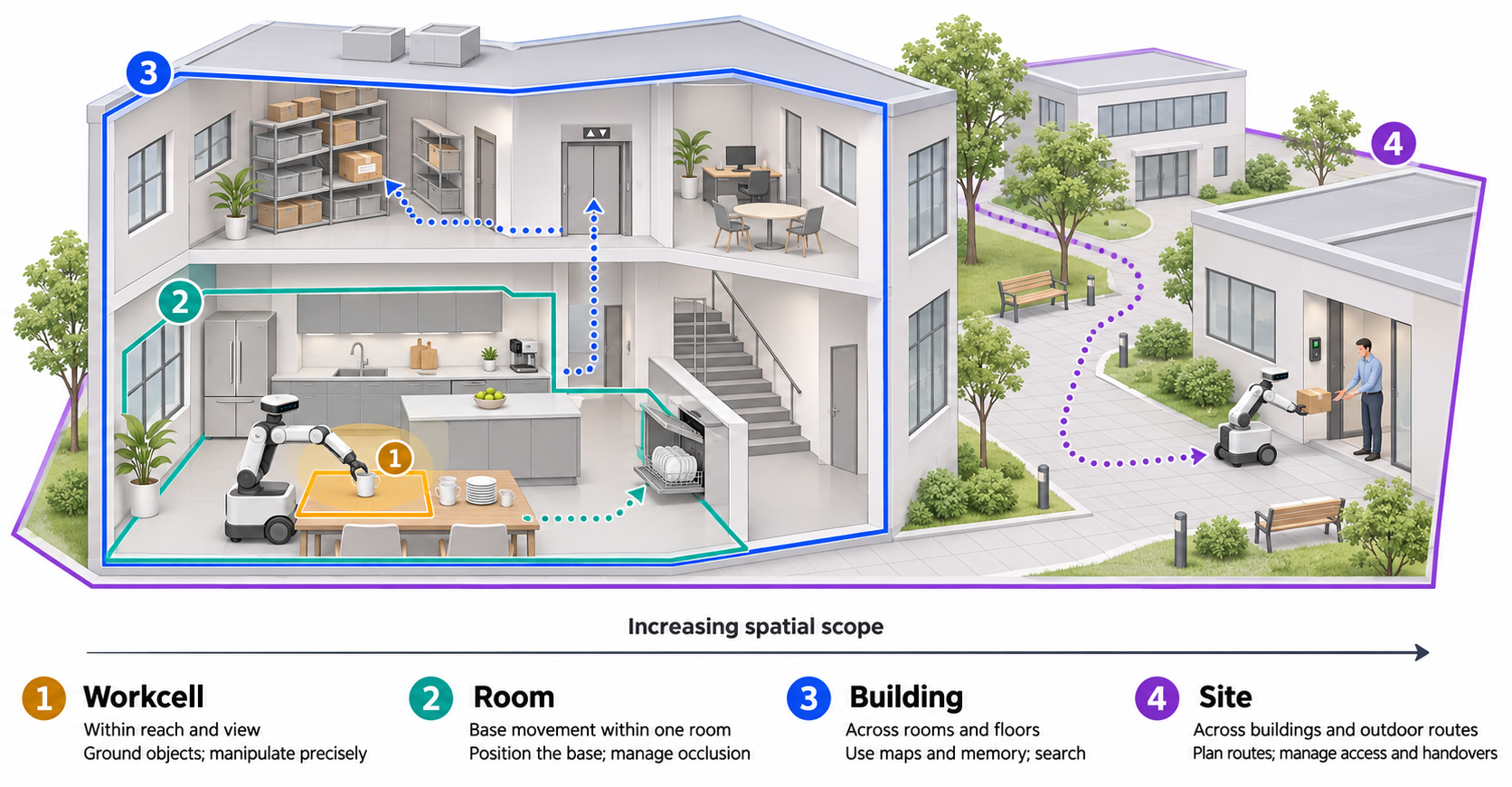}
\caption{\textbf{Spatial scope of delegated work, from a workcell within the robot's reach to a site with several buildings and outdoor routes.} The illustration is conceptual.}
\label{fig:spatial-scope}
\end{figure}

Spatial scope and complexity complement the ladder: within each level, progress means reliably fulfilling the delegated responsibility across broader environments and more demanding activities.
As the operating scope expands from a workcell to a room, a building, or a site (Figure~\ref{fig:spatial-scope} and Table~\ref{tab:spatial-scope}), robots must integrate observations across locations, remember relevant objects and events, and seek information beyond their immediate view. 
These demands, like those of complex activities, fall on Embodied S2, which maintains goals and context while guiding action through the loop of observation, reasoning, execution, and verification.
We therefore expect the value of agentic robotics to grow with scope and complexity.

For current systems, they can execute primitives and complete tasks in unstructured environments, but have only begun to reach the workflow level, and have yet to reach the role level.
\begin{itemize}
\item \emph{Primitives and tasks.} Reliability on narrow tasks is rising. Sunday reports 99.1\% success in folding laundry in unseen homes with a single model checkpoint and no adaptation per home \citep{sunday2026act2}. Dyna Robotics reports that Dyna-2 passed customer acceptance criteria at a rate of 87\% at deployment sites whose data it had not seen \citep{dyna2026dyna2}. Longer tasks are now completed too: $\pi_{0.5}$ performs long-horizon cleaning in unseen homes \citep{pi2025pi05}, and Figure AI reports a four-minute task of unloading and reloading a dishwasher \citep{figure2026helix02}.
\item \emph{Workflows.} The first workflow-level results are single-company demonstrations, such as the hour-long laundry-room workflow that Dyna Robotics reports for Dyna-2.1 \citep{dyna2026dyna21}. In simulation, a mobile manipulator with persistent memory has chained tasks for more than three hours \citep{banwasi2026messymem}.
\item \emph{Roles.} To our knowledge, no robot system in an unstructured environment has been reported to hold a role without human repair.
In structured settings, an autonomous materials laboratory operated for 17 days \citep{szymanski2023autonomous}, although the novelty of its products was later disputed \citep{leeman2024challenges,szymanski2026author}; sustained operation alone does not show that a system holds a role. Capabilities that a role relies on have begun to appear as harness prototypes that keep state outside the model. Some keep notes about a house that persist across tours and raise navigation success \citep{zhao2026navharness} or learn properties such as ``locked'' through interaction \citep{banwasi2026messymem}. Others carry user preferences across a program restart on a real robot \citep{wang2026carryon} or maintain a revisable model of the robot's own body \citep{lou2026knowbody}. Remembering such information, however, is not the same as organizing work around it. It remains to be shown that a robot can use these capabilities to manage goals across tasks and sessions and to sustain a role. %
\end{itemize}

\begin{table}[!tbp]
\centering\small
\caption{\textbf{Spatial scope of delegated work.} A claim of reaching a level should state the scope, duration, and complexity of the work.}
\label{tab:spatial-scope}
\begin{tabular}{@{}L{0.10\linewidth}L{0.35\linewidth}L{0.45\linewidth}@{}}
\toprule Scope & Extent & What it adds for S2 \\ \midrule
Workcell & Within reach and view & Grounding objects and actions in view; contact and precision \\ \addlinespace
Room & Within view, but the base must move & Where to place the base; reachability that changes as the robot moves \\ \addlinespace
Building & Beyond what the robot can observe at once & Maps and memory of places and objects; search and exploration \\ \addlinespace
Site & Several buildings and outdoor routes & Long routes; other people and robots; handover between areas \\
\bottomrule
\end{tabular}
\end{table}

\subsection{Future Research Directions}
\label{sec:directions}

Climbing the ladder, and doing so across larger spaces and longer periods, will require
progress well beyond what current systems demonstrate. The evidence reviewed in this paper
points to four directions that need further research. They concern embodied reasoning
that meets physical deadlines, control whose state and outcomes remain
trustworthy across tasks and spaces, adaptation to new environments and bodies at low cost,
and continual improvement from physical experience. The list is not exhaustive.

\subsubsection{Embodied Reasoning Under Physical Constraints}
S2 must understand the world, anticipate the consequences of its actions, and decide
within limited time and computation, yet current S2 foundation models fall short in both
speed and decision quality. With a frontier model as its S2 foundation model, S2 decides once every
several to forty seconds \citep{xu2026rtsafe,yang2026astranomy,galbot2026systematically},
far slower than the S1 skills it directs (Figure~\ref{fig:latency}). More deliberation
helps when the world waits \citep{gemini2025robotics15} but hurts when it moves \citep{xu2026rtsafe}. Decisions remain
weak where embodiment matters most: judging success from ambiguous evidence
\citep{navasardyan2026failbench}, anticipating the consequences of actions
\citep{sun2026roboharm}, and deciding when to stop exploring \citep{liu2026roboquest}.

We therefore advocate making capped scaling an explicit
training and system objective: maximizing decision quality subject to a response deadline,
evaluated without pausing the world and reported as success--latency--cost curves. Three
questions follow. First, when is further reasoning worth its time, and when should the
robot observe again or act at once? Second, which decisions must be made online, and which
can be reused or turned into skills? Partial answers to both include routing
each subtask to a planner of appropriate strength \citep{dao2026direct} and launching
chains of primitives with one call \citep{xu2026chunkact}. A fast policy can also consult a
frontier model only for recovery \citep{sun2026call}, and authoring skills offline removes
the slow model from the loop \citep{bai2026compile}. Third, how can the task benefit per unit of time and computation be
raised? Distillation into small models that run locally is one candidate \citep{izzo2026multimodal,liu2026guava,baumann2026localnav,liauto2026mevlm,liu2026guava}.

\subsubsection{Persistent and Verifiable Control Across Tasks and Spaces}
As responsibility extends to workflows and roles, and the operating scope to buildings and sites, the system must keep its state trustworthy while organizing and verifying the work.
However, current harnesses are built and evaluated almost entirely on single tasks in a workcell or a room (Section~\ref{sec:op-eval}).

Three questions stand out.
The first is how to represent and maintain state that spans rooms, floors, and days, so that the robot knows what it still knows.
Candidates include augmenting spatially grounded scene graphs with outcomes learned through interaction \citep{banwasi2026messymem} and recording a subgoal as complete only after evidence of completion \citep{zhao2026eventsync}.
The second is how to coordinate the interdependent tasks of a workflow, which share information, resources, and deadlines.
The third is how to divide judgment between the model and the execution system: which checks belong inside skills and verifiers, and which decisions must return to S2.
Such a division can also keep a slow decision-maker compatible with continuous control. A precedent exists one level down, within S1: real-time chunking lets a VLA generate the next action chunk while the current one executes, remaining robust to inference delays of more than 300\,ms \citep{black2025rtc}. S2 decisions, however, take seconds rather than milliseconds.

\subsubsection{Efficient Adaptation Across Environments and Embodiments}

Deployment demands that capabilities transfer to new sites and embodiments with minimal configuration, data, and human effort. Yet current success remains highly dependent on the specific embodiment (Section~\ref{sec:eval-cross}).
We advocate three lines of work. 
The first is reuse of knowledge and experience: in-context adaptation and skill libraries of validated repairs reduce what must be learned anew \citep{lu2026aspire}, and compiled skills can be frozen and carried to other agents \citep{bai2026compile}.
The second is capability interfaces, such as capability cards and device descriptions, that let S2 plan for a body it has not used before.
The third is skill adaptation: in one study, a single policy, conditioned on examples of commanded actions and resulting motions, adapted in context to unseen controller modes and gains \citep{yang2026crosscontroller}.

Fitting a system to a body also includes choosing loop rates and the placement of computation, as deployed stacks do. Helix~02 runs whole-body control at 1\,kHz and a visuomotor policy at 200\,Hz \citep{figure2026helix02}, beneath an upper module that ran at 7 to 9\,Hz in the first Helix \citep{figure2025helix}. 
An on-robot model avoids the network round trips and outages that a cloud model must tolerate.
Conversely, skills, sensory feedback, and the body should be designed so that S2 can plan, execute, and verify more easily.
Examples include skills that report their terminal states with evidence, and touch sensing: adding tactile observations to the in-context demonstrations of
an embodied task agent raised its average success on four simulated tasks from 15.75\% to 24.00\% \citep{wang2026tactileeta}.

\subsubsection{Continual Improvement from Physical Experience}
\label{sec:rsi}

A robot holding a role for weeks keeps meeting new tasks and changes in its environment and body that no training run anticipates. It must therefore learn from physical experience which
object properties, user preferences, causes of failure, and skills are worth keeping. We distinguish three kinds of change. \emph{Adaptation} adjusts
behavior to the current task, environment, or feedback. \emph{Continual improvement}
retains a change to memory, tools, skills, the harness, or a policy, and verifies that it
improves later work. \emph{Recursive self-improvement} (RSI) improves the process that
produces such changes, making subsequent improvement more effective.

Robots have long improved through autonomous data collection, learned rewards, and reinforcement learning, and DrEureka used language models to design reward functions and domain randomization for sim-to-real transfer \citep{ma2024dreureka}. What recent agentic systems add, mostly around frozen foundation models, is a loop that diagnoses failures from operating experience and revises the system, usually with a foundation model proposing the revisions \citep{tian2026physevo,lee2026robotuse,zhang2026robocop,deng2026dynaharness}. RegenHarness specifies such a loop as an evidence-gated protocol \citep{wang2026regenharness}.
This loop starts with attribution: deciding whether a failure came from perception, planning, a skill, or an interface, which part should change, and why the change should help future tasks.
Mechanisms differ in what they change.
Some carry lessons forward in context: retrospective, cross-stage failure attribution let GPT-6 Astra learn manipulation skills on a real Franka arm without demonstrations or parameter updates \citep{yu2026retrace}, and LEXP reuses lessons extracted from execution evidence \citep{zeng2026lexp}. 
Others change tools, skills, and harness code around a frozen model \citep{lu2026aspire,tian2026physevo,deng2026dynaharness}. 
For example, RAX~\citep{huang2026rax} converts successful online exploration into reusable programs for perception, planning, and control.
Still others change weights, adopting a fine-tuned policy only after verification \citep{zhang2026robocop}. 
Since every physical trial costs robot time and may be irreversible, a system must also decide which experience to acquire next, a question current methods rarely address explicitly.

A change counts as an improvement only if it helps later work and preserves existing ability; it must also be retired once outdated. 
Helping later work depends on practice tasks that grow harder and can be verified automatically.
Constructing simulated practice tasks from capabilities identified in offline data \citep{wang2026reconstruct} and practicing in
self-built simulated twins \citep{google2026argonrobotics} are first steps, although, to our
knowledge, progressive difficulty has not yet been demonstrated physically.
Preserving ability requires an adoption rule: in one study, accepting every proposed revision degraded a harness, whereas a champion--challenger rule improved held-out results \citep{wei2026scale}.
Retirement matters because a replaced gripper can invalidate learned behavior
\citep{luo2026toward}.

Gains do not yet accumulate reliably.
In EmbodiedCLBench, experience from basic tasks improved advanced ones, but the improvement did not grow with
more learning samples \citep{huang2026embodiedclbench}. With equal total development time, repeated short attempts matched or outperformed longer learning sessions in some model--task combinations; which did better depended on whether a model reached its first success within one short attempt \citep{wang2026sear}. In a 123-round study of agentic skill discovery in
simulation, accepted changes kept passing their tests while the target task never
succeeded: ``the number of accepted changes measures activity, not progress''
\citep{wang2026what}. A credible claim of improvement should therefore report gains on held-out tasks and show no regression on earlier tasks. It should also separate the contributions of the S2 foundation model, the harness, and S1, under the controls of Section~\ref{sec:eval-how} and within applicable safety requirements.

Claiming RSI further requires showing that an improved procedure for selecting practice, proposing
changes, and verifying them makes later improvement more effective, including on tasks
beyond those used to develop it. A survey of physical RSI reserves its
highest level for ``independently verified recursive gain'' and concludes that ``the
surveyed literature does not yet demonstrate Physical RSI as defined by our framework''
\citep{zhou2026who}. Demonstrating RSI requires reliable feedback from physical trials,
whose success detection is itself error-prone \citep{navasardyan2026failbench}, and causal
attribution of gains to the improved procedure rather than to more attempts.

\paragraph{Dependence among the Directions.}
The four directions depend on one another. Decisions within deadlines rely on state and
feedback that remain trustworthy across tasks and spaces. Reasoning and control reach new
environments and bodies only through adaptation, which relies on clear capability
interfaces. Continual improvement and recursive self-improvement rely on all three: a system
should retain only the changes that verified feedback shows to be better, and a retained change must pay off in the complete system
rather than in an isolated component. Conversely, each retained improvement can make
decisions faster, state more reliable, and adaptation cheaper. Progress in the other three
directions thus accumulates through the fourth.

\section{Conclusion}
\label{sec:conclusion}

Agentic robotics organizes capabilities around user intent through a closed loop of reasoning, action, verification, and adaptation. Our reference framework distinguishes three complementary functional roles: System~2 (S2) for goal-directed reasoning and decision-making, System~1 (S1) for executable skills, and System~0 (S0) for low-level control. S2 can invoke S1 skills or issue commands directly to S0. We refer to its realization through a foundation model and an embodied harness as \emph{Embodied S2}, the decision-making core that connects general-purpose intelligence with embodied execution.

Four ideas underpin this perspective. First, foundation models make S2 a promising starting point for robot generalization. Their broad knowledge and reasoning capabilities can help robots understand unfamiliar goals and situations, organize available capabilities, and adapt through feedback. This offers a complementary path to improving action policies through larger and more diverse training datasets. Second, physical autonomy requires the model and embodied harness to jointly support effective, verifiable action under constraints of time, embodiment, uncertainty, and safety. These constraints must shape how systems reason, execute, and adapt. Third, meaningful evaluation must distinguish model capabilities from closed-loop system performance while examining how those capabilities contribute to reliable outcomes. Finally, the long-term objective is sustained autonomy with progressively broader responsibility, extending from individual primitives and tasks to coordinated workflows and standing roles.

Realizing this ambition requires robots to track goals and progress over time, coordinate interdependent activities, recover from failures, and adapt to changing environments and embodiments. Improvement from physical experience must carry over to subsequent work while preserving existing capabilities. The evidence reviewed here does not yet establish reliable autonomy across these broader and more persistent responsibilities, making sustained performance a central target for both system development and evaluation.

This work calls for three community priorities. First, develop reliable evaluation frameworks that complement simulation with real-robot testing and extend task coverage to higher levels of responsibility and broader spatial scopes. Evaluation should assess sustained performance under changing conditions, including recovery from failure. Second, improve reasoning efficiency for time-sensitive, safety-critical, and highly dynamic settings. When test-time scaling relies on longer deliberation, gains in reasoning quality may come at the cost of timely action. Research should therefore prioritize better decisions within the time available for action, with the longer-term goal of developing embodied-native reasoning that integrates perception, reasoning, and action in response to the dynamics and urgency of physical tasks. Third, standardize reporting of the model, harness, available capabilities, evaluation conditions, inference budgets, response times, and human assistance. Clear reporting supports reproducible comparisons and helps distinguish model capabilities from the support provided by the surrounding system. Together, these priorities can help turn promising results into credible progress toward sustained physical autonomy.

We envision a new era of robotic systems that understand human intent, adapt to changing circumstances, and reliably take on broader responsibilities over longer time horizons.
Agentic robotics offers a system-level path toward this vision, bringing the general-purpose capabilities of foundation models into systems that people can trust with sustained work in the physical world.

\clearpage
\appendix
\section{Related Work}
\label{sec:related-work}

This appendix reviews the work on which our reference architecture, evaluation framework,
and roadmap build. Each subsection ends with our position on that work.

\subsection{Surveys and Position Papers}
\label{sec:rw-surveys}

\paragraph{Foundation Models and Robotics.}
\citet{bommasani2021opportunities} coined the term foundation models and discussed
robotics among their capabilities. Subsequent surveys organized robotic uses of
foundation models by component (perception, decision-making, and control)
\citep{firoozi2025foundation,hu2023toward} and reviewed real-world
applications in which foundation models replace specific modules of existing robot systems
\citep{kawaharazuka2024real}. Surveys of vision-language-action models place high-level
planners inside a VLA taxonomy, distinguishing monolithic from hierarchical designs
\citep{ma2026survey,shao2025large}. Others classify robot autonomy driven by language and
vision-language models by the model's role \citep{salimpour2025embodied}, or
argue that world models, including models of users, are central to embodied agents
\citep{fung2025embodied}. Our System~1/System~2 vocabulary comes from
cognitive science \citep{kahneman2011thinking} and was popularized in machine learning by
\citet{bengio2019system2}.

\paragraph{Definitions Closest to Ours.}
Several 2026 works define agentic robots by function rather than architecture.
\citet{wang2026survey} characterize agentic robotics by ``this
decision-making role of the FM, rather than any particular model or interface'', where FM
stands for foundation model. They code 2,343 systems by where foundation-model decisions
occur: in task orchestration, motion generation, or improvement. \citet{qin2026towards}
define agentic robots as systems in which agents autonomously plan and coordinate task
execution using observations and execution feedback, and organize their survey around
model, data, environment, and harness.
\citet{chen2026multimodal} characterize embodied agents by
five functions (perceive, anticipate, plan, act, verify) and find that only 5 of 64
benchmarks assess anticipation and only 3 assess verification. \citet{zhu2026language} separate model,
harness, and environment and conclude that action-interface expansion is documented more
convincingly than robust completion, recovery, authorization, or independent
verification. Harness-specific accounts include a survey that defines the embodied agentic
harness as the runtime control plane between an agent's proposals and embodied execution
\citep{huang2026harnesssurvey}. They also include a roadmap from world-action models to
embodied brains \citep{liang2026wams} and a proposal that robot middleware is the harness
layer \citep{lee2026harnessengineering}. Frameworks for self-improvement define ladders of
their own, for robots \citep{zhou2026who} and for large reasoning models \citep{yang2026scaling}. Table~\ref{tab:definitions} relates our
definition of Embodied S2 to the closest published definitions.

\begin{table}[!tbp]
\centering\small
\caption{\textbf{Closest definitions in the literature and how ours relates to them.}}
\label{tab:definitions}
\begin{tabular}{@{}L{0.17\linewidth}L{0.33\linewidth}L{0.19\linewidth}L{0.23\linewidth}@{}}
\toprule Work & What it says & What we adopt & What we add or change \\ \midrule
RoboBench \citep{luo2025robobench} &
``We refer to System~2 as the embodied brain, the cognitive core for decision-making in
manipulation.'' &
S2 as the decision-making core &
Embodied S2 includes the harness, not only a multimodal language model; criterion for when a component acts as S2 \\
From WAMs to Embodied Brains \citep{liang2026wams} &
The embodied brain is ``a model-level target'', while ``the harness manages memory,
routing, logging, verification, and capability calls''. &
Separation of model and harness &
Harness counted inside Embodied S2; S1 and S0 are tools that S2 calls \\
Thea \citep{wang2026thea} &
``The world, however, withholds two abilities that software grants for free: reading
the state of the world, and judging the outcome of an action.'' &
World, user, and body state built and maintained by the harness &
User context as a separate category; the harness, including its outcome judgments, counted as part of Embodied S2 \\
SCOPER \citep{wei2026scoper} &
``An embodied brain, or System~2, must turn an abstract goal into actions while
discovering relevant scene conditions and checking whether executed skills achieved
their intended effects.'' &
Choosing the next skill and judging the last one &
Harness treated as part of Embodied S2 rather than an add-on; persistence across episodes \\
Embodied agentic harnesses survey \citep{huang2026harnesssurvey} &
Harness as the ``runtime control plane'' that grounds observations, coordinates
capabilities, monitors outcomes, and manages intervention. &
Harness responsibilities and contracts &
Each ability divided between the model and the harness, each evaluated with the other held fixed \\
Agentic robotics survey \citep{wang2026survey} &
``The distinguishing feature is this decision-making role of the FM, rather than any
particular model or interface.'' (FM: foundation model) &
A decision-based criterion &
Adjudication of vendor ``S2'' labels; two evaluation levels; ladder of delegable work \\
Towards Agentic Robots \citep{qin2026towards} &
Agents ``autonomously plan and coordinate task execution using environmental
observations and execution feedback''; System~2 planners, System~1 visuomotor
policies, System~0 controllers. &
The S2/S1/S0 vocabulary &
Harness counted as part of Embodied S2; identifiability requirement \\
Harness Engineering for Physical AI \citep{lee2026harnessengineering} &
Robot middleware is the harness layer, mediating control, computing, and communication. &
The harness bounds what the model sees and does &
Different scope: our harness assembles context for S2's decisions and carries them out; the middleware described is infrastructure beneath it \\
\bottomrule
\end{tabular}
\end{table}

\paragraph{Our Position.}
Like these works, we define S2 by function: it interprets intent, decides in light of goals
and constraints, orchestrates skills and tools, and verifies outcomes to adapt later
decisions. We add a criterion that identifies S2 by its decisions on how to proceed and on
outcomes, and we count the embodied harness as part of Embodied S2. Unlike the self-improvement ladders, our four levels measure delegated
responsibility (Section~\ref{sec:ladder}), with stated conditions for each level claim
(Section~\ref{sec:eval-how}).

\subsection{Language-Model-Based Robot Agents}
\label{sec:rw-agents}

\paragraph{Early Systems (2022--2025).}
The basic S2 loop is not new. Early work used language models to decompose tasks into
steps \citep{huang2022language}. Later work grounded them in the affordances of learned
skills \citep{ahn2022saycan}, composed them with other pretrained models through language
\citep{zeng2023socratic}, and closed the loop with feedback from success detectors and
humans \citep{huang2022inner}. Code became an interface for composing perception and
control \citep{liang2022code,singh2023progprompt}. Language-model planners were coupled to
classical planners \citep{liu2023llmp} and geometric feasibility checks
\citep{lin2023text2motion}, grounded in 3D scene graphs \citep{rana2023sayplan}, taught when
to ask for help \citep{ren2023knowno}, and equipped to explain their failures
\citep{liu2023reflect}. Language models also wrote value maps and keypoint constraints for
motion solvers \citep{huang2023voxposer,huang2024rekep} and summarized user preferences
\citep{wu2023tidybot}. OK-Robot integrated open-vocabulary modules for mobile manipulation
in homes \citep{liu2024okrobot}. In Minecraft, a language-model agent followed an automatic
curriculum and built a skill library of code \citep{wang2024voyager}. In parallel, embodied
multimodal models and VLAs moved reasoning into the policy
\citep{driess2023palme,mu2023embodiedgpt,brohan2023rt2,gemini2025robotics}.

\paragraph{Agentic Robotic Systems of 2026.}
Frontier models that act through code and tools broadened the scope of this work.
Frameworks study coding agents as robot controllers across levels of abstraction
\citep{fu2026capx}, let agents write and call their own tools \citep{ge2026urai}, and
compose primitives and VLA policies in interactive code \citep{si2026pyrualean}. HexaShell runs
a thin harness with tools written and tested by optimizer agents \citep{gao2026hexashell};
a later version uses a small set of 14 general tools \citep{gao2026hexashellv3}.
Others maintain persistent playbooks \citep{lee2026robotuse}. General-purpose agents drive
physical robots directly \citep{jia2026agentaspolicy} and learn in context from
demonstrations \citep{cheng2026gptpolicy}. One study argues that, given the right
interface, coding and computer-use agents are in a sense already robot-control agents
\citep{hu2026via}. Orchestration frameworks connect agents to ROS~2 and to learned skills and policies
\citep{li2026roboclaw,cardenas2026rosclaw,galanti2026orchestration}, and other methods
reduce how often the expensive model is consulted \citep{sun2026call,xu2026chunkact}.

\paragraph{Our Position.}
These systems implement all or part of the S2 loop through layered orchestration, direct
control, or code as policy (Figure~\ref{fig:exec-paths}). We identify S2 by its decisions, not its
interface, and treat verification, persistent context, and revision as part of S2. From
these works we draw comparisons that fix the model or the harness, and we report cost
and time semantics alongside success.

\subsection{Hierarchical and Dual-System Robot Foundation Models}
\label{sec:rw-dual}

Many robot foundation models separate high-level reasoning from low-level action. Helix
pairs a slow vision-language model with a fast visuomotor policy that receives a latent
vector \citep{figure2025helix}, and Helix~02 adds a learned whole-body controller \citep{figure2026helix02}. GR00T~N1 couples a
vision-language module with a diffusion action module trained end to end
\citep{nvidia2025gr00t}. In Hi Robot, a high-level VLM chooses the next step for a low-level
policy \citep{shi2025hirobot}. The $\pi_0$ model follows commands from a high-level VLM
\citep{black2025pi0}, and $\pi_{0.5}$ co-trains subtask prediction with action generation
\citep{pi2025pi05}. Gemini Robotics~1.5 interleaves natural-language reasoning with actions
and is paired in an agentic system with an embodied reasoning model, Gemini Robotics-ER~1.5, that acts as orchestrator \citep{gemini2025robotics15}. Other
dual-system designs use value-guided sampling \citep{song2025hume}, combine a generalist
with a specialist policy \citep{bu2024towards}, or embed the fast system inside the slow
one \citep{chen2025fastinslow}. Still others switch adaptively between reasoning and acting
\citep{lin2025onetwovla} or communicate through 2D paths \citep{li2025hamster} and editable trajectory traces \citep{lee2025molmoact}. Unified models generate reasoning and actions with one decoder \citep{liu2026g05} or within one tightly coupled mixture-of-experts model \citep{zhai2025igniting}. Recent work adds world-model-guided
test-time search to subtask generation \citep{cai2026tau0} and studies design principles
for hierarchical VLAs \citep{hu2026whatmatters}. One study finds that the executor of a
hierarchical VLA underuses its plans \citep{xie2026fastplans}.

\paragraph{Our Position.}
We identify S2 by its decisions, not by its place in a hierarchy: a module acts as S2 only
if it makes identifiable decisions on how to proceed and on outcomes. An upper module that
emits only a latent representation may perform this function, but we do not
classify it as S2 unless replacement experiments reveal these decisions. The subtask stage
of a unified model, by contrast, makes an
identifiable decision on how to proceed and so implements part of S2. That executors can
underuse their plans suggests that S1's adherence to them should be measured.

\subsection{Embodied Reasoning Models}
\label{sec:rw-er}

A family of models aims to supply the embodied reasoning that S2 needs. The Gemini
Robotics-ER line introduced the ERQA benchmark and has evolved from a reasoning model into
an orchestrator of VLAs \citep{gemini2025robotics,gemini2025robotics15,google2026er2}.
Cosmos-Reason1 trains physical common sense and embodied reasoning with supervised
fine-tuning and reinforcement learning \citep{nvidia2025cosmosreason1}. Vesta combines a
generalist model with a multimodal memory harness \citep{bjorck2026vesta}. RoboBrain
models cover planning, affordance, and trajectory prediction
\citep{ji2025robobrain,baai2025robobrain2,tan2026robobrain25}. Embodied-R1 uses pointing as
an embodiment-agnostic interface, and Embodied-R1.5 adds a closed
planner--grounder--corrector loop \citep{yuan2025embodiedr1,yuan2026embodiedr15}. Other
models vary in scale and training recipe: RynnBrain
\citep{dang2026rynnbrain,li2026rynnbrain11}, PhysBrain \citep{deepcybo2026physbrain15},
MiMo-Embodied \citep{hao2025mimoembodied}, HY-Embodied \citep{tencent2026hyembodied}, Robix
\citep{fang2025robix}, ACE-Brain \citep{ace2026acebrain05}, Pelican-VL
\citep{zhang2025pelican}, and the MolmoER backbone of MolmoAct2 \citep{fang2026molmoact2}. Related spatial-grounding
models include SpatialVLM, RoboPoint, RoboRefer, and Magma
\citep{chen2024spatialvlm,yuan2024robopointx,zhou2025roborefer,yang2025magma}. Vlaser
studies how VLM initialization affects VLA fine-tuning \citep{yang2025vlaser}. Progress
estimation can be framed as temporal ordering of shuffled frames \citep{ma2025gvl}. Recent
models also train on data with execution feedback, goal revisions, and failure recovery
\citep{liauto2026mevlm,li2026joyinside}.

\paragraph{Our Position.}
Reports on these models mostly evaluate offline question answering, pointing, and
spatial skills. We treat such a model as a candidate S2 foundation model whose value for the
S2 loop must be shown at the system level.

\subsection{Agent Harnesses}
\label{sec:rw-harness}

\paragraph{Digital Precedents.}
ReAct interleaved reasoning and acting \citep{yao2023react}, Toolformer taught models to
call APIs \citep{schick2023toolformer}, and Reflexion stored verbal feedback in episodic
memory instead of updating weights \citep{shinn2023reflexion}. Generative agents
introduced memory streams with reflection and retrieval \citep{park2023generative}, and
CodeAct used executable Python as a unified action space \citep{wang2024codeact}.
SWE-agent showed that the agent--computer interface affects agent performance
\citep{yang2024sweagent}, and OpenHands provided an open sandboxed platform
\citep{wang2025openhands}. OSWorld benchmarked computer use \citep{xie2024osworld}.
Computer-use models and the Model Context Protocol gave agents general ways to reach software \citep{anthropic2024computeruse,anthropic2024mcp}. 
As a caution, automatic harness evolution can fail to outperform simple test-time scaling at a matched budget \citep{wang2026rethinking}.

\paragraph{Physical Harnesses.}
Thea adapts the coding-agent harness to robots, adding a scene graph as context and exit
codes that report action outcomes \citep{wang2026thea}. Runtime governance layers add
policy checks, admission, monitoring, rollback, and human override
\citep{qin2026governance}. PhyAgentOS separates execution termination from semantic
completion \citep{liu2026phyagentos}, and in one study of LLM-controlled devices, scoped capability
tokens rejected every capability-escalation call that JSON-schema validation let through
\citep{yang2026device}. Harnesses with
execution contracts and verification include DynaHarness, RegenHarness, SCOPER's harness,
and a harness for safe robot control
\citep{deng2026dynaharness,wang2026regenharness,wei2026scoper,xu2026safeharness}. Programs
serve as evolvable interfaces \citep{wang2026harnesspai}, critics run on separate
timescales \citep{ding2026zetta}, and frozen VLAs become retryable primitives
\citep{zhang2026harnessvla}. Navigation and persistence harnesses validate planner
proposals \citep{chen2026harnessvln}, support lifelong sessions \citep{zhao2026navharness},
and carry user and scene context across sessions \citep{wang2026carryon}.

\paragraph{Our Position.}
We count the harness, that is, the non-model components that assemble the context for
S2's decisions, carry them out, and verify their outcomes, as part of Embodied S2 rather
than as a layer around the model. Harness cards disclose its configuration so that
system-level results can be attributed.

\subsection{Memory and World Models}
\label{sec:rw-memory}

Explicit world memories for robots include open-vocabulary 3D scene graphs
\citep{gu2024conceptgraphs}, retrieval-augmented spatiotemporal memory
\citep{anwar2025remembr}, memory snapshots for exploration \citep{yang2025threedmem}, and
combined long- and short-term memories \citep{wang2025karma}. Policies can instead carry
memory internally \citep{shi2026memoryvla,torne2026mem}. User and long-term context remains
less developed: in one study, agents struggled to apply sequential user
patterns to planning \citep{kwon2025embodied}. Recent harnesses keep memories of people and preferences
\citep{jin2026glimpse} and record the outcomes of interactions in the scene graph
\citep{banwasi2026messymem}. In one manufacturing study, an external world-state model
improved state tracking \citep{tang2026vlmdewm}. Predictive world models have progressed
from learned simulators \citep{ha2018world,hafner2025mastering} to generative interactive
environments \citep{bruce2024genie,deepmind2025genie3}, physical-AI platforms
\citep{nvidia2025cosmos,nvidia2026cosmos3}, and self-supervised video world models used for
planning \citep{assran2025vjepa2}. They increasingly enter the
decision and evaluation loop, both for imagination-based search
\citep{zhang2026worldactionplanner} and for evaluating policies
\citep{gemini2025evaluating}. Criteria have also been proposed for when their verdicts can
count as evidence \citep{oefinger2026admissibility}.

\paragraph{Our Position.}
We treat cross-session world, user, and body state as a harness responsibility: each item
records when and how it was obtained and can be revised. World models supply evidence for S2 but make neither kind
of S2 decision, so they do not act as S2.

\subsection{Benchmarks}
\label{sec:rw-bench}

Embodied benchmarks range from instruction following in simulated homes
\citep{shridhar2020alfred} and suites of manipulation tasks
\citep{liu2023libero,zhou2025liberopro,li2024behavior1k} to real-to-sim evaluation
\citep{li2024simpler} and crowdsourced real-world evaluation \citep{atreya2025roboarenax}.
Benchmarks for embodied agents and S2 include EmbodiedBench, the Embodied Agent Interface,
VLABench, RoboCerebra, VSI-Bench, RoboBench, and ASIMOV
\citep{yang2025embodiedbench,li2024embodied,zhang2025vlabench,han2025robocerebra,yang2024vsibench,luo2025robobench,sermanet2025asimov}.
Closed-loop benchmarks from 2026 include RoboDojo for manipulation policies
\citep{chen2026robodojo} and RobotWorld, SCOPER, LIBERO-Agent, and CodeActionBench for agents
\citep{yang2026robotworld,wei2026scoper,diao2026liberoagent,lyu2026codeactionbench}.
Embodied First Exam evaluates coding agents across diverse simulated robots and tasks, allowing them to use direct control, generated programs, or trained policies under different protocols \citep{wang2026embodiedexam}.
Other studies test the reliability of success detectors, runtime monitors, and safety
refusals under robot control
\citep{navasardyan2026failbench,ahmedt2026rescue,sun2026roboharm}.
Section~\ref{sec:evaluation} discusses many of these benchmarks.

\paragraph{Our Position.}
Many benchmarks evaluate the model alone or the whole stack without stating
which. Few fix S1 and vary S2 \citep{han2025robocerebra,wei2026scoper}, and few score
verification directly; exceptions include FailBench \citep{navasardyan2026failbench}, RESCUE
\citep{ahmedt2026rescue}, and the completion-claim accuracy of CodeActionBench
\citep{lyu2026codeactionbench}. We therefore evaluate at two levels, model and system, with
a harness card for every result.

\subsection{Self-Improvement and Continual Learning}
\label{sec:rw-self}

Before 2025, robots improved through autonomous data collection and relabeling
\citep{xiao2023dial,bousmalis2023robocat,ahn2024autort,zhou2024autonomous} and through
reward functions written by language models \citep{ma2024eureka}. Weight-level
improvement of S1 policies now uses learned progress signals
\citep{ghasemipour2025selfimproving}, reinforcement learning from experience
\citep{pi2025pistar06}, fleet learning during deployment \citep{wang2026lwd}, and agentic
loops that verify rollouts and refine policies \citep{liu2026mobiagent,xiao2026enpire}.
Orchestrators and policies can co-evolve \citep{zhang2026robocop}. Around frozen
foundation models, systems revise tools, skills, and harnesses
\citep{tian2026physevo,wang2026reconstruct,lu2026aspire,elmaaroufi2026rho,lee2026emho},
and some co-evolve policy, curriculum, and judge \citep{zhou2026verifine}. Some
results are cautionary. In one simulation study, accepted changes kept passing their
tests while the target task never succeeded \citep{wang2026what}. In another, accepting
every proposed revision degraded a harness, whereas a champion--challenger rule improved
it \citep{wei2026scale}. In a third, continual-learning gains did not grow with more
samples \citep{huang2026embodiedclbench}.

\paragraph{Our Position.}
Whereas S2 adapts to verified outcomes within a task, a claim of improvement asserts that
a retained change helps later work. Such a claim should pass the three-part test of
Section~\ref{sec:rsi}: gains on held-out tasks, no regression, and separate contributions
of the S2 foundation model, the harness, and S1. This test is stricter than the standard
of most current reports but weaker than the recursive-gain criterion of physical recursive
self-improvement \citep{zhou2026who}.

\section{Applications: What Agentic Robotics Changes}
\label{sec:applications}

Embodied S2 could expand the range of work that robots undertake and improve how reliably and efficiently they complete it.
For each of several representative domains, we ask what earlier approaches already deliver, where they stop, what Embodied S2 could add, and how far reported deployments reach.

Two earlier approaches set the baseline. \emph{Programmed automation}, from industrial arms
to warehouse systems, executes engineered motions and rules with high speed and precision,
but only where the environment and the task are fixed in advance. Each new product, layout,
or exception requires engineering effort. \emph{Learned policies}, including VLA models,
generalize across objects and scenes within the skills on which they were trained, and now
complete whole tasks in homes they have never seen \citep{pi2025pi05,figure2026helix25}.
Each new task, however, still requires demonstrations, and a policy does not by itself
decide which task to do next, check that the result is correct, or recover when it is not.
In reported deployments, these responsibilities fall to remote experts and operators: \emph{human
backup performs the role of S2}, a pattern that recurs across the domains below.

If Embodied S2 took over these responsibilities, three changes would follow. First, a robot
could be given new work by instruction rather than by reprogramming or new demonstrations,
which would make automation worthwhile for varied, low-volume tasks that never justified
the engineering. Second, a robot could verify its own results and handle routine
exceptions, so that one person could supervise many robots by exception instead of guiding
one. Third, a robot could carry state across tasks and days, a prerequisite for entrusting
it with a workflow or a role rather than a single task. We state the
reach of each domain on the ladder of delegable work (Section~\ref{sec:ladder};
Figure~\ref{fig:teaser}), from a single primitive (L1) through a task (L2) and a workflow
(L3) to a standing role (L4). Because people still fill much of the role of S2, human interventions per
task, or per shift for continuous operations, should become a standard deployment metric
alongside reliability and execution efficiency. Table~\ref{tab:applications} summarizes
the reports and our assessment for each domain.

\begin{table}[!tbp]
\centering\small
\caption{\textbf{Application domains: what earlier approaches deliver, what Embodied S2
could add, and the reach reported so far.} Levels of delegable work are L1 primitive, L2
task, L3 workflow, and L4 role; the level needed and the contributions of S2 are our
assessment. A reported duration of operation is not a level reached: no report fixes in
advance all the conditions that a level claim requires (Section~\ref{sec:eval-how}).
Multi-day laboratory runs show sustained operation in a
restricted domain rather than a role held across open-ended work.}
\label{tab:applications}
\begin{tabular}{@{}L{0.14\linewidth}L{0.27\linewidth}L{0.24\linewidth}L{0.27\linewidth}@{}}
\toprule Domain & Earlier approaches and where they stop & What Embodied S2 could add &
Reported reach and evidence; level needed \\ \midrule
Home & Single-purpose appliances for fixed chores; learned policies complete single tasks
in unseen homes; remote experts bridge longer chores & Interpreting open requests;
sequencing chores; verification and recovery; user and home memory; interruption; privacy &
Task; company-run evaluation in unseen homes; remote experts disclosed for some products.
Needed: L3 to L4 \\ \addlinespace
Manufacturing & Programmed cells for fixed products, with every change an engineering
project; humanoid pilots on short tasks & Per-cycle verification; exception handling;
re-tasking from work instructions & Primitive to task, within shifts of up to ten hours;
company reports; intervention counts not disclosed. Needed: L3 to L4 \\ \addlinespace
Logistics & Large fleets in structured warehouses; items that the picking robot cannot
handle go to people & Exception handling; fleet orchestration & Primitive level per robot;
fleet operation in structured warehouses; company reports; human takeover disclosed for
picking. Needed: L3 to L4 \\ \addlinespace
Inspection & Preplanned routes; recorded data interpreted by people & Verified reading of
physical state; follow-up on anomalies; persistent site model & Task to workflow per
inspection round, in structured sites; company reports. Needed: L4 \\ \addlinespace
Healthcare and eldercare & Delivery between fixed points; eldercare only in trials &
Varied requests; timing of intervention; long-term memory; safety & Task for hospital
delivery (company report); eldercare only in trials. Needed: L4 for eldercare \\
\addlinespace
Laboratories & Fixed protocols on automated equipment; autonomous runs in restricted
domains & Choosing the next experiment; outcome verification; provenance & Sustained
operation for 8 to 17 days in structured laboratories (peer-reviewed). Needed: L4 \\
\addlinespace
Agriculture and mobility & Machines that each perform one job; vehicles in restricted
operating domains & Switching between jobs from observations; safety case; real-time S0 and
S1 & Primitive (agriculture; manufacturer reports); task to workflow per trip, in
restricted domains (mobility; company reports). Needed: L3 to L4, in wider domains \\
\bottomrule
\end{tabular}
\end{table}

\subsection{Homes}
Homes are the most numerous and the least structured of these settings. Single-purpose
appliances such as robot vacuum cleaners work reliably because their task is fixed.
General-purpose home robots are now emerging. Figure AI reports 56\% whole-task success in
30 unseen homes in a company-run evaluation that counts any rollout requiring a safety
intervention as a failure \citep{figure2026helix25}. Consumer humanoids are offered for
sale, although their shipping volumes have not been disclosed. According to its
manufacturer, the 1X NEO is sold with remote supervision by a company expert for chores
that the robot cannot yet perform alone \citep{onex2026neoproduct}. An assessment by MIT
Technology Review noted that the robot still needs a remote operator for most tasks
\citep{condliffe2026breakthroughs}. Reported home autonomy thus reaches L2, the task level,
and remote operation bridges the rest.

Much of what the remote operator supplies is the work of S2. A request such as ``get the
living room ready for guests, but leave the children's puzzle on the table'' leaves
most objects, all steps, and the completion criterion unstated. Fulfilling it requires interpreting what the
household wants, deciding where each item belongs, sequencing the work, checking the
result, and responding to interruptions. Throughout, the robot must remember the home, its
occupants, and their preferences while protecting their privacy. An Embodied S2 that
handled such requests would change what a home robot is. It would no longer be an
appliance for one chore or a teleoperated body, but a helper to which a household delegates
an open list of chores. Remote experts would then approve exceptions rather than perform
the work.

\subsection{Manufacturing and Logistics}
Factories and warehouses are where robots already work at large scale
\citep{ifr2026industrial,amazon2025million}, and where the
limits of programmed automation are most visible. Industrial arms repeat engineered motions
with high speed and precision inside fixtures and cells. This approach pays off in
high-volume production but turns every change of product or layout into an engineering
project. Humanoids and learned policies are now entering these settings. Figure AI reports
that its Figure~02 deployment at BMW ran ten-hour shifts and contributed to the production
of more than 30,000 vehicles \citep{figure2025bmw}. Apptronik agreed to pilot its Apollo humanoid at Mercedes-Benz, runs data collection at
customer sites including Mercedes-Benz, and uses the resulting fleet data to train Gemini Robotics models
\citep{apptronik2024mercedes,apptronik2026robotpark,gemini2025robotics}. Hyundai has
announced plans to deploy Boston Dynamics' Atlas in its plants
\citep{bostondynamics2026rmac}. Skild AI reports more than 60 paying customers ten months
after its first commercial deployment; mobility accounts for 10\% of its revenue, and the
remaining 90\% comes mainly from manipulation tasks \citep{skild2026pillar}. In logistics, Agility reports that its Digit
robots have moved more than 100,000 totes at GXO \citep{agility2025totes,agility2026gxo}.
Amazon reports operating more than a million robots coordinated by what it describes as a
fleet foundation model \citep{amazon2025million}, and its tactile picking robot asks human
workers to take over items that it cannot handle \citep{amazon2025vulcan}. All of these
figures are company-reported.

Because each cycle in these settings is short and repetitive, Embodied S2 would contribute
less within a cycle than around it. It could verify each cycle and handle exceptions such
as a jam or an unfamiliar item, diagnosing the cause and then either retrying differently
or handing over with a precise report. It could also orchestrate fleets and re-task robots
when production changes. Re-tasking may bring the largest gain. An Embodied S2 that
composed existing skills into a new sequence from a work instruction, and verified the
first parts before running at volume, would turn a reprogramming project into a
configuration step. Automation could then extend from high-volume lines to high-mix,
low-volume work that fixed automation has rarely reached.

\subsection{Inspection}
Inspection is an early commercial application of embodied reasoning models, which are
candidate S2 foundation models. It suits current models because it tolerates latency and rewards
verified reading of physical state, and it illustrates the step from collecting data to
making judgments. A conventional inspection robot repeats a preplanned route and records
data for people to interpret. Boston Dynamics reports more than 1,500 Spot robots with
customers \citep{bostondynamics2026spot} and has integrated Gemini Robotics-ER into its
inspection software for reading gauges and interpreting site conditions
\citep{deepmind2026er16,bostondynamics2026aivi}. Google reports a demonstration in which Gemini~4 Argon controls
a real Spot after practicing the task in a simulated twin that it built itself
\citep{google2026argonrobotics}. An Embodied S2 could connect reading and control by acting on what it reads.
When a gauge is out of range, it could look again from another angle, check the related
valve, and file a report with the evidence attached, instead of leaving every anomaly for a
person to find in the recordings. At the role level, inspection would also require a
persistent model of the site that is updated on every round.

\subsection{Healthcare and Laboratories}
Hospital delivery robots already navigate elevators and corridors autonomously
\citep{diligent2024elevator}, but they carry items between fixed points. An Embodied S2
could turn such a robot into an assistant for varied requests, such as fetching a
particular item from a supply room. Eldercare is the most demanding domain considered here.
Delegating it to a robot would require responsibility at the role level (L4), long-term
memory of a person, well-timed intervention, and the strongest privacy and safety
protections. Current evidence is limited to trials.

Laboratory automation has long executed fixed protocols, and structured laboratories also
host autonomous research campaigns that run continuously for days. The A-Lab synthesized 36 of 57
target compounds over 17 days of continuous operation \citep{szymanski2023autonomous},
although the novelty of its products and the automated analysis that identified them were later disputed; the count given here is the corrected one
\citep{leeman2024challenges,szymanski2026author}. A
mobile robotic chemist ran 688 experiments over eight days \citep{burger2020mobile}.
These runs show sustained operation in a restricted domain
rather than a role held across open-ended work. A language-model agent, Coscientist, has also planned and executed chemistry experiments
\citep{boiko2023autonomous}. In open-ended work, an Embodied S2 could decide
which experiment to run next given the results so far, verify that each step produced what
it should, and record provenance. In laboratories,
verification of outcomes, rather than manipulation, is likely to be the bottleneck.

\subsection{Agriculture and Mobility}
Agricultural robots operate commercially at the primitive level
(L1): laser weeders work in 15 countries according to their manufacturer, which reports more than 200 growers across its laser-weeding and tractor-autonomy products
\citep{carbonrobotics2026laserweeder}, but each
machine performs one job. An Embodied S2 could let one platform scout a field, decide where
weeding or harvesting is needed, and switch between these jobs as conditions change.
In mobility, robotaxis provide hundreds of thousands of paid rides per week in restricted
operating domains \citep{korosec2026waymo}. Cars and drones are physical agents as well,
and the same S2--S1--S0 organization applies to them. Their history also suggests how
agentic robotics is likely to scale: domain by domain, with each operating domain widened
only as evidence of safety accumulates.

\paragraph{Common Patterns.}
Across these domains, evidence of unaided work in unstructured settings is strongest at L2. 
Some demonstrations extend toward workflows, but their reports do not yet establish all the conditions needed for a reliable L3 claim.
Longer operation has been reported only in structured or restricted domains, and wherever the fallback is disclosed, it is a person. 
The same three changes recur: work assigned by instruction rather than by engineering or new data, supervision by exception rather than
one operator per robot, and responsibility extended from single tasks to workflows and
roles. Each change would widen the range of work for which deploying a robot is
worthwhile. The research directions in Section~\ref{sec:directions} address what delivering these changes would take.

\section{Contributors}
\label{sec:contributions}

\begingroup
\renewcommand{\arraystretch}{1.25}
\noindent\begin{tabular*}{\linewidth}{@{\extracolsep{\fill}}llll@{}}
Qingfeng Lan\textsuperscript{*} & Duo Li & Siyao Li & Yiheng Lyu \\
Shijian Lu & Zhenghao Qi\textsuperscript{*} & Shizun Wang\textsuperscript{*} & Xinchao Wang \\
Ming Yang\textsuperscript{*} & Zhiqin Yang\textsuperscript{*} & Zuhao Yang\textsuperscript{*} & Tao Yu\textsuperscript{*} 
\end{tabular*}
\par
\endgroup

\smallskip
\noindent Contributors are listed in alphabetical order by surname, then by first name within each surname, read from left to right across each row. \textsuperscript{*} denotes core contributors.

\bibliographystyle{plainnat}
\bibliography{references}

\end{document}